\documentclass[journal,letterpaper]{IEEEtran}
\usepackage{amsmath,amssymb,amsfonts,mathtools,bm}
\usepackage{siunitx}

\usepackage{booktabs}
\usepackage{multirow}
\usepackage{tabularx}
\usepackage{makecell}
\usepackage[table]{xcolor}

\usepackage{graphicx}
\usepackage{float}
\graphicspath{{figures/}}
\usepackage{tikz}
\usetikzlibrary{positioning,arrows.meta,fit,calc,patterns,decorations.pathreplacing}

\usepackage{algorithm}
\usepackage{algpseudocode}

\usepackage{cite}
\usepackage[hyphens]{url}
\usepackage[colorlinks=true,allcolors=blue]{hyperref}
\usepackage[capitalize]{cleveref}
\crefname{figure}{Fig.}{Figs.}
\Crefname{figure}{Fig.}{Figs.}
\crefname{section}{Sec.}{Secs.}
\Crefname{section}{Sec.}{Secs.}

\usepackage{xspace}
\newcommand{\pol}[1]{\textit{#1}}
\newcommand{\EarthFlatRigid}{\pol{Earth-Flat-Rigid}\xspace}
\newcommand{\EarthFlatRFT}{\pol{Earth-Flat-RFT}\xspace}
\newcommand{\EarthRoughRigid}{\pol{Earth-Rough-Rigid}\xspace}
\newcommand{\EarthRoughRFT}{\pol{Earth-Rough-RFT}\xspace}
\newcommand{\MoonFlatRigid}{\pol{Moon-Flat-Rigid}\xspace}
\newcommand{\MoonFlatRFT}{\pol{Moon-Flat-RFT}\xspace}
\newcommand{\MoonRoughRigid}{\pol{Moon-Rough-Rigid}\xspace}
\newcommand{\MoonRoughRFT}{\pol{Moon-Rough-RFT}\xspace}
\newcommand{\FtFlatRigid}{\pol{Earth$\to$Moon-Flat-Rigid}\xspace}
\newcommand{\FtRoughRFT}{\pol{Earth$\to$Moon-Rough-RFT}\xspace}
\newcommand{\JointFlatRigid}{\pol{Earth+Moon-Flat-Rigid}\xspace}
\newcommand{\JointRoughRFT}{\pol{Earth+Moon-Rough-RFT}\xspace}
\newcommand{\JointRoughRigid}{\pol{Earth+Moon-Rough-Rigid}\xspace}

\newcommand{\hdrstrut}{\rule{0pt}{2.3ex}}
\renewcommand{\arraystretch}{1.15}

\newcommand{\tabnote}[1]{\par\smallskip\parbox{\linewidth}{\footnotesize #1}}

\IEEEoverridecommandlockouts
\begin{document}
\title{Do Earth-Trained Humanoid Controllers Walk on the Moon? Gravity, Terrain, and Soil Transfer}
\author{Guillem Casadesus~Vila, Sri~Aditya~Deevi, and Grace~X.~Gao%
\thanks{This work has been submitted to the IEEE for possible publication. Copyright may be transferred without notice, after which this version may no longer be accessible.}%
\thanks{The authors are with the Department of Aeronautics and Astronautics, Stanford University, Stanford, CA 94305, USA. {\tt\small \{guillemc, adeevi, gracegao\}@stanford.edu}}}
\maketitle
\begin{abstract}
% Context: how to train lunar humanoid locomotion for gravity, terrain, and soil
Humanoid robots could aid lunar operations, but how to train their locomotion controllers for lunar gravity, terrain, and soil is unclear. We compare 12 strategies for a simulated Unitree G1 across gravity, terrain, and rigid or granular 3D resistive force theory (3D RFT) contact. Earth-trained policies keep 0.64 to 0.75 survival at the Moon on flat rigid ground in 10~s episodes, but falls continue, so longer episodes would see lower survival. Moon-trained and fine-tuned policies, in contrast, fall in every Earth episode, nearly all within 2~s. Rough-terrain training helps on uneven rigid ground only at training gravity. 3D RFT soil training helps on Earth sand but little on lunar soil. On three held-out soil models, Moon-trained policies lose no detectable progress at lunar gravity. Among tested strategies, only gravity-randomized rough 3D RFT training keeps mean survival above 0.83 at Earth and 0.95 at the Moon on all key tests up to 20\textdegree. Its ablation without soil training differs consistently only on soil at Earth gravity, falling in most sand episodes. These predictions depend on the contact model and soil stiffness, so lunar soil claims require agreement across soil models.
\end{abstract}


\IEEEpeerreviewmaketitle
\section{Introduction}
\label{sec:intro}

\begin{figure}[!tb]
  \centering
  \includegraphics[width=\columnwidth]{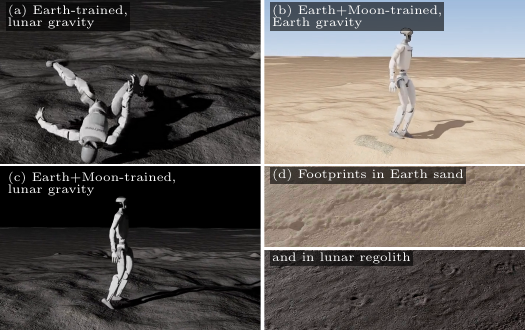}
  \caption{\textbf{An Earth-trained controller falls at lunar gravity, while training across both gravities walks at both.} (a) A G1 policy trained at Earth gravity on rough terrain with 3D resistive force theory (3D RFT) soil (Earth-Rough-RFT) falls on a held-out lunar tile at lunar gravity. (b, c) A policy trained with gravity randomized from Earth to the Moon (Earth+Moon-Rough-RFT) walks on sand at Earth gravity and on regolith at lunar gravity. (d) Footprints in material point method (MPM) sand and regolith.}
  \label{fig:hero}
\end{figure}

% Context: renewed lunar surface activity motivates humanoids, walking on the Moon is hard even for people, and lunar gravity, terrain, and soil all differ from the Earth conditions controllers are trained for
NASA's Artemis program aims to establish a sustainable human presence at the Moon~\cite{nasa_artemis_2020}. Humanoid robots could take mundane and dangerous surface tasks off future lunar crews~\cite{nasa_experience_2023}, and a humanoid form suits tools and interfaces designed for people, a premise of NASA's Robonaut~2~\cite{diftler_robonaut_2011}. However, walking on the Moon is hard even for people, as suited Apollo crews often loped~\cite{carr2009apollo} and fell 27 times across the Apollo surface missions~\cite{thuro_characterization_2021}. Learned humanoid controllers now walk over rough, deformable, and sloped terrain on Earth~\cite{radosavovic_realworld_2024,radosavovic_challenging_2024}, while lunar operation changes several of these conditions at once (\Cref{fig:hero}). Compared with Earth, gravity falls to \SI{1.62}{\metre\per\second\squared}, craters and rocks, whose size statistics are measured~\cite{hartmann_saturation_1984,di_rock_2016}, change the available footholds, and the regolith yields under contact~\cite{carrier_physical_1991}.

% Gap: prior work trains for one condition at a time, so gravity randomization from Earth to Moon, the Earth re-test (needed because a lunar controller is verified on Earth), and the cross-soil-model check are open
Prior learned-locomotion studies adapt one recipe to one condition, such as reduced gravity on rigid ground~\cite{klein_training_2025,arm_efficient_2025} or granular media at Earth gravity~\cite{kamohara_granular_2026}. Similarly, quadruped studies pair lunar gravity with compliant ground~\cite{belmontebaeza_learning_2026} or granular contact~\cite{vyas_locomotion_2026}. For the Unitree G1 humanoid, Klein and Roberts compare an Earth baseline, lunar fine-tuning, and lunar training from scratch~\cite{klein_training_2025}. However, to our knowledge, no humanoid study has randomized gravity across the Earth-to-Moon range, checked whether policy comparisons hold across soil models, or re-tested lunar-adapted policies at Earth gravity, which matters because a lunar controller would be tested on Earth before flight.

% Scope: how to train a lunar controller, 12 policies as candidate strategies tested at both gravities, one reward family, half the soil-training environments on 3D RFT, SONIC as a frozen reference on rigid ground
We therefore ask how a humanoid locomotion controller should be trained for lunar gravity, terrain, and soil, and what each training strategy buys and costs. We train 12 policies for the 29-degree-of-freedom G1 as candidate strategies, plus one ablation without soil training, with one reward family whose settings depend on gravity (\Cref{fig:training}), and test each at both gravities. Eight are trained from scratch and cover every combination of Earth or Moon gravity, flat or rough terrain, and rigid or granular ground. The other four are two Earth-to-Moon fine-tunes and two gravity-randomized policies trained from scratch. Soil training places half the environments on rigid ground and half on granular ground modeled with resistive force theory (RFT) in its three-dimensional form (3D RFT)~\cite{li_terradynamics_2013,agarwal_mechanistic_2023}. As a frozen reference, we also test the released SONIC planner and tracker pipeline~\cite{luo_sonic_2026} on rigid ground.

% Contribution: the analyses of the letter, one per item, findings live in the abstract and results
The contributions of this letter are the following:
\begin{itemize}
  \item A comparison of the 12 training strategies at both gravities, which shows that transfer between the two gravities is asymmetric.
  \item An analysis of what rough-terrain and soil training add at each gravity.
  \item A comparison of the gravity-randomized policies with the best single-gravity policies at both gravities.
  \item A cross-check of the Earth- and Moon-trained policies on three soil models held out of training.
\end{itemize}

\section{Related Work}
\label{sec:related}

\subsection{Humanoid Controllers and Adaptation}
\label{sec:related_s2r}
% Context: task-specific humanoid locomotion is trained at Earth gravity, and randomizing simulator parameters, here gravity, is the standard route to robustness that we test, and ULTRA randomizes gravity only as a small perturbation, so we analyze the Earth-to-Moon range
History-based policies transfer humanoid locomotion to hardware~\cite{radosavovic_realworld_2024}, including blind walking over rough, deformable, and sloped surfaces~\cite{radosavovic_challenging_2024}, but these policies are developed and deployed at Earth gravity. Randomizing simulator parameters makes a policy robust to changed dynamics~\cite{peng_simtoreal_2018}, so randomizing gravity is one way to train a single policy for Earth and the Moon. Humanoid controllers have so far randomized gravity only slightly, as $\pm\SI{0.1}{\metre\per\second\squared}$ in ULTRA~\cite{he_ultra_2026}, so we analyze randomization across the Earth-to-Moon range.

% Context: generalist whole-body controllers have not been tested under reduced gravity and their transfer is unclear, so SONIC is one focus of the study
Generalist whole-body controllers instead learn broad motor skills from human motion. SONIC~\cite{luo_sonic_2026}, for example, pairs a kinematic planner with a motion-tracking controller trained on about 700 hours of motion capture. However, to our knowledge, no such controller has been tested under reduced gravity, so SONIC's transfer to the Moon is unclear.

\subsection{Reduced-Gravity and Planetary Legged Locomotion}
\label{sec:related_lowg}
% Context: reduced-gravity robots from SpaceBok and Olympus hopping to quadrupedal walking on lunar ground and lunar fine-tuning of the G1, and we add gravity randomization across the Earth-to-Moon range, the Earth re-test, and the cross-soil-model comparison
SpaceBok and Olympus study hopping and in-flight attitude control under reduced gravity~\cite{kolvenbach_spacebok_2019,olsen_olympus_2025}. For walking, Arm et al. study power-optimized quadrupedal gaits under scaled gravity with a base-offload rig and evaluate them on flat rigid ground~\cite{arm_efficient_2025}. Belmonte-Baeza et al. train a quadrupedal mobile manipulator at lunar gravity on rough, hilly, and cratered terrain with randomized friction. They then test it on compliant ground, unseen in training, whose stiffness and damping are set for a regolith simulant~\cite{belmontebaeza_learning_2026}. Vyas compares rigid and granular foot contact for quadrupeds on lunar regolith and finds that granular contact makes training harder, changes the gait, and raises energy use~\cite{vyas_locomotion_2026}. For humanoids, Klein and Roberts train the G1 with Unitree's training code~\cite{unitree_rl_lab_2025} and find that lunar fine-tuning matches lunar training from scratch in less than half the time~\cite{klein_training_2025}. This comparison, however, excludes the \SI{14.5}{\hour} of Earth pretraining. Building on these studies, we add Earth-to-Moon gravity randomization, an Earth re-test of the lunar-adapted policies, and a comparison across soil models.

% Context: at low gravity human walking loses pendular exchange and bouncing gaits cost about as much as Earth walking, so a lunar gait need not look like an Earth walk, Froude scaling only partly predicts gait changes (the walk-run transition Froude number rises at 0.2 and 0.1 g), and we report gait against Froude at both gravities and compare toe-first landing at the Moon with Klein and Roberts
Human gait changes at low gravity, so a lunar gait need not resemble an Earth walk. In simulated reduced gravity, the pendular energy recovery of walking at \SI{1}{\metre\per\second} averages \SI{64}{\percent} from 0.5 to 1\,g but falls to \SI{48}{\percent} at 0.25\,g~\cite{griffin_walking_1999}. Moreover, at lunar gravity skipping and running cost about as much as walking on Earth~\cite{pavei_skipping_2015}. In such simulations, humans switch from walking to running at a Froude number near 0.5 from 1 to 0.4\,g but at 0.83 at 0.2\,g and 1.13 at 0.1\,g~\cite{kram1997effect}. In actual lunar gravity, they switch at a mean Froude number of 1.39~\cite{de_witt_lunar_2014}, so equal Froude numbers do not imply equal gaits. We therefore relate gaits at both gravities through the Froude number and compare toe-first landings at the Moon with the toe-dominant walking of Klein and Roberts~\cite{klein_training_2025}.

\subsection{Granular Contact and Prediction Credibility}
\label{sec:related_soil}
% Context: reduced-order granular models and granular legged locomotion at Earth gravity, which we extend to lunar gravity
Resistive force theory predicts local intrusion stresses in granular media~\cite{li_terradynamics_2013}, and its three-dimensional form covers arbitrary intruder motion~\cite{agarwal_mechanistic_2023}. For bipeds, Chen et al. validate sinkage and slip on sand~\cite{chen_bipedal_2026}, and Kerimoglu et al. show that foot-induced yielding and foot shape change performance on granular slopes~\cite{kerimoglu_flowable_2026}. GM-Loco trains a G1 with 3D RFT at Earth gravity and checks the policy in the material point method (MPM)~\cite{newton_contributors_2025} and on basalt, dry sand, and beach sand~\cite{kamohara_granular_2026}, and we extend its 3D RFT training to lunar gravity.

% Gap: experiments disagree on how granular resistance scales with gravity, so we test whether comparisons hold across models
How granular resistance scales with gravity remains open, because experiments disagree on whether it is proportional to gravity or nearly independent of it. In fluid-saturated beds of spheres, where buoyancy sets the effective gravity, low-velocity drag is proportional to effective gravity over a 20-fold range~\cite{costantino_lowvelocity_2011}. In contrast, the penetration depth of low-velocity impacts into sand is independent of effective gravity over the tested range~\cite{murdoch_experimental_2017}, and cone resistance on a lunar simulant is only 9 to \SI{29}{\percent} lower at 1/6\,g than at 1\,g~\cite{chen_lowgravity_2026}. Moreover, parabolic-flight wheel tests show that gravity acting on the soil itself changes wheel mobility at matched load~\cite{niksirat_reducedgravity_2020}. Because the soil models encode different scalings (\Cref{sec:rft}) and no Earth test resolves this for lunar regolith, we do not pick the correct soil model but test whether policy comparisons agree in sign across models.

\section{Method}
\label{sec:method}
\begin{figure}[!tb]
  \centering
  \input{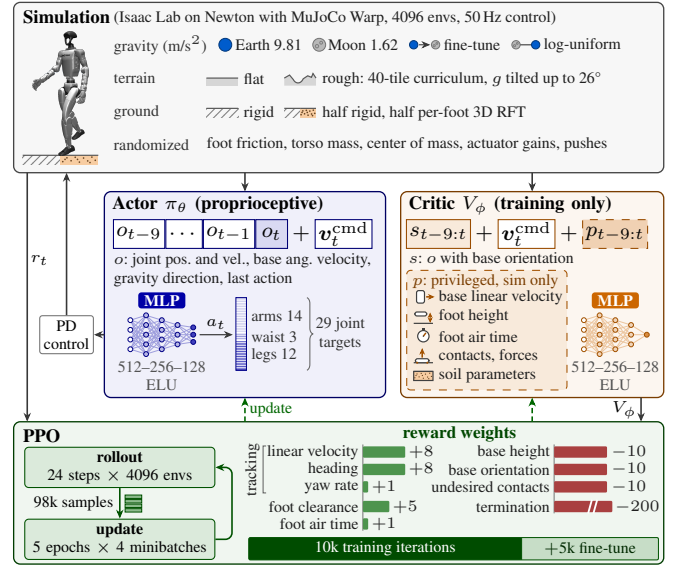}
  \caption{\textbf{Training pipeline.} The environment sets gravity, terrain geometry, and ground for each policy. The actor maps ten frames of proprioception and the velocity command to 29 joint proportional-derivative (PD) targets at \SI{50}{\hertz}. The critic also observes privileged base velocity, foot contacts, and soil parameters. Proximal policy optimization (PPO) updates both networks from the GM-Loco reward terms, with selected terms shown at their final weights. At test time, the actor is frozen and purely proprioceptive.}
  \label{fig:training}
\end{figure}

% Scope: section overview, 12 candidate training strategies with one reward family, SONIC as frozen reference
This section describes the 12 policies we train as candidate strategies for a lunar locomotion controller. All share one reward family and differ only in gravity, terrain geometry, ground, initialization, soil ranges, and the reward settings tied to gravity (\Cref{fig:training,sec:recipe,sec:lunar}). Their names give training gravity (Earth, Moon), geometry (Flat, Rough), and ground (Rigid, RFT), while Earth$\to$Moon marks a fine-tune and Earth+Moon marks gravity randomization (\Cref{tab:gravity}).

\subsection{Training Setup}
\label{sec:recipe}
% Method: GM-Loco reward terms on Unitree's G1 model, trained with PPO in Newton with MuJoCo Warp, our history and mirror augmentation
Every trained policy uses the reward terms of GM-Loco~\cite{kamohara_granular_2026}, which we reimplement. We train with proximal policy optimization (PPO) and mirror augmentation in Isaac Lab~3.0, on the Newton engine with its MuJoCo Warp solver~\cite{newton_contributors_2025}, using the 29-degree-of-freedom G1 model and actuators~\cite{unitree_rl_lab_2025}.
% Source: recipe LuPNT-isaac src/lupnt_lab_tasks/g1_best/env_cfg.py:1-35,104-114 (robot, scene, solver from g1_unitree RobotEnvCfg); actor/critic groups gm_loco/agents/rsl_rl_ppo_cfg.py:24 and gm_loco/env_cfg/observation_cfg.py:68-263,301-310 (history 10); mirror augmentation gm_loco/agents/rsl_rl_ppo_cfg.py:50-53 (use_data_augmentation=True), inherited by g1_best/agents/rsl_rl_ppo_cfg.py:9; Aditya-progress-2026-10-16.txt sec. 5.1. Full observation, reward, and curriculum lists are in the supplement.

% Method: randomized friction, commands, and pushes, noise-free observations as a stated simplification
Training samples the commands and randomizes foot friction, torso mass, center of mass, actuator gains, and pushes, while the observations carry no noise.
% Note: observation noise is configured (observation_cfg.py:76-119) but never enabled: PolicyHistoryCfg.__post_init__ (observation_cfg.py:188-190) overrides PolicyCfg's without calling it, so enable_corruption stays False (env.yaml observations.policy.enable_corruption: false). The paper therefore makes no noise-robustness claim.

% Method: soil term, GM-Loco per-foot 3D RFT with two force corrections
RFT-trained policies add the per-foot 3D RFT force of GM-Loco~\cite{kamohara_granular_2026} on half of the environments, without GM-Loco's teacher-student terrain encoder. We sum the 3D RFT stresses, which depend on the orientation, velocity direction, and depth of each element~\cite{agarwal_mechanistic_2023}, over a 5 by 5 element grid on each sole, while rocks remain rigid. Our implementation matches the published coefficients and the per-element forces of the authors' code, which verifies the code, not lunar material behavior.
% Source: coefficients match the published table to 5e-6; normal force linear in depth to 0.17% over a 600x stiffness range (vault RAL-results-2026-09, 2026-10-05).
% Source: EMA filter g1_best/env_cfg.py:165 (step2b_env_cfg.py:110); substep re-apply and COM torque gm_loco/soft_contact/actions_cfg.py:65-77 and gm_loco/newton_wrench_fix.py:1-20 (without it a standing robot sinks until the RFT force is 1.8x Earth / 2.5x Moon its weight, 1.08x with it); half rigid g1_study/terrain.py:46-56,72-95 and g1_best/env_cfg.py:85; union contact gm_loco/mdp/contact_state.py:12-13,38,57.

% Method: soil ranges per gravity, lunar range calibrated to bootprint stiffness
Earth training uses the soil ranges of GM-Loco~\cite{kamohara_granular_2026} (\Cref{tab:soil_models}). Because the 3D RFT force depends on density $\rho_c$ and internal friction $\mu$ only through $\rho_c\,g\,f(\mu)$, lunar training fixes $\mu$ at 0.9~\cite{agarwal_mechanistic_2023} and varies the density. We calibrate this range by having the G1 walk on each candidate soil at lunar gravity, and thus under its lunar weight, and matching its foot sinkage to the bootprint depth that the Apollo stiffness gives at the G1 foot pressure~\cite{carrier_physical_1991}. The gravity-randomized policies use the union of both ranges, so on soil the Earth and Moon conditions differ in soil strength as well as in gravity.
% Source: Agarwal 2023 PNAS SI sec. S4, Fig. S4 and Table S1: xi_n fit over mu_int 0.1 to 0.9. Verified 2026-10-20.
% Source: g1_best/env_cfg.py:55-57 (EARTH_SOIL, MOON_SOIL mu_surf 0.343, JOINT_SOIL union), :190-211 (rho_c = packing x bulk; Moon packing 1.0); dump env.yaml of Moon-Rough-RFT_s43 events.randomize_* (0.343-1.0, 0.9, 150-8000). Equivalent rho_c and f(mu) fit range: LuPNT-isaac docs/regolith_calibration/README.md. 8000/150 = 53. Calibration method: insim_calibration.md and Aditya-progress-2026-10-16.txt sec. 3.

\subsection{Policies and Training Budgets}
\label{sec:policies}
% Method: 2x2x2 grid of training choices, rough training, fine-tunes, gravity randomization
% Note (env mix, resolved 2026-10-27): S1 states the pinned software release and that training ran on several GPU machines, compared by seed means. Known: the 36 grid/fine-tune/randomized runs were trained by a coauthor on LuPNT-isaac lunar-g1-best 16cbc17 (IsaacLab-ea), host per run not recorded (design note: ws4/NAVLab RTX 4090); the rigid control EarthMoon-Rough-Rigid s42-44 on ws6 RTX 5090 at 12608bc. Eval: eval_best on ws6 (cells at 03a7821) with later evaluator changes 00bd229 (eval joint-reset pin 10-16), 1063614 (soil recalibration), 9f30a4d; s42/s43 slope cells predate the pin, s44 is mixed (78 of 672 files predate it); a 15-cell repro under the current evaluator moved at most 0.023; soil cells are not bitwise deterministic (26 of 53 reruns changed survival by up to 0.055). Some tables mix ws6 and Sherlock cells. Sherlock image work (own Apptainer image, per-run run_env.json) is on LuPNT-private branch ral-sherlock-image 4f923e710, not yet run.
Eight policies trained from scratch, the grid policies, cross the training choices of gravity, geometry, and ground. The rigid Earth and Moon pairs differ only in gravity and the settings scaled with it (\Cref{sec:lunar}), whereas the RFT pairs also differ in soil range (\Cref{sec:recipe}), so we compare them as combinations of training choices rather than as a one-factor contrast. Rough training likewise combines 40 tiles, 10 from each of the four tile families (\Cref{sec:terrain}), a terrain curriculum~\cite{rudin_learning_2022}, and gravity tilted by up to \SI{26}{\degree} toward a random direction in one third of episodes. The remaining four policies combine both gravities. The fine-tunes \FtFlatRigid and \FtRoughRFT continue their Earth parents for 5k iterations at lunar gravity with the lunar settings, including the lunar soil range for \FtRoughRFT. The gravity-randomized policies \JointFlatRigid and \JointRoughRFT instead train from scratch with gravity drawn per episode, log-uniform on $[1.62, 9.81]$\,\si{\metre\per\second\squared}. However, the observation carries only the direction of gravity, so these policies can infer its magnitude only from their history. To separate the soil-training part of \JointRoughRFT, we also train a rigid control, \JointRoughRigid, with the same gravity randomization and rough training on rigid ground. Because this control trained on another machine and was evaluated later, 15 cells of \JointRoughRFT re-evaluated with the control's evaluator changed survival by at most 0.023.
% Source: output/RAL_Humanoid/eval_c3_joint_rough_rigid/c3_summary.md, Caveats: the twin trained on the ws6 RTX 5090, EarthMoon-Rough-RFT on ws4 or a coauthor's 4090 host; re-running EarthMoon-Rough-RFT with the current evaluator (LuPNT-isaac 9f30a4d, ws6) reproduces 7 of 15 cells exactly and moves 8 by -1 to +6 of 256 episodes, at most 6/256 = 0.023; no verdict changes.
% Source: tilt g1_best/env_cfg.py:80,255 (26 deg, fraction 1/3) and g1_study/mdp.py:20-75 (per-episode draw, uniform angle, uniform azimuth); 40 tiles g1_study/env_cfg.py default_tile_families; promotion isaaclab terrain_levels_vel, g1_best/env_cfg.py:220.
% Source: fine-tunes run the Moon cell from model_10000 with GM_LOCO_RESUME_ITERATION=10000 (g1_best/__init__.py:8, curricula shifted so they stay finished, step2b_env_cfg.py:132); log-uniform g g1_study/mdp.py:61-62; joint range g1_best/env_cfg.py:78.

% Method: seeds, steps per run, no Earth-to-Earth control, SONIC as a frozen reference
Each policy has three seeds, evaluated at the final checkpoint. A from-scratch run of 10k PPO iterations takes about $9.8\times10^{8}$ environment steps, and a fine-tune, including its parent, about $1.5\times10^{9}$. The fine-tunes see no Earth episodes, and we run no Earth-to-Earth continuation as a control. SONIC~\cite{luo_sonic_2026} enters as a frozen reference whose planner has no gravity or terrain input, so its references stay Earth-timed and level. We run it in the same Isaac Lab environment with its released gains and map each command profile to the planner inputs. This port matches the planner output and first tracker observation of the reference implementation exactly, and on flat Earth ground it survives every episode with per-profile speed ratios within 0.09 of the released deployment (tolerance 0.10). We test it on rigid ground only and compare complete controllers, not architectures.
% Source: SONIC port check, S1 sec:sonic-supp (acceptance: planner output for a fixed input sequence, first tracker observation, Earth survival >= 0.95 and speed fraction within 0.10 of the released MuJoCo deployment on every profile; result: exact match, 256/256 survivors on flat Earth, speed ratios within 0.09).
% Source: max_iterations 10000 g1_best/agents/rsl_rl_ppo_cfg.py:12 (fine-tunes --max_iterations 5000, final model_15000); num_steps_per_env 24 gm_loco/agents/rsl_rl_ppo_cfg.py:21; num_envs 4096 gm_loco/rough_env_cfg.py scene and run dump env.yaml. 4096 x 24 = 98,304 steps/it; 10,000 x 98,304 = 9.83e8; 5,000 x 98,304 = 4.92e8; 15,000 x 98,304 = 1.47e9.

\subsection{Gravity-Dependent Design}
\label{sec:lunar}
% Method: Froude number, no gait clock
We relate gaits at the two gravities through the Froude number $\mathrm{Fr}=v^2/(gL)$, with walking speed $v$ and leg length $L=\SI{0.78}{\metre}$, so that at \SI{1.0}{\metre\per\second} Fr is 0.79 on the Moon and 0.13 on Earth. Because the policies have no gait clock, cadence and stride length emerge at each gravity.

% Method: energy normalization and gravity-scaled reward terms
We keep the energy and torque weights equal at both gravities and add no feedforward torque, so at the Moon the same weights act on smaller torques and powers. Only the contact thresholds and the landing-impact weight change with gravity. The contact-detection thresholds scale with the gravity of each episode, while the undesired-contact penalty and critic contact-force thresholds are fixed at their lunar values for lunar and gravity-randomized training. Lunar training also multiplies the landing-impact penalty by $9.81/1.62$.
% Source: g1_best/env_cfg.py:135-144 (undesired-contact threshold and privileged force filter x 1.62/9.81 on Moon and Joint cells, contact_impulse weight / (1.62/9.81) on Moon cells only, Joint keeps the Earth landing weight); run dump: undesired_contacts threshold 0.165, contact_impulse -0.0303 = -0.005 x 6.06.

\section{Experimental Setup}
\label{sec:setup}

\subsection{Terrain and Slopes}
\label{sec:terrain}
% Method: training and test tiles from one generator with disjoint seeds, crater, rock, and undulation statistics
Training and test tiles come from the same generator in LuPNT, an open-source lunar simulator~\cite{casadesusvila_lupnt_2025}, in four families with craters, rocks, or undulations alone or all three combined. Evaluation uses 32 held-out tiles per family from disjoint generator seeds, with 8 starts each. All tiles are \SI{8}{\metre} at \SI{0.1}{\metre} per pixel, and craters from \num{0.4} to \SI{2.4}{\metre} follow Hartmann's equilibrium size-frequency distribution~\cite{hartmann_saturation_1984}, rocks follow a size-frequency model applied to the Moon by Di et al.~\cite{di_rock_2016}, and undulations are a fractional Brownian surface anchored to the median slope of lunar highlands, \SI{4.66}{\degree} at a \SI{4}{\metre} baseline~\cite{cai2020meter}. However, the crater and rock statistics are extrapolated below their measured sizes, and rocks under \SI{0.2}{\metre} are not represented. The held-out undulation tiles are also smoother than the cited lunar value, with a median slope of \SI{1.6}{\degree} at the \SI{4}{\metre} baseline.
% Source: held-out tile statistics, LuPNT-private output/RAL_Humanoid/review_2026-10-24/summary.md sec. C19 (generator metric, median |dz|/L over x and y pairs, mean over the 32 undulation tiles): 1.59 deg at 4 m with 0.4 m margin, 1.92 deg (0.94-5.02) with 1.5 m margin; the generator anchors 4.66 deg on a 256 m field and high-passes each tile at 8 m.

% Method: slopes by tilted gravity, inclined mesh for SONIC
Slopes tilt the gravity vector over flat rigid ground by 5, 10, 15, 20, 26, and \SI{30}{\degree}, so only \SI{30}{\degree} lies beyond the \SI{26}{\degree} of rough training. Each grade is tested uphill and downhill, with the tilt about the axis perpendicular to the initial heading and the robot upright. Test foot friction is 0.5 and ground friction is 0.3, and because the solver uses the larger of the two, a foot cannot stand still on a grade steeper than the friction-cone limit $\arctan 0.5=\SI{26.6}{\degree}$ at either gravity. The \SI{26}{\degree} grade thus lies within \SI{0.6}{\degree} of this limit and \SI{30}{\degree} beyond it. Because survival checks only pelvis height and tilt, we also report traversal $T$, the fraction of episodes that survive with positive progress along the command and with less than 1.5 times the median progress of the same policy's flat-ground survivors at that gravity and command. Because the SONIC planner assumes a vertical axis, we test SONIC on slopes only on a \SI{10}{\degree} inclined mesh.
% Source: friction and traversal, review_2026-10-24/summary.md sec. C1: foot friction static = dynamic = 0.5 and ground 0.3 in all 2014 eval_best logs that print it, max-combine (g1_unitree/velocity_env_cfg.py:96-104); tan 26 = 0.488, tan 30 = 0.577. T per policy, seed, and cell in slope_table.md and slope_table_perseed.csv (bound 1.5x the FlatRigid survivor median of the same policy, seed, profile, and gravity).
% Source: LuPNT-private projects/guillemc/RAL_Humanoid/run_best_eval_queue.py:42-53 (grades, azimuth 180 = Up, 0 = Down, Flat cell) and eval_locomotion.py:566-610 (tilted gravity, zero spawn yaw, upright spawn); g1_best/env_cfg.py:80 (26 deg training).

\subsection{Soil Models}
\label{sec:rft}
% Source: sections/3-method.tex "Soil Models" (lines 109, 112, 119), "Training Setup" (lines 55, 60), "Evaluation Protocol" (line 135), S1-method-details.tex (lines 100-108, 178, 182), tables/tab_soilrft.tex, tables/tab_route.tex. Engines: training and 3D RFT in Isaac Lab on Newton with MuJoCo Warp, MPM in Newton, SCM and CRM in Chrono (3-method.tex Soil Models). Friction is the test foot friction per engine (3-method.tex line 109). Episodes per cell are per seed: rigid and 3D RFT 256 per seed, SCM 32 per seed and ground, MPM and CRM one per seed and forward profile, 6 per policy and cell (3-method.tex line 135). MPM covers the 8 grid policies at training gravity plus Earth+Moon-Rough-RFT and Earth2Moon-Rough-RFT at the Moon (tab_route).
\begin{table}[!tb]
  \centering\scriptsize
  \caption{Ground models of the study: 3D resistive force theory (3D RFT), material point method (MPM), soil contact model (SCM), and continuous representation model (CRM). The held-out engines use lunar regolith at both gravities}
  \label{tab:soil_models}
  \renewcommand{\arraystretch}{1.1}
  \setlength{\tabcolsep}{2pt}
  \begin{tabularx}{\linewidth}{|l|l|c|>{\raggedright\arraybackslash}X|c|}
    \hline
    \hdrstrut \textbf{Model} & \textbf{Use} & \textbf{$\mu_f$} & \textbf{Soil parameters} & \textbf{Ep.} \\
    \hline\hline
    Rigid & Train, test & 0.5 & None & 256 \\
    3D RFT~\cite{agarwal_mechanistic_2023} & Train, test & 0.6 & Training: Earth $\rho_c$ 500 to 3000, $\mu$ 0.1 to 1.0, Moon $\rho_c$ 150 to 8000, $\mu$ 0.9. Test: nominal lunar $\rho_c$ about 5200, $\mu$ 0.9, Earth sand $\rho_c$ 1500, $\mu$ 0.55. \SI{0.12}{\metre} layer & 256 \\
    \hline
    MPM~\cite{sulsky_particle_1994,newton_contributors_2025} & Held out & 0.84 & $\rho$ 1500, $\phi$ \SI{42}{\degree}, $c$ \SI{0.52}{\kilo\pascal}, \SI{0.12}{\metre} bed & 2 \\
    SCM~\cite{tasora_overview_2017} & Held out & 0.5 & Bekker, $\phi$ \SI{35}{\degree}, $c$ \SI{0.17}{\kilo\pascal}, $k_c$ \SI{1.4}{\kilo\newton\per\metre\squared}, $k_\phi$ \SI{820}{\kilo\newton\per\metre\cubed}, $n=1$ & 32 \\
    CRM~\cite{unjhawala_physicsbased_2025} & Held out & 0.5 & As MPM, \SI{0.15}{\metre} flat bed only & 2 \\
    \hline
  \end{tabularx}
  \tabnote{Rigid ground, 3D RFT, and MPM run in Newton, and SCM and CRM in Chrono. $\mu_f$ is the test foot friction, $\rho_c$ the effective 3D RFT density and $\rho$ the bulk density in \si{\kilo\gram\per\metre\cubed}, $\mu$ the internal friction, $\phi$ the friction angle, and $c$ the cohesion. Rigid ground and 3D RFT test every policy at both gravities, and the held-out engines test the eight grid policies at their training gravity, with MPM also run for the rough strategies at the Moon. Ep. is episodes per cell and seed, over the eight profiles in rigid ground, 3D RFT, and SCM, and one per forward profile 2 and 3 in MPM and CRM.}
\end{table}

% Method: 3D RFT as the cheapest test soil, MPM, SCM, and CRM as held-out models of increasing fidelity
The test grounds are rigid, 3D RFT, and three held-out engines on which no policy trains (\Cref{tab:soil_models}). Because 3D RFT runs in the training engine, it covers every policy and geometry, but it is in distribution only for the RFT-trained policies. MPM uses a Drucker--Prager yield surface in Newton~\cite{sulsky_particle_1994,newton_contributors_2025}. Chrono provides the other two: SCM, a Bekker-type soil contact model~\cite{tasora_overview_2017}, and CRM, a smoothed-particle-hydrodynamics continuum called the continuous representation model~\cite{unjhawala_physicsbased_2025}.

% Method: engines share a regolith source but not one material, MPM stiffness uncalibrated, cohesion absent from 3D RFT
The engines take their lunar regolith from one source~\cite{carrier_physical_1991} but do not model one identical material. In MPM, for example, the bed stiffness follows from the incompressibility constraint and the yield surface rather than from the bootprint modulus of \SI{8}{\kilo\pascal\per\centi\metre} that fixes the 3D RFT soils. Cohesion and footprint history are absent from 3D RFT, so MPM adds contact effects that training never saw.
% Source: MPM treats the bed as incompressible Drucker-Prager with yield pressure and Young modulus 1e15 and no plate calibration (mpm_g1best/mpm_walk_g1best.py:412, TIMING.md:12, PLAN.md item 8). The 8 kPa/cm modulus enters only the calibrated 3D RFT soils (S1).
% Note (2026-10-19): the old MPM Earth sand was replaced by the lunar regolith at Earth gravity, because in a static plate test it sank 5.5/10.3 cm at 12/24 kPa against 1.2/2.3 cm for the 3D RFT Earth soil with the same parameters.

% Method: 3D RFT test soils at each gravity, effective densities fitted to one policy's gait
At the Moon we test loose, crater-rim, and nominal soils of 1346, about 2900, and about \SI{5200}{\kilo\gram\per\metre\cubed}, fitted so that the walking foot sinkage of \MoonFlatRFT matches Apollo bootprints~\cite{carrier_physical_1991}. These densities are effective parameters of the 3D RFT force, not physical ones, and the nominal value is about 3.5 times the \SI{1500}{\kilo\gram\per\metre\cubed} bulk density of the regolith. Lunar soil results are therefore conditional on a calibration tied to the gait of one policy, and the fitted densities are point estimates. The compacted soil of \SI{16000}{\kilo\gram\per\metre\cubed}, twice as stiff as the stiffest lunar training soil, is a stress test and not a calibrated soil. Every other 3D RFT soil test uses the nominal lunar soil at both gravities, matching the regolith of the held-out engines. At Earth gravity, however, this soil has no physical counterpart because its effective density was fitted at lunar gravity. Since it is also stiffer than any Earth training soil, we call these cells lunar-calibrated soil at Earth gravity. Only the Earth soil cells of \Cref{tab:gravity} use an Earth medium sand (\SI{1500}{\kilo\gram\per\metre\cubed}, internal friction 0.55) as in GM-Loco, which we call Earth sand.
% Source: lunar soils docs/regolith_calibration/insim_calibration.md table and Aditya-progress-2026-10-16.txt sec. 3 (rho_c 1346 / ~2900 / ~5200 / 16000 held out, targets 17.8 / 8.1 / 4.5 mm; old values 4000 and 8000 measured 5.3 and 2.9 mm); Earth soil g1_best/env_cfg.py:301 (EARTH_TEST_SOIL rho_c 1500, mu_int 0.55, mu_surf 0.6; the medium sand of scripts/gm_loco/eval_policy.py:64); Moon mu_surf 0.6 g1_best/env_cfg.py:302-305 (the surface friction the in-sim calibration ran at).
% Source: sinkage sweep v4 (2026-10-18), shared paper_results/evals/sinkage_sweep/v4_final, power-law fit depth ~ rho_c^-1.0; label fit on Moon-Flat-RFT, mean of seeds 42-44 (v4_final/README.md: Moon-Rough-RFT loads a pitched foot and is not used for the labels).
% Note (2026-10-19 protocol): the nominal lunar soil (rho_c 5200, mu_int 0.9) is used by MPM, SCM, CRM and the 3D RFT column of tab:soil_results at both gravities (the Earth-gravity "Regolith" cell). EARTH_TEST_SOIL is used only by the 3D RFT soil columns of tab:gravity.

% Method: the soil models span different gravity scalings of regolith strength, exact for 3D RFT and approximate for MPM
The soil models differ in how soil strength scales with gravity: linearly in 3D RFT by construction~\cite{agarwal_mechanistic_2023}, with the overburden $\rho g z$ but not linearly in MPM and CRM, and not at all in SCM, whose moduli do not depend on $g$~\cite{tasora_overview_2017}. Measurements on granular media fall between these cases (\Cref{sec:related_soil}), so we claim held-out lunar-soil effects only where their sign holds in both MPM and SCM. However, the Earth-sand results and survival on lunar-calibrated soil at Earth gravity come from 3D RFT alone.

\subsection{Evaluation Protocol and Metrics}
\label{sec:protocol}
% Setup: both gravities, eight profiles, horizon, test conditions, fall rule
We test every policy at Earth and lunar gravity in episodes that each command one of eight profiles: forward at 0.25, 0.5, and \SI{1.0}{\metre\per\second} (profiles 1 to 3), backward at \SI{0.5}{\metre\per\second} (4), lateral at $\pm\SI{0.3}{\metre\per\second}$ (5), forward at \SI{0.5}{\metre\per\second} while turning at $\pm\SI{0.5}{\radian\per\second}$ (6), and two \SI{4}{\second} segments stepping from 0.25 to \SI{1.0}{\metre\per\second} (7) or from 1.0 to \SI{0}{\metre\per\second} (8). An episode lasts $T_{\mathrm{ep}}=\SI{10}{\second}$, \SI{2}{\second} of standing followed by \SI{8}{\second} of the profile, and its commands are body-frame velocities with no outer loop on heading. Tests remove the pushes and the mass, center-of-mass, and gain randomization, and fix foot friction at 0.5 on rigid ground (soil friction in \Cref{tab:soil_models}). Tiles, starts, commands, and seeds are shared across policies. An episode fails when the pelvis drops below \SI{0.2}{\metre} above the local ground or the base tilts beyond \SI{0.8}{\radian}.
% Source: profiles LuPNT-private projects/guillemc/RAL_Humanoid/command_profiles.py _SEGMENTS (1-3 fwd 0.25/0.5/1.0, 4 vx -0.5, 5 vy +-0.3, 6 vx 0.5 with wz +-0.5, 7 4 s at 0.25 then 4 s at 1.0, 8 4 s at 1.0 then 4 s at 0; +- = halves of the episodes) and S1 command profiles.
% Source: Eval variant g1_best/env_cfg.py:261-283 (no push, mass, CoM, gains; foot friction 0.5; obs corruption off) and eval_locomotion.py:373-378 (every Best task run through its Eval variant); terminations g1_best/env_cfg.py:223-243.

% Setup: test cells and episodes per engine
Rigid ground is tested on a flat bed, the four tile families, and the slopes. Each tile cell uses the 32 held-out tiles of one family without gravity tilt, and each robot spawns above the local ground. Every soil engine is tested on the flat bed. 3D RFT and SCM are also tested on beds that follow the 32 held-out combined tiles, and MPM on only two of these tiles, one per forward profile. Every 3D RFT bed is a \SI{0.12}{\metre} layer, the deepest seen in training. On the MPM combined beds, we exclude any episode with more than \SI{5}{\percent} of its frames off the bed, together with its rigid pair, an exclusion that depends on the outcome. For the eight grid policies at their training gravity, however, keeping these episodes changes no seed-mean MPM ratio by more than 0.01.
% Source: test cells g1_best/env_cfg.py:293-355 (held-out tiles seeds 2000-2031, no tilt asserted :333, all-soil columns and 0.12 m layer :336-345, fixed soil :346-351); above-ground spawn g1_best/mdp.py:61-103 (highest ground within 0.3 m); 256 episodes run_best_eval_queue.py:76; 256 envs over 32 tile columns = 8 starts per tile (g1_study/terrain.py:72-95).

% Definition: survival, signed progress, soil progress ratio
Survival $S$ is the fraction of episodes that reach $T_{\mathrm{ep}}$ without failure. Let $\hat{\bm c}$ be the unit base-frame direction of the first nonzero translational command of a profile. Signed progress $d_\parallel=\int_{2\,\mathrm{s}}^{T_{\mathrm{ep}}}\bm v_{xy}\cdot\hat{\bm c}\,dt$ projects the base-frame planar velocity $\bm v_{xy}$ on $\hat{\bm c}$ over the motion phase, so heading drift does not count. This phase includes the hold of profile 8 and ends at a fall. The soil progress ratio $\rho_v$, averaged over seeds, divides the mean signed progress on soil over profiles 2 and 3 (all eight profiles in SCM) by that of the same engine's rigid run on the same geometry and starts. Values above 1 mean more progress on soil, but a rigid-run fall can also raise the ratio, so we report survival with it.

% Note (2026-10-26): the four combined-tile cells were re-run with per-episode tile ids and a paired per-seed tile-cluster bootstrap computed (output/RAL_Humanoid/review_2026-10-26/c11_tiles/c11_summary.md, c11_signrule_cluster.csv, script_snapshot/); design effect median 1.12 (0.47-2.32), none of the 27 comparisons on these cells changes verdict. The Newcombe interval remains the stated rule. Rule applied by LuPNT-private projects/guillemc/RAL_Humanoid/figures/sign_rule_table.py, output tables/tab_signrule.tex and tab_signrule_withdrawn.tex (54 comparisons, 42 stated and 12 withdrawn or narrowed since 2026-10-27, the 42 including the 13 rigid-control rows added 2026-10-27; 44 pass and 10 fail after the 26 and 30 deg slope cells were scored on traversal T, 2026-10-24; 32/9 on survival before the control rows).
% Method: approximate Wilson and Newcombe intervals per seed, sign agreement over seeds as a replication heuristic, re-run spread
Each rate has a \SI{95}{\percent} Wilson interval per cell and seed~\cite{brown_interval_2001}, and a difference in survival has the per-seed Newcombe interval built from the two Wilson intervals~\cite{newcombe_interval_1998}. Although the 256 episodes of a cell share 32 tiles, a tile-cluster bootstrap on the combined-tile cells changes no verdict. Under our sign rule, a survival difference is consistent if it has the same sign in all three seeds and each seed's interval excludes zero, and we report only consistent differences as findings. However, the sign rule is a seed-replication heuristic rather than a test with a controlled error rate. If seed effects are symmetric about zero, a null difference passes the rule with probability at most 1/4, so with no multiplicity control the 54 comparisons we scored, including those we do not state, would give at most about 14 chance passes if every difference were null. Because simulation on graphics processors is not bitwise deterministic, re-running every 3D RFT soil evaluation changes seed-mean survival by at most 0.035 and no sign-rule verdict.
% Source: run-to-run spread, eval_best_gaitfix (18 soil cells x 36 policy-seeds = 648, 2026-10-20) vs eval_best: 313 changed, max |change| 0.0625 per seed (Earth-Flat-Rigid_s43 SoilFlatCompacted_Moon), seed-mean cell values max |change| 0.0352 (review_2026-10-24/summary.md sec. C18); sign_rule_table.py on the merged set gives identical pass/fail for all 41 claims and the rough-vs-soil table. Binomial SE sqrt(0.25/256) = 0.031. Chance pass bound: P(three seeds share a sign) = 2 x (1/2)^3 = 1/4, 54/4 = 13.5 (41/4 = 10.25 before the 13 rigid-control rows).

\section{Results}
\label{sec:results}

\subsection{Gravity Transfer}
\label{sec:res_gravity}
% Source: output/RAL_Humanoid/eval_best/<policy>_s<seed>__<cell>.json, tabular generated by projects/guillemc/RAL_Humanoid/figures/make_tab_keycells.py
\begin{table}[!tb]
  \centering\scriptsize
  \caption{Survival on the key test cells at both gravities, with Earth sand in the soil columns at Earth and the nominal lunar soil at the Moon. RFT in policy names is 3D resistive force theory (3D RFT).}
  \label{tab:gravity}
  \setlength{\tabcolsep}{2pt}
  % Source: tables/tab_keycells.tex (written by projects/guillemc/RAL_Humanoid/figures/make_tab_keycells.py), seed means of the Flat, Combined tiles, and 20 deg columns copied without the seed ranges and without the 30 deg columns; the six all-zero Earth rows of the Moon-trained and fine-tuned policies collapsed into one; SONIC rows from tables/tab_terrain.tex (Flat and Comb. survival, 256 episodes per cell). The JointRoughRigid rows (rigid control) are read from eval_c3_joint_rough_rigid. The full table is tab:keycells-full in the supplement.
\begin{tabular*}{\linewidth}{@{\extracolsep{\fill}}|l|cc|cc|cc|}
\hline
& \multicolumn{2}{c|}{\textbf{Flat}} & \multicolumn{2}{c|}{\textbf{Comb. tiles}} & \multicolumn{2}{c|}{\textbf{Slope \SI{20}{\degree}}} \\
\hdrstrut \textbf{Policy} & Rigid & Soil & Rigid & Soil & Up & Down \\
\hline\hline
\multicolumn{7}{|l|}{\textbf{At Earth}} \\ \hline
\EarthFlatRigid & 1.00 & 0.96 & 0.53 & 0.27 & 0.36 & 0.19 \\
\EarthFlatRFT & 1.00 & 1.00 & 0.70 & 0.69 & 0.08 & 0.26 \\
\EarthRoughRigid & 1.00 & 0.33 & 0.98 & 0.30 & 1.00 & 0.68 \\
\EarthRoughRFT & 1.00 & 0.99 & 0.98 & 0.87 & 0.99 & 0.98 \\
\hline
Moon-trained, fine-tuned (6) & 0.00 & 0.00 & 0.00 & 0.00 & 0.00 & 0.00 \\
\hline
\JointFlatRigid & 0.92 & 0.00 & 0.29 & 0.00 & 0.17 & 0.13 \\
\JointRoughRFT & 0.99 & 0.98 & 0.95 & 0.84 & 1.00 & 0.86 \\
\JointRoughRigid & 0.99 & 0.13 & 0.96 & 0.10 & 0.98 & 0.82 \\
\hline
SONIC & 1.00 & -- & 0.90 & -- & -- & -- \\
\hline
\multicolumn{7}{|l|}{\textbf{At Moon}} \\ \hline
\EarthFlatRigid & 0.64 & 0.92 & 0.45 & 0.80 & 0.08 & 0.01 \\
\EarthFlatRFT & 0.67 & 0.96 & 0.43 & 0.76 & 0.05 & 0.11 \\
\EarthRoughRigid & 0.74 & 0.96 & 0.43 & 0.77 & 0.37 & 0.21 \\
\EarthRoughRFT & 0.75 & 0.84 & 0.54 & 0.78 & 0.56 & 0.16 \\
\hline
\MoonFlatRigid & 1.00 & 1.00 & 0.80 & 0.96 & 0.26 & 0.47 \\
\MoonFlatRFT & 1.00 & 1.00 & 0.74 & 0.96 & 0.26 & 0.44 \\
\MoonRoughRigid & 1.00 & 1.00 & 0.99 & 0.98 & 1.00 & 0.67 \\
\MoonRoughRFT & 0.95 & 1.00 & 0.97 & 1.00 & 0.81 & 0.99 \\
\hline
\FtFlatRigid & 1.00 & 1.00 & 0.84 & 0.96 & 0.31 & 0.67 \\
\FtRoughRFT & 1.00 & 1.00 & 0.98 & 1.00 & 0.95 & 0.99 \\
\JointFlatRigid & 1.00 & 1.00 & 0.53 & 0.86 & 0.06 & 0.17 \\
\JointRoughRFT & 1.00 & 1.00 & 0.99 & 1.00 & 0.99 & 0.96 \\
\JointRoughRigid & 1.00 & 1.00 & 0.99 & 0.99 & 1.00 & 0.90 \\
\hline
SONIC & 0.10 & -- & 0.08 & -- & -- & -- \\
\hline
\end{tabular*}

  \tabnote{Each value is the fraction of 256 episodes per seed without a fall, averaged over three seeds. Soil cells are 3D RFT under a \SI{0.12}{\metre} layer. Comb. tiles are the 32 held-out combined tiles, and slopes are rigid. SONIC is one checkpoint, tested on rigid ground only.}
\end{table}

% Source: soil columns, SoilFlat_Earth / SoilTilesCombined_Earth use EARTH_TEST_SOIL (rho_c 1500, mu_int 0.55); SoilFlat_Moon / SoilTilesCombined_Moon use the nominal lunar soil (rho_c 5200, mu_int 0.9), equal to the Nominal columns of tab_lunarsoils.

% Result: lunar gravity gives every policy group flight phases and toe-first landing, while no policy is airborne at Earth
At the Moon, every policy group has flight phases, with flight fractions up to 0.66, and lands toe-first, as Klein and Roberts report for their G1 policies~\cite{klein_training_2025}, whereas at Earth no policy, including SONIC, is airborne.
% Source: fig_froude.json group means: duty factor at Earth 0.61-0.66 (Earth-trained), 0.59-0.87 (SONIC); at the Moon 0.24-0.32 (Moon-trained), 0.26-0.35 (fine-tuned); flight fraction means up to 0.66 (Earth-trained at the Moon at 1 m/s), 0.52 Moon-trained, 0.29 gravity-randomized; 0 for every group at Earth.

% Result: training at one gravity transfers asymmetrically, and only gravity randomization with rough RFT training holds at both gravities up to 20 degrees
Policies trained at one gravity transfer asymmetrically to the other (\Cref{tab:gravity}). The Earth-trained policies keep 0.64 to 0.75 survival at the Moon on flat rigid ground over the \SI{10}{\second} horizon, while the rough-trained Earth policies lose more on the combined tiles and the \SI{20}{\degree} slopes. However, these values are upper bounds for a longer horizon, because the Earth-trained policies keep falling until the end, with 30 to \SI{46}{\percent} of falls in the last \SI{2}{\second}. At Earth, in contrast, the Moon-trained policies and the fine-tunes fall in every episode, nearly always during the standing phase, at per-seed median times of about 1.0 to \SI{1.7}{\second}. For five of these six policies the ankles reach their pitch and torque limits while knee and hip torque stay below theirs, a failure sequence consistent with ankle overload in the learned crouched posture rather than a lack of knee strength. At the Moon SONIC survives at most 0.10 on every rigid test cell, below every Earth-trained grid policy (\Cref{tab:gravity}). Only the gravity-randomized \JointRoughRFT keeps a seed-mean survival above 0.83 at Earth and 0.95 at the Moon on every key cell up to \SI{20}{\degree}, with lowest seeds of 0.74 and 0.91. Still, at Earth it trails \EarthRoughRFT on the \SI{20}{\degree} downhill grade, with 0.86 against 0.98, but the sign rule does not license this loss. At the Moon, likewise, no lunar-trained policy consistently outsurvives \JointRoughRFT on these cells. The rigid control \JointRoughRigid shows no difference from \JointRoughRFT that the sign rule licenses on any rigid key cell or on the 3D RFT key cells at the Moon, although the control falls below the thresholds on the \SI{20}{\degree} downhill grade, with 0.82 at Earth and 0.90 at the Moon. At Earth, however, it survives only 0.13 on flat sand and 0.10 on the sand tiles, against 0.98 and 0.84 for \JointRoughRFT, a difference the sign rule licenses. Only the survival of \JointRoughRFT on soil at Earth gravity therefore depends measurably on soil training, most on Earth sand in 3D RFT, the training contact model.

\subsection{Terrain and Slopes}
\label{sec:res_terrain}

% Result: rough training at the training gravity, and how it compares with soil training
At the training gravity, rough-terrain training holds every tile family (per-seed values in the data attachment), and it raises the survival of the rigid-trained policies on the combined tiles at both gravities (\Cref{tab:gravity}). On these rigid tiles and the \SI{20}{\degree} uphill grade, rough training gains more survival than soil training in every seed at both gravities. On the combined tiles under Earth sand, in contrast, soil training gains more in the seed means, although the sign rule is not met. Because rough training includes tilted gravity (\Cref{sec:policies}), the slope gain belongs to the rough-training package rather than the tile geometry.

% Result: the rough-training gain does not carry across gravity on tiles or on the lunar 20-degree grade, and SONIC at Earth
However, on the tiles the rough-training gain does not carry across gravity. At the Moon, for example, \EarthRoughRigid loses survival on every family, with no consistent difference from \EarthFlatRigid on the combined tiles (per-seed differences of $-0.18$ to $+0.21$). Similarly, on the \SI{20}{\degree} uphill grade at the Moon, the rough-versus-flat difference of the Earth-trained policies is not consistent across seeds, and the survivors of \EarthRoughRigid slide back, with per-seed median progress of $-0.8$ to \SI{-4.7}{\metre}. By comparison, at Earth SONIC survives 1.00 on flat rigid ground and 0.90 on the combined tiles (\Cref{tab:gravity}), as well as 1.00 on the crater and rock tiles and on the inclined mesh.
% Source: SONIC at Earth, tables/tab_terrain.tex SONIC Earth row (Flat 1.00, Crat. 1.00, Rock 1.00, Und. 0.91, Comb. 0.90, mesh Up 1.00, Down 1.00) and data/terrain.csv; Table II (tab_keycells_main) carries only Flat and Comb. for SONIC.

\begin{figure*}[!tb]
  \centering
  \includegraphics[width=\textwidth]{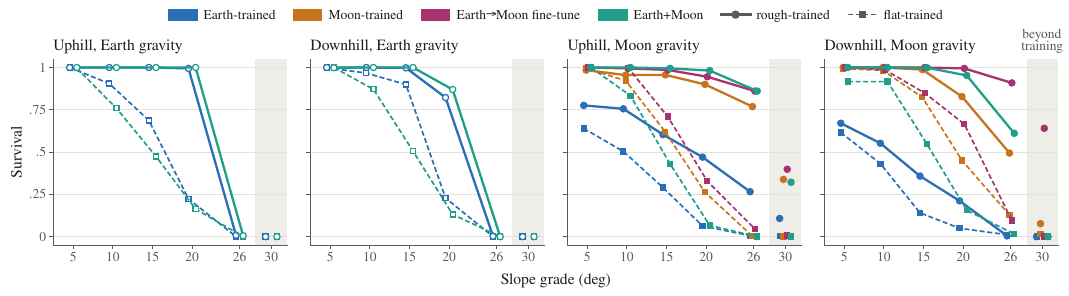}
  \caption{\textbf{Slope envelope at both gravities.} Survival against uphill and downhill grade for rough-trained (solid) and flat-trained (dashed) policy groups, with gravity tilted over a borderless rigid plane. Values are averaged over seeds and, where a group has both, over its policies trained on rigid ground and on resistive force theory (RFT) soil among the 12 of \Cref{sec:policies}, excluding the rigid control. The shaded \SI{30}{\degree} column lies beyond the training range and the \SI{26.6}{\degree} friction-cone limit. At \SI{26}{\degree} most survivors slide, so survival is not traversal.}
  \label{fig:slope}
\end{figure*}
\label{sec:res_slope}
% Result: rough-trained policies hold slopes up to 20 degrees, and from 26 degrees the survivors mostly slide
Up to \SI{20}{\degree}, the rough-trained policies walk the grade at their training gravity, whereas from \SI{26}{\degree} the survivors mostly slide (\Cref{fig:slope}). In particular, at Earth the rough-trained policies survive the \SI{20}{\degree} grade more often in every direction and pair, except \EarthRoughRigid downhill, where one seed mostly falls, and at the Moon all survivors of \JointRoughRFT and \FtRoughRFT on the \SI{20}{\degree} uphill grade progress forward. From \SI{26}{\degree} every policy fails at Earth, with seed-mean survival of at most 0.01, whereas rough-trained policies that saw lunar gravity keep seed-mean survival of 0.64 to 0.96 on the \SI{26}{\degree} uphill grade at the Moon. However, 0.59 to 0.64 of their survivors move backward, and only 0.14 to 0.29 of their episodes traverse the grade. At \SI{30}{\degree}, beyond the training range, traversal is at most 0.07 for each of the 12 policies at both gravities and at most 0.13 for the rigid control. The 0.64 downhill survival of \FtRoughRFT at the Moon comes from survivors covering 2.8 to 3.5 times the commanded distance and so slide. Beyond \SI{20}{\degree} we claim only failures.

\subsection{Soil Training}
\label{sec:res_soil}
\begin{table}[!tb]
  \centering\footnotesize
  \caption{Soil progress ratio $\rho_v$ of the grid policies at their training gravity in each engine: 3D resistive force theory (3D RFT, written RFT in policy names), material point method (MPM), soil contact model (SCM), and continuous representation model (CRM)}
  \label{tab:soil_results}
  \setlength{\tabcolsep}{3pt}
  \begin{tabular*}{\linewidth}{@{\extracolsep{\fill}}|l|cc|cc|cc|c|}
\hline
& \multicolumn{2}{c|}{\textbf{3D RFT}} & \multicolumn{2}{c|}{\textbf{MPM}} & \multicolumn{2}{c|}{\textbf{SCM}} & \textbf{CRM} \\
\hdrstrut \textbf{Policy} & Flat & Comb. & Flat & Comb. & Flat & Comb. & Flat \\
\hline\hline
\EarthFlatRigid & 1.04 & 0.80 & 0.76 & 1.26 & 0.96 & 1.37 & 0.98 \\
\EarthFlatRFT & 1.07 & 0.91 & 0.86 & 1.24 & 1.05 & 1.18 & 1.07 \\
\EarthRoughRigid & 1.05 & 0.93 & 1.03 & 1.04 & 1.06 & 0.91 & 1.03 \\
\EarthRoughRFT & 0.94 & 0.93 & 1.06 & 1.02 & 1.03 & 0.95 & 1.07 \\
\hline
\MoonFlatRigid & 1.04 & 1.35 & 1.00 & 1.05 & 1.04 & 1.01 & 1.09 \\
\MoonFlatRFT & 1.01 & 1.46 & 1.00 & 1.49 & 0.96 & 1.05 & 0.97 \\
\MoonRoughRigid & 0.89 & 0.87 & 0.92 & 0.92 & 1.19 & 0.92 & 1.13 \\
\MoonRoughRFT & 1.14 & 1.08 & 1.01 & 1.03 & 1.05 & 0.99 & 1.01 \\
\hline
\end{tabular*}

  \tabnote{Earth rows are at Earth gravity on the lunar-calibrated soil of \Cref{sec:rft}. $\rho_v$ is the seed-mean soil progress ratio of \Cref{sec:protocol}, including progress before a fall. Combined (Comb.) beds follow the held-out combined tiles.}
\end{table}
% Note (2026-10-19): the four-soil survival table (tab:lunar_soils) moved to the supplement, sections/S2-results-details.tex.

% Result: soil training pays on Earth sand, on lunar soil helps only the flat pair on tiles at Earth, rho_v near 1 reflects speed for the lunar policies, and lunar stiffness hardly matters for them

In 3D RFT, which is in distribution for the RFT-trained policies only, soil training helps on Earth sand (\Cref{tab:gravity}): both RFT-trained Earth policies survive the combined tiles under sand more often than their rigid-trained neighbors. On flat sand, in particular, \EarthRoughRigid survives only 0.33 and \JointFlatRigid none. Nearly all falls of \EarthRoughRigid and \JointFlatRigid are orientation failures, at per-seed median times of 1.9 to \SI{5.7}{\second}. Because both survive the stiffer lunar-calibrated soil at Earth gravity, these falls are plausibly artifacts of meeting the per-foot 3D RFT wrench for the first time, although we did not log force or sinkage traces to confirm this. However, on the lunar-calibrated soil at Earth gravity, outside the Earth training range, the gain shrinks: the rough pair shows no consistent survival difference on the combined tiles, and only the flat pair keeps a survival gain there.
% Result: at the Moon soil training changes survival little, rho_v near 1 reflects speed, and soil stiffness hardly matters to the lunar-trained policies
At the Moon, in contrast, soil training changes survival little: \MoonFlatRFT and \MoonRoughRFT are within a few hundredths of their rigid-trained neighbors on every soil cell, and \EarthRoughRFT shows no consistent survival difference from \EarthRoughRigid on flat soil. Because the Moon-trained policies survive nearly all soil episodes, their $\rho_v$ near 1 reflects speed (\Cref{tab:soil_results}), while the high 3D RFT values on the combined tiles come mainly from falls on the rigid reference (\Cref{tab:gravity}). Across the three calibrated lunar soils and the compacted stress-test soil, the Moon-trained and fine-tuned policies keep at least 0.93 of tile episodes and nearly all flat ones.
% Source: flat-sand falls, review_2026-10-24/summary.md sec. C15: EarthMoon-Flat-Rigid SoilFlat_Earth S 0.00 in all seeds, 256/256/256 falls, bad_orientation 256/256/247, median 2.18/1.88/2.04 s, falls during stand 61/157/125; SoilFlatRegolith_Earth S 0.88/0.99/1.00. Earth-Rough-Rigid SoilFlat_Earth S 0.54/0.11/0.34, all bad_orientation, median 5.66/3.86/3.14 s.
% Source: lunar soil at Earth gravity, eval_best SoilFlatRegolith_Earth and SoilTilesCombinedRegolith_Earth success_rate, 256 episodes per seed, mean over seeds 42-44: flat 0.99-1.00 for all four Earth policies; combined tiles Earth-Flat-Rigid 0.38, Earth-Flat-RFT 0.68, Earth-Rough-Rigid 0.98, Earth-Rough-RFT 0.98. Four lunar soils: tab_lunarsoils (Moon-only and fine-tuned rows, tile minimum 0.93 for Moon-Flat-Rigid on loose soil).

\subsection{Training Strategies for the Moon}
\label{sec:res_route}

% Result: the three lunar strategies survive at the Moon on flat rigid ground, only gravity randomization keeps Earth walking, the fine-tune costs 15k iterations, and the three rough strategies are close at the Moon
At the Moon, the three flat rigid policies \MoonFlatRigid, \FtFlatRigid, and \JointFlatRigid survive every flat rigid episode (\Cref{tab:gravity}). At Earth, however, only \JointFlatRigid walks (0.92), while the fine-tuned policies forget Earth walking, as expected in fine-tuning~\cite{wolczyk_finetuning_2024}. Moreover, the fine-tune costs 15k iterations including its parent, against 10k for either from-scratch policy. On the rigid combined tiles at the Moon, the three rough RFT strategies survive within a few hundredths of each other, but only \JointRoughRFT also walks at Earth, at the 10k-iteration budget of \MoonRoughRFT.

\subsection{Dependence on the Soil Model}
\label{sec:res_soilmodel}
% \begin{figure}[!tb]
%   \centering
%   \fbox{\parbox[c][0.26\columnwidth][c]{0.93\columnwidth}{\centering\todo{static plate sinkage against pressure for 3D RFT, MPM, SCM, and CRM at Earth and lunar gravity, fixed and gravity-scaled load, with the Apollo bootprint reference}}}
%   \caption{\textbf{Static plate sinkage at both gravities.} A plate of the G1 sole size is loaded from 1 to \SI{10}{\kilo\pascal} in each soil model.}
%   \label{fig:plate}
% \end{figure}
% Note (2026-10-19): the footprint figure (fig:footprints) moved to the supplement, commented out until it is regenerated without the withdrawn Earth sand.

% Method: the soil models scale resistance with gravity differently, so only MPM-and-SCM agreement (the agreement rule, descriptive, no intervals) licenses a lunar-soil claim
Following \Cref{sec:rft}, we claim a held-out lunar-soil effect only where the difference has the same sign in all three seeds in both MPM and SCM (our descriptive agreement rule).

% Result: in the sampled held-out rollouts no loss of progress is detected for the Moon-trained policies, the seeds resolve only large losses and the MPM combined beds are inconclusive, at the low G1 foot pressure
In the three held-out engines we detect no loss of progress on soil for the Moon-trained policies (\Cref{tab:soil_results}). Their seed-mean $\rho_v$ is 0.92 to 1.19 in every MPM, SCM, and CRM cell except \MoonFlatRFT on the MPM combined bed, and they survive all but one MPM combined-bed episode. The exception is one seed that fell on the rigid tile but not on soil, which shortens its rigid reference. However, MPM and CRM run at most 6 episodes per cell, and three seeds resolve only large losses. Per-policy \SI{95}{\percent} Student $t$ intervals over the three seeds therefore remain compatible with progress losses of about \SI{30}{\percent} in some held-out cells, as their lower ends reach 0.69 for \MoonRoughRigid. The MPM combined-bed result is also inconclusive (intervals in the data attachment). Apart from the exception, mean progress per episode differs between soil and rigid ground by $-0.38$ to \SI{0.56}{\metre}. No gain of \MoonFlatRFT or \MoonRoughRFT over their rigid-trained neighbors meets the agreement rule. The result applies only to a light humanoid like the G1, at its low foot pressure: standing on one 0.20 by \SI{0.07}{\metre} sole, the \SI{33.3}{\kilo\gram} robot exerts about \SI{3.9}{\kilo\pascal} at the Moon and \SI{23}{\kilo\pascal} at Earth.
% Source: 95% half-width and progress differences, review_2026-10-24/summary.md sec. C16 (half-width t(0.975,2)/sqrt(3) = 2.48 x pooled between-seed SD, an approximate t half-width and not a power-based minimum detectable effect, reproduced by make_tab_keycells.py write_rhov_intervals, Moon-trained (4): 3D RFT flat 0.370, comb 0.326; MPM flat 0.146, comb 1.118; SCM flat 0.241, comb 0.128; CRM flat 0.126). Delta progress of the four Moon-trained policies in MPM, SCM, CRM -0.38 to +0.56 m excluding Moon-Flat-RFT MPM comb (+1.03); soil survival 0.83 to 1.00 (SCM comb 0.85 for both flat Moon policies, MPM comb Moon-Flat-Rigid 0.83).
% Source: per-policy intervals, tables/tab_rhov_intervals.tex and data/rhov_intervals.csv (make_tab_keycells.py write_rhov_intervals, mean +- t(0.975,2) s/sqrt(3)) and S2 sec. tab:rhov_intervals: Moon-trained lower ends in MPM, SCM, CRM reach 0.69 (Moon-Rough-Rigid, flat MPM bed and combined SCM bed); Moon-Flat-RFT MPM comb. interval [-0.71, 3.69]; pooled MPM comb. half-width 1.12.
% Source: comb_summary.{md,json}, Moon-Flat-RFT seed 43 2.51 (fell on the rigid tile, survived on MPM), seeds 42/44 1.03/0.93.

% Caveat: at lunar gravity 3D RFT raises survival over rigid ground mainly for Earth-trained policies, cause not isolated, so soil claims rest on the held-out engines
By comparison, 3D RFT at lunar gravity makes walking easier than rigid ground for most policies (\Cref{tab:gravity}), and on both grounds their falls are orientation falls. The effect is largest for the Earth-trained policies, which meet lunar gravity outside training, and the rigid-trained ones among them also meet 3D RFT outside training. Across the four soil stiffnesses, the Earth-trained policies gain up to 0.45 in seed-mean survival over the rigid cell and do better on softer soil, with 0.99 on loose flat soil against 0.87 on compacted soil for \EarthFlatRigid. In contrast, the rough-trained policies that saw lunar gravity survive within 0.05 of their rigid-ground values on 3D RFT at the Moon. Our lunar-soil claims therefore rest on the held-out engines.
% Source: sign rule (tables/tab_signrule_withdrawn.tex, rows G6): flat soil vs flat rigid fails for Earth-Rough-RFT (+.05 interval incl. 0, +.06, +.11) and for Moon-Flat-Rigid and Moon-Flat-RFT (1.00 on both grounds); combined-tile soil passes for all six. Earth sand: Earth-Flat-RFT vs Earth-Flat-Rigid flat sand +.08, +.03, +.01 (seed 44 interval incl. 0); Earth-Rough-Rigid flat-sand falls all orientation, median about 3.9 s, Earth-Rough-RFT 5 falls in total (coordinator audit 2026-10-19).
% Source: largest gain +0.45 Earth-Flat-Rigid SoilTilesCombinedLoose_Moon vs TilesRigidCombined_Moon (review_2026-10-24/summary.md sec. C15 list); rough lunar-capable policies (Moon-Rough-Rigid, Moon-Rough-RFT, Earth2Moon-Rough-RFT, EarthMoon-Rough-RFT) flat 1.00/1.00, 0.95/1.00, 1.00/1.00, 1.00/1.00 and tiles 0.99/0.98, 0.97/1.00, 0.98/1.00, 0.99/1.00 rigid/soil in tab_keycells.
% Source: tab_keycells at the Moon, Earth-Flat-Rigid flat 0.64 rigid vs 0.92 soil, tiles 0.45 vs 0.80; Earth-Rough-Rigid 0.74 vs 0.96, 0.43 vs 0.77; Moon-Flat-Rigid tiles 0.80 vs 0.96; Moon-Flat-RFT tiles 0.74 vs 0.96. Fall cause: all falls are orientation (tilt) terminations on both grounds (verified by the coordinator, 2026-10-19).

% Result: at Earth gravity on lunar soil no held-out soil-training difference in progress meets the rule
At Earth gravity on lunar regolith, MPM and SCM disagree for the rough pair, so no soil-training difference in progress meets the agreement rule there.

\section{Discussion and Conclusion}
\label{sec:conclusion}
% Takeaway: gravity transfer is asymmetric, rough training does not carry across gravity, soil training pays mainly on Earth sand, held-out soils detect no loss but resolve only large ones, and gravity randomization with rough 3D RFT training is the tested package that works at both gravities
Transfer between the two gravities is asymmetric. At the Moon, Earth-trained policies keep 0.64 to 0.75 survival on flat rigid ground over \SI{10}{\second}, while SONIC fails. Yet falls continue, so longer episodes would see lower survival. Conversely, at Earth the Moon-trained and fine-tuned policies fall in every episode, nearly always in the standing phase. Rough training helps on uneven rigid ground at the training gravity but not across gravity. Soil training helps on Earth sand in 3D RFT, whereas on lunar soil the held-out models detect no loss of progress for the Moon-trained policies, although our seeds resolve only large losses. The lunar soil behaves close to rigid partly by construction, since the 3D RFT soils are fitted to Apollo bootprint stiffness and the SCM Bekker set lies close to it. Fine-tuning for the Moon adds half a from-scratch run, and with no Earth episodes, the fine-tuned policies forget how to walk at Earth. Gravity randomization with rough and 3D RFT training, in contrast, costs one from-scratch run and keeps a seed-mean survival above 0.83 at Earth and 0.95 at the Moon on every key cell up to \SI{20}{\degree}. Compared with the best single-gravity policy, the randomized policy shows no consistent loss at either gravity (\Cref{sec:protocol}), although on the \SI{20}{\degree} Earth downhill grade it is lower in two of three seeds. A rigid control with the same gravity randomization and rough training shows no consistent difference from the recommended policy on rigid ground or on lunar soil, but it falls in most episodes on Earth sand in 3D RFT, so soil training is what keeps the recommended policy walking there. Among the tested strategies, we therefore recommend combining gravity randomization, rough training, and 3D RFT training with these reward settings, in simulation with fixed friction and without observation noise, latency, calibrated actuator dynamics, or test pushes.
% Source: rigid control, output/RAL_Humanoid/eval_c3_joint_rough_rigid/c3_summary.md (Earth flat sand 0.13 vs 0.98, sand tiles 0.10 vs 0.84, both licensed; no licensed difference on rigid cells or at the Moon).
% Source: compute, tab_route Iter. 10k vs 15k and Steps 0.98 vs 1.47 billion (1.47/0.98 = 1.5).

% Limitation: geometry, soil inputs, held-out engine coverage, and the spread of contact-model gravity scalings
Terrain statistics are extrapolated below their measured sizes and omit rocks under \SI{0.2}{\metre}. Lunar soil parameters combine sites, depths, and test methods, and how regolith strength scales with gravity is uncertain~\cite{chen_lowgravity_2026}, so we claim lunar soil results only where MPM and SCM agree. In the held-out engines, CRM is tested on flat beds only, the MPM bed stiffness is not calibrated to the bootprint modulus, and the articulation exported to Chrono is not checked dynamically against the training engine.

% Note (2026-10-19): the sentence on 3D RFT stiffness under tilted gravity (overstated by 1/cos(theta) - 1, 6.4% at 20 deg and 15.5% at 30 deg) was removed, because every reported slope test is rigid.

% Limitation: simulation only, single robot, single generalist checkpoint, and the next measurement
All results are simulation predictions for the light \SI{33.3}{\kilo\gram} G1, and SONIC is one released checkpoint that does not represent generalist controllers. A robot of \SI{70}{\kilo\gram} or more with similar feet would roughly double the foot pressure, so we expect deeper sinkage and a more nonlinear, yielding soil response that our soil models capture only partly, and soil training would likely matter more there, including at lunar gravity.
% Source: G1 model mass 33.3 kg, S1-method-details.tex:95 (Unitree spec about 35 kg); one-foot pressure 33.3 kg x 9.81 m/s^2 / (0.20 x 0.07 m^2) = 23 kPa at Earth, stated in 5b-granular.tex. Heavier-robot part is a scaling argument (pressure proportional to mass at equal foot area), not a simulation result. The measurement that would most reduce the soil uncertainty is soil resistance at reduced gravity under a foot-sized load, which no ground test yet provides.
% Scope: release of code, checkpoints, and evaluation records
Code, trained checkpoints, and per-episode evaluation records will be released upon acceptance.

\section*{Acknowledgment}
This manuscript benefited from the use of AI-based assistants, specifically Anthropic's Claude models, which were used for coding assistance, evaluation tooling, and manuscript drafting. All final content was reviewed and edited by the authors.
\appendices
\renewcommand{\thetable}{S\arabic{table}}
\renewcommand{\thefigure}{S\arabic{figure}}
\setcounter{table}{0}
\setcounter{figure}{0}
\section*{Supplementary Material}
This supplement follows the order of the main text, and three parts of it carry most of the evidence behind its claims. The seed sign rule is applied to every stated survival comparison in Tables~\ref{tab:signrule-supp} and~\ref{tab:signrule-c3}, while Table~\ref{tab:signrule-withdrawn} lists the tested comparisons that the text narrowed or withdrew (Section~\ref{sec:signrule-supp}). The rigid control of Section~\ref{sec:rigidcontrol-supp} isolates the effect of soil training in the recommended policy, and Table~\ref{tab:rhov_intervals} gives the soil progress ratio of each policy with its own interval over the three seeds. The per-seed and per-profile tables behind the seed means are CSV files in the folder \texttt{supplement\_tables} released with the code, whose README gives the experiment, cells, units, and episode counts of each file, and the text names the file wherever it summarizes one.
\section{Study Design Details}
\label{sec:method-supp}
% Scope: what this supplement adds to the study design of the main text
This section gives the training, terrain, soil-model, and evaluation details that the Method and Experimental Setup sections of the main text summarize.

\subsection{Training Recipe}
\label{sec:recipe-supp}
% Scope: the 12 trained policies, the rigid control, and their names
Table~\ref{tab:policies} lists the 12 trained policies and the rigid control \JointRoughRigid of \JointRoughRFT. Each name gives the training gravity, the training geometry, and the training ground, Earth$\to$Moon marks a fine-tune of an Earth policy at lunar gravity, and Earth+Moon marks gravity randomization.

\begin{table}[!tb]
  \centering\footnotesize
  \caption{\textbf{Policies trained in this study.} RFT in the Ground column and in policy names is 3D resistive force theory (3D RFT).}
  \label{tab:policies}
  \renewcommand{\arraystretch}{1.1}
  \setlength{\tabcolsep}{2.3pt}
  \begin{tabular*}{\linewidth}{@{\extracolsep{\fill}}|l|c|c|c|l|}
    \hline
    \hdrstrut \textbf{Name} & \textbf{$g$ [m/s$^2$]} & \textbf{Geometry} & \textbf{Ground} & \textbf{Init.} \\
    \hline\hline
    \EarthFlatRigid & 9.81 & Flat & Rigid & Scratch \\
    \EarthFlatRFT & 9.81 & Flat & RFT & Scratch \\
    \EarthRoughRigid & 9.81 & Rough & Rigid & Scratch \\
    \EarthRoughRFT & 9.81 & Rough & RFT & Scratch \\
    \hline
    \MoonFlatRigid & 1.62 & Flat & Rigid & Scratch \\
    \MoonFlatRFT & 1.62 & Flat & RFT & Scratch \\
    \MoonRoughRigid & 1.62 & Rough & Rigid & Scratch \\
    \MoonRoughRFT & 1.62 & Rough & RFT & Scratch \\
    \hline
    \FtFlatRigid & 1.62 & Flat & Rigid & Earth parent \\
    \FtRoughRFT & 1.62 & Rough & RFT & Earth parent \\
    \hline
    \JointFlatRigid & 1.62--9.81 & Flat & Rigid & Scratch \\
    \JointRoughRFT & 1.62--9.81 & Rough & RFT & Scratch \\
    \hline
    \JointRoughRigid & 1.62--9.81 & Rough & Rigid & Scratch \\
    \hline
  \end{tabular*}
  \tabnote{Each policy has three seeds, and all are frozen at test. Init. is the initialization, from scratch or from an Earth parent. Each from-scratch policy trains for 10k iterations of proximal policy optimization (PPO). RFT-trained policies train with half of the environments on 3D RFT and half rigid. Each fine-tune continues the final checkpoint of its Earth parent, \EarthFlatRigid or \EarthRoughRFT, one per parent seed, for 5k more iterations. The rigid control \JointRoughRigid differs from \JointRoughRFT only in its rigid ground. It trained on a different GPU host.}
\end{table}

% Method: observations, rewards, and curricula of the GM-Loco recipe
The actor observes ten frames of base angular velocity, gravity direction, joint positions and velocities, and previous action, together with the current velocity command. The critic also observes base orientation and linear velocity, the height of each foot above the local ground, its contact state, contact force, and air time, and the soil parameters, none of which the actor sees. Rewards cover velocity tracking, foot air time, foot clearance, orientation, spacing, slip, and landing impact, base height and orientation, joint limits, and energy. The penalties on joint torque, joint acceleration, action magnitude and rate, and joint speed near its limit start at a tenth of their weight and reach it after 1500 iterations, and the velocity-tracking reward narrows and doubles in weight until iteration 7500. On terrain the height terms are measured above the local ground.
% Source: actor/critic groups gm_loco/agents/rsl_rl_ppo_cfg.py:24 and gm_loco/env_cfg/observation_cfg.py:68-263,301-310 (history 10); ramp gm_loco/step2b_env_cfg.py:97,130-138 (1500 it x 24 steps = 36000 steps); tracking std 1.0->0.316 and weight 4->8 to step 180000 = 7500 it (CURRICULUM_SCALE 0.5, g1_best/env_cfg.py:52, confirmed in the run dump params/env.yaml of Moon-Rough-RFT_s43).

% Method: table of network, PPO, and reward settings, and what differs between the gravity configurations
Table~\ref{tab:training} lists the settings that all 36 training runs share and the few that depend on the gravity configuration. The Earth policies use gravity 9.81, the Moon policies 1.62, and the Earth+Moon policies draw the gravity of each episode log-uniformly between the two. This range is far wider than the gravity perturbations used for robustness in humanoid controllers such as ULTRA, $\pm\SI{0.1}{\metre\per\second\squared}$~\cite{he_ultra_2026}. The landing-impact weight is divided by the gravity ratio at the Moon and kept at its Earth value for Earth+Moon, and the force threshold of the undesired-contact penalty and of the critic's foot contact force is multiplied by the gravity ratio at the Moon and for Earth+Moon, so for Earth+Moon it stays at the lunar value in every episode. The thresholds of the foot-contact detection scale with the gravity of each episode. On rough ground, a third of the episodes tilt gravity by up to \SI{26}{\degree}. The soil ranges of the RFT-trained policies are those of the table, and rigid-ground policies have no soil.
% Source: ckpt_best/<run>/params/agent.yaml (all 36 identical but seed, run name, and max_iterations 10001, or 5001 for the Earth-to-Moon fine-tunes; learn() runs max_iterations iterations, saved checkpoints at 500-iteration intervals) and env.yaml (rewards, curriculum, events.gravity, soil events, num_envs 4096, decimation 4, 20 s episodes) of Earth-Rough-RFT_s42, Moon-Rough-RFT_s42, EarthMoon-Rough-RFT_s42, Earth2Moon-Rough-RFT_s42; gravity draw is log-uniform in g1_study/mdp.py:23,63-64; scaling of thresholds and landing weight g1_best/env_cfg.py:122-129. Saved params agree with the source except the dump of Earth-Flat-RFT_s42 (slope_tilt_max_deg 20 with tilt fraction 0, so unused) and the dumped tracking std 0.707 and tracking weight 4, which are the constructor values that the curriculum overwrites from the first step.

\begin{table*}[!tb]
  \centering\scriptsize
  \caption{\textbf{Training configuration.} Reward weights are final values. Tracking terms narrow their kernel width from 1.0 to 0.316 and, for linear velocity and heading, double their weight, linearly over the first 7500 iterations. Terms marked with an asterisk start at a tenth of their weight and reach it linearly after 1500 iterations. Soil values are ranges drawn per episode. Network and PPO rows use proximal policy optimization (PPO), a multilayer perceptron (MLP), exponential linear unit (ELU) activations, and a target Kullback-Leibler divergence (KL), with coefficient (coef.), standard deviation (std), and environment (env.).}
  \label{tab:training}
  \renewcommand{\arraystretch}{1.1}
  \setlength{\tabcolsep}{3pt}
  \begin{tabular*}{\linewidth}{@{\extracolsep{\fill}}|l|r|l|r|l|r|}
    \hline
    \multicolumn{6}{|l|}{\hdrstrut \textbf{Network and PPO}} \\
    \hline
    Actor and critic & \multicolumn{5}{l|}{MLP 512-256-128, ELU, no observation normalization, initial action std 1.0} \\
    Learning rate & $10^{-3}$, adaptive (target KL 0.01) & Clip, entropy coef. & 0.2, 0.005 & $\gamma$, $\lambda$ & 0.99, 0.95 \\
    Epochs, minibatches & 5, 4 & Steps per env., envs & 24, 4096 & Gradient norm clip & 1.0 \\
    Optimizer & Adam & Value loss & clipped, coef. 1.0 & Augmentation & mirror symmetry \\
    Iterations & \multicolumn{5}{l|}{10000 from scratch, 5000 more for each fine-tune} \\
    \hline\hline
    \multicolumn{6}{|l|}{\hdrstrut \textbf{Reward weights}} \\
    \hline
    Linear velocity tracking & 8 & Heading tracking & 8 & Yaw rate tracking & 1 \\
    Termination & $-200$ & Foot air time & 1 & Foot clearance & 5 \\
    No flight & $-1$ & Foot slip & $-0.5$ & Foot spacing & $-2$ \\
    Foot roll & $-1$ & Foot pitch & $-1$ & Posture deviation & $-1$ \\
    Base height & $-10$ & Base orientation & $-10$ & Undesired contacts & $-10$ \\
    Vertical velocity & $-0.2$ & Roll-pitch rate & $-0.05$ & Joint limits & $-2$ \\
    Energy & $-10^{-3}$ & Joint speed & $-2\times10^{-4}$ & Landing impact & see below \\
    Action rate, lower body* & $-0.05$ & Action rate, upper body* & $-0.05$ & Action magnitude* & $-0.01$ \\
    Joint acceleration* & $-10^{-6}$ & Joint torque* & $-10^{-5}$ & Joint speed limit* & $-1$ \\
    \hline\hline
    \multicolumn{6}{|l|}{\hdrstrut \textbf{Gravity configurations}} \\
    \hline
    & \multicolumn{1}{c|}{\textbf{Earth}} & \multicolumn{2}{c|}{\textbf{Moon}} & \multicolumn{2}{c|}{\textbf{Earth+Moon}} \\
    \hline
    $g$ [m/s$^2$] & 9.81 & \multicolumn{2}{c|}{1.62} & \multicolumn{2}{c|}{1.62 to 9.81, log-uniform} \\
    Landing impact weight & $-0.005$ & \multicolumn{2}{c|}{$-0.030$} & \multicolumn{2}{c|}{$-0.005$} \\
    Undesired-contact threshold [N] & 1.0 & \multicolumn{2}{c|}{0.165} & \multicolumn{2}{c|}{0.165} \\
    Soil density $\rho_c$ [kg/m$^3$] & 500 to 3000 & \multicolumn{2}{c|}{150 to 8000} & \multicolumn{2}{c|}{150 to 8000} \\
    Soil friction, internal & 0.1 to 1.0 & \multicolumn{2}{c|}{0.9} & \multicolumn{2}{c|}{0.1 to 1.0} \\
    Soil friction, surface & 0.1 to 1.0 & \multicolumn{2}{c|}{0.343 to 1.0} & \multicolumn{2}{c|}{0.1 to 1.0} \\
    \hline
  \end{tabular*}
\end{table*}

% Method: robot model, actions, and foot geometry
We use Unitree's 29-degree-of-freedom G1 model, \SI{33.3}{\kilo\gram}, with joint-space PD control at a policy rate of \SI{50}{\hertz}, actions scaled by 0.25 about the default pose, and joint targets clipped to $\pm\SI{2}{\radian}$. The robot keeps Unitree's stock foot collision geometry, a sole box with corner spheres, and the 3D RFT element grid, 5 by 5, spans its sole footprint of 0.20 by \SI{0.07}{\metre} from the sphere contact points.
% Asset: Hugging Face unitreerobotics/unitree_model 323e3502, g1_29dof_rev_1_0.usd sha256 cfed730b...; gains per actuator group in LuPNT-isaac g1_unitree/robot_cfg.py:114-172
% Source: action scale 0.25 and clip +-2 rad on the absolute target, step2b_env_cfg.py:124 and env.yaml actions.joint_pos; PPO clip_actions null (the old 13.6 clip is gone); foot box g1_study/env_cfg.py:64.

% Setup: simulator, actuator, and soil-engine settings for reproduction
Table~\ref{tab:sim_settings} lists the simulator, actuator, and soil-engine settings needed to reproduce the training and evaluation runs.

\begin{table*}[!tb]
  \centering\scriptsize
  \caption{\textbf{Simulation settings.} Values used by all 36 training runs and by the evaluations in each engine. Gains, limits, and armature are Unitree's G1 configuration, unchanged. Each engine also runs its own rigid-ground baseline with the same robot and control period. Actuators use proportional-derivative (PD) control. The held-out soil engines are the material point method (MPM), soil contact model (SCM), and continuous representation model (CRM), and 3D resistive force theory (3D RFT) is the training soil model.}
  \label{tab:sim_settings}
  \setlength{\tabcolsep}{3pt}
  \begin{tabular*}{\linewidth}{@{\extracolsep{\fill}}|l|l|}
    \hline
    \multicolumn{2}{|l|}{\hdrstrut \textbf{Training and 3D RFT evaluation}} \\
    \hline
    Physics step, substeps & \SI{5}{\milli\second}, 2 \\
    Control rate & \SI{50}{\hertz} (decimation 4) \\
    Solver & MuJoCo Warp, Newton method, 100 iterations, \\
    & 50 line-search iterations, implicit-fast integrator, pyramidal cone \\
    Ground and foot friction & 0.3, 0.3 to 1.0 in training, 0.5 in tests \\
    \hline\hline
    \multicolumn{2}{|l|}{\hdrstrut \textbf{Actuators (PD stiffness [N\,m/rad], damping [N\,m\,s/rad])}} \\
    \hline
    Hip pitch, roll, yaw & 100, 2 \\
    Knee & 150, 4 \\
    Ankle pitch and roll & 40, 2 \\
    Waist yaw & 200, 5 \\
    Waist roll and pitch & 40, 5 \\
    Shoulder, elbow, wrist & 40, 1 \\
    Torque limit [N\,m] & hip pitch and yaw, waist yaw 88, hip roll and knee 139, \\
    & ankle, waist roll and pitch, shoulder, elbow, wrist roll 25, wrist pitch and yaw 5 \\
    Velocity limit [rad/s] & hip pitch and yaw, waist yaw 32, hip roll and knee 20, \\
    & ankle, waist roll and pitch, shoulder, elbow, wrist roll 37, wrist pitch and yaw 22 \\
    Armature [kg\,m$^2$] & 0.01 on every joint \\
    \hline\hline
    \multicolumn{2}{|l|}{\hdrstrut \textbf{Held-out soil engines}} \\
    \hline
    MPM & \SI{1.25}{\milli\second} step (\SI{800}{\hertz}), decimation 16, \SI{2}{\centi\metre} voxels, \\
    & 2 particles per voxel, \SI{0.12}{\metre} bed, \SI{1.2}{\metre} by \SI{9}{\metre} lane \\
    SCM & \SI{0.5}{\milli\second} step, \SI{2}{\centi\metre} grid \\
    CRM & \SI{0.5}{\milli\second} multibody step, \SI{0.1}{\milli\second} fluid step, \\
    & \SI{2}{\centi\metre} particle spacing, \SI{0.15}{\metre} bed \\
    \hline\hline
    \multicolumn{2}{|l|}{\hdrstrut \textbf{Software}} \\
    \hline
    Training and 3D RFT & Isaac Lab 3.0, Newton --, MuJoCo Warp -- \\
    SCM and CRM & Chrono -- \\
    \hline
  \end{tabular*}
\end{table*}
% Source: ckpt_best/*/params/env.yaml, all 36 runs identical: sim.dt 0.005, physics.num_substeps 2, solver_cfg (mujoco_warp, solver newton, iterations 100, ls_iterations 50, integrator implicitfast, cone pyramidal, njmax/nconmax 300), decimation 4, ground friction 0.3; actuators N7520-14.3 / N7520-22.5 / N5020-16 / W4010-25 from LuPNT-isaac g1_unitree/robot_cfg.py:115-171 (unitree_rl_lab 4960b847 unitree.py:420-478); eval JSON summary sim_dt 0.005, decimation 4, physics NewtonCfg. Note: gm_loco/robot.py (ankle 50 N m, armature-derived gains) is NOT the robot of these runs. MPM: collab_results/mpm_g1best{,_comb} run JSONs, physics_hz 800, num_substeps 1, decimation 16, mpm.voxel 0.02, ppc 2, depth 0.12, lane 1.2 x 9.0 m, max_iterations 1000 or 3000 (both occur). SCM/CRM: collab_results/chrono_g1best/evals/*.json args dt 0.0005, scm_delta 0.02, crm_spacing 0.02, crm_cfd_dt 1e-4, crm_depth 0.15, mu 0.5.
% Method: software versions and training hosts
All policies train and evaluate with the same Isaac Lab 3.0.0 early-access release (commit ae37b02), which pins Isaac Sim 6.1.0, Newton 1.5.2, MuJoCo Warp 3.11.0, Warp 1.16.0, PyTorch 2.11.0 with CUDA 12.8, and rsl-rl 5.4.1, while SCM and CRM run in PyChrono 10.0.0. Training ran on several GPU machines, and because GPU simulation is not bitwise deterministic across devices, we compare policies by seed means and seed spreads rather than by matched seeds.
% Source: 2026-10-27, read-only on ws4: /home/shared_ws4/lunar_humanoid_paper (maintainer's shared results). Package versions from the training env site-packages (LuPNT-isaac-ea/.pixi/envs/default, installed 2026-10-07, before the 2026-10-12 start): isaaclab 17.0.2, isaacsim 6.1.0.0, newton 1.5.2, mujoco_warp 3.11.0, warp_lang 1.16.0, torch 2.11.0+cu128, rsl_rl_lib 5.4.1. Run git diffs (laptop runs): IsaacLab-ea ae37b028e clean, LuPNT-isaac 8e822d013 (ea-port). Hosts from tfevents names: laptop sri-aditya-deevi-Raider-GE78-HX-14VIG (flat policies, Earth2Moon-Flat-Rigid, EarthMoon-Flat-Rigid), navlab-ws4 RTX 4090 driver 580.178.04 (rough policies, Earth2Moon-Rough-RFT s42/s43, EarthMoon-Rough-RFT s42/s43), Sherlock sh04 nodes (Earth-Flat-RFT s44, Earth-Flat-Rigid s44, Earth2Moon-Rough-RFT s44, EarthMoon-Rough-RFT s44; Sherlock logs: Warp 1.16.0, torch 2.11.0+cu128, Kit 110.3.0, CUDA toolkit 12.9). Rigid control: ws6 RTX 5090, IsaacLab-ea same versions, LuPNT-isaac 12608bc. PyChrono 10.0.0 from LuPNT-private .pixi/envs/chrono conda-meta. Verification level: ws4 env read directly; Sherlock logs print Warp and torch versions; laptop runs record only the IsaacLab commit ae37b028e (same lock), env not read.

% Method: settings taken from the study design rather than from GM-Loco
The robot model and actuators, the ground friction, and the commands come from the study design rather than from GM-Loco. A ground friction of 1.0 would make the foot-friction randomization inert, because the solver uses the larger of the two friction coefficients. Commands are drawn uniformly from $-0.5$ to \SI{1.0}{\metre\per\second} forward, $\pm\SI{0.3}{\metre\per\second}$ laterally, and $\pm\SI{1.0}{\radian\per\second}$ in yaw rate, resampled every \SI{10}{\second} in \SI{20}{\second} episodes, with 20\% zero commands and no command curriculum. The fixed test profiles are never used in training. Robot randomization follows GM-Loco, with torso mass $\pm\SI{5}{\kilo\gram}$, torso center of mass $\pm\SI{3}{\centi\metre}$, and actuator gains 0.9 to 1.1. Every 10 to \SI{15}{\second} a push sets a velocity change of up to \SI{1}{\metre\per\second} along each horizontal axis at Earth gravity, scaled by $\sqrt{g/9.81}$ for each environment, or \SI{0.41}{\metre\per\second} on the Moon, so that the push has the same Froude number at both gravities.
% Source: mass/CoM gm_loco/step2b_env_cfg.py:71-77; gains gm_loco/env_cfg/event_cfg.py:89-101; push interval and range event_cfg.py:153-158, sqrt(g/9.81) scaling g1_best/mdp.py:39-58 (sqrt(1.62/9.81)=0.406).

% Method: 3D RFT force corrections and contact detection
The 3D RFT force is smoothed by GM-Loco's moving-average filter. The two corrections to the force apply in every RFT cell. Applying the soil wrench only on the first solver substep, with its torque about the link origin rather than the foot's center of mass where the solver applies it, lets a standing robot sink until the soil force reaches 1.8 times its weight at Earth and 2.5 times at the Moon, against 1.08 times with both corrections. Depth and surface normal come from the local soil surface. RFT was fitted on level beds and is extrapolated on inclined surfaces. The contact-dependent rewards and critic observations count a foot as in contact when the contact sensor force exceeds \SI{1}{\newton} or the 3D RFT force on that foot exceeds \SI{40}{\newton}, both scaled by $g/9.81$, because 3D RFT forces bypass the contact sensor.
% Source: substep re-apply and COM torque gm_loco/soft_contact/actions_cfg.py:65-77 and gm_loco/newton_wrench_fix.py:1-20 (without it a standing robot sinks until the RFT force is 1.8x Earth / 2.5x Moon its weight, 1.08x with it); union contact gm_loco/mdp/contact_state.py:12-13,38,57.

% Method: lunar soil range, ranges in detail, why walking sinkage and not a plate test
Earth training conditions express density as a packing ratio of 0.5 to 1.0 times a bulk density of 1000 to \SI{3000}{\kilo\gram\per\metre\cubed}. For the lunar range the robot walks on each candidate soil at lunar gravity, and its measured foot sinkage is matched to the bootprint depth that the Apollo stiffness gives at the G1 foot pressure~\cite{carrier_physical_1991}. A static plate test is not enough, because 3D RFT has no static bearing capacity, and a sliding foot receives only about \SI{14}{\percent} of the vertical force of a foot pushed straight down. Lunar surface friction is 0.34 to 1.0, and the lunar density range spans a factor of 53 in stiffness. Its stiff edge, \SI{8000}{\kilo\gram\per\metre\cubed}, is stiffer than the nominal bootprint soil.
% Source: g1_best/env_cfg.py:190-211 (rho_c = packing x bulk; Moon packing 1.0); in-sim mapping rho_c -> walking sinkage at the 3562 Pa one-foot pressure (0.80 MPa/m target 4.5 mm), sinkage sweep v4 (2026-10-18); measured with the old values, 4000 gave 5.3 mm and 8000 gave 2.9 mm. Calibration method and the 14% sliding-foot force: insim_calibration.md and Aditya-progress-2026-10-16.txt sec. 3.

\subsection{Rough Training and Budgets}
\label{sec:policies-supp}
% Method: tile families, slope tilt, and promotion rule of rough training
Only the rigid-trained and RFT-trained pairs differ in a single training setting. The Earth and Moon RFT policies also differ in soil range, and rough training is a package of tiles, curriculum, and gravity tilt, so those contrasts compare training packages. The rough-training gain at a gravity is the rough-trained policy against the flat-trained policy, for example \EarthRoughRigid against \EarthFlatRigid, and the soil-training gain is the RFT-trained policy against the rigid-trained policy, for example \MoonFlatRFT against \MoonFlatRigid. Rough training uses 10 tiles from each of the four tested tile families, crater-only, rock-only, undulation-only, and combined. The gravity tilt of rough training is drawn per episode, uniformly up to \SI{26}{\degree}, toward a uniformly random direction. The promotion rule moves a robot up after walking past the tile border and down after covering less than half the commanded distance~\cite{rudin_learning_2022}. Compounding of rough and soil training is read by comparing the rough RFT policy, for example \MoonRoughRFT, with its two single-factor neighbors.

% Method: warm starting, and budget arithmetic of the fine-tunes
Warm starting can generalize differently from scratch training at equal training loss~\cite{ash_warmstarting_2020}, which is why the fine-tunes are compared with the Moon-trained policies \MoonFlatRigid and \MoonRoughRFT. A fine-tune sees half the lunar iterations of a Moon-trained policy but 1.5 times its total, about $1.5\times10^{9}$ environment steps against $9.8\times10^{8}$.

% Method: what a fine-tune restores from the Earth checkpoint, and how the schedules continue
A fine-tune restores the actor, the critic, the Adam state, and the iteration counter from the final Earth checkpoint, so it continues at iteration 10000 with the learning rate the adaptive schedule had reached, and it keeps the standard PPO settings of Table~\ref{tab:training}. The iteration-dependent schedules do not restart. Their milestones are shifted back by 10000 iterations, so the tracking schedule and the regularization ramps are already at their final values on the first step. The terrain curriculum is driven by success rather than by iteration and starts again at its first level.
% Source: ckpt_best/2026-10-13_17-43-29_Earth2Moon-Rough-RFT_s42/params/agent.yaml (load_checkpoint .../model_10000.pt, max_iterations 5001) and env.yaml curriculum (tracking step_0 -240000, step_1 -60000, ramps -240000 to -204000, in env steps of 24 per iteration, and every ramp_* term likewise); resume path isaaclab_rl/entrypoints/backends/train_rsl_rl.py:230 calls runner.load(path) with no load_cfg, rsl_rl on_policy_runner.py:145-160 loads all of actor, critic, optimizer, iteration (algorithms/ppo.py:372-394, defaults all True), and learn() continues from start_it = current_learning_iteration (on_policy_runner.py:77); shift gm_loco/step2b_env_cfg.py:132-138 and budget_env_cfg.py:58 via GM_LOCO_RESUME_ITERATION=10000 (g1_best/__init__.py:8); checkpoint model_10000.pt holds iter 10000 and optimizer lr 1.14e-4 (Earth-Rough-RFT_s42), model_15000.pt of the fine-tune holds iter 15000; terrain curriculum starts at max_init_terrain_level 0, g1_best/env_cfg.py (terrain.max_init_terrain_level = 0), level state is not in the checkpoint.

\subsection{SONIC Port}
\label{sec:sonic-supp}
% Method: released checkpoint and planner inputs
The released G1 SONIC checkpoint~\cite{luo_sonic_2026,sonic_modelcard_2026} serves as a frozen external reference. Its kinematic planner maps a motion mode, directions, and a target speed to whole-body references with no gravity or terrain input~\cite{sonic_planner_2026}, so its references stay Earth-timed and level. SONIC is tested on rigid ground only, on the eight command profiles mapped to the planner inputs.
% SONIC: HF nvidia/GEAR-SONIC 6733128a, planner_sonic.onnx sha256 39b553e1..., model_encoder 013ab028..., model_decoder c7241a12... (downloaded; hashes checked against the HF LFS oids)

% Method: planner timing, port acceptance criteria, variant
The planner maps a motion mode, movement and facing directions, and a target speed to whole-body references at \SI{30}{\hertz}. Its context comes from its own previous plan rather than the measured state. We run the gains, armatures, and torque limits of the released training configuration, which match the released deployment, and reimplement the planner loop in Python. The planner output and the first tracker observation match the reference implementation exactly. We also do not fine-tune the tracker because the training code of the planner is documented as to be released later~\cite{sonic_planner_2026}. SONIC runs in the same Isaac Lab~3.0 environment as the trained policies, with its own action interface and default pose. The port must reproduce the planner output for a fixed input sequence, the first tracker observation, and the Earth survival and speed of the released MuJoCo deployment, with survival of at least 0.95 and a speed fraction within 0.10 of the deployment on every command profile. On flat Earth ground the port survives all 256 episodes, with speed ratios within 0.09 of the deployment. We use the default variant of the release, whose tracker runs at \SI{50}{\hertz} and whose planner replans every \SI{1}{\second} and whenever the commanded mode, speed, or direction changes, with joint mapping by name and the release's default pose.

% Method: command mapping and reference feasibility check
For SONIC, the commanded speed sets the planner's target speed and mode, slow walking up to \SI{0.8}{\metre\per\second} and walking above, as in the release's gamepad mapping, the commanded direction sets its movement direction, and the integral of the commanded yaw rate sets its facing direction, refreshed at each replan with a \SI{0.5}{\second} lead. The \SI{10}{\degree} grade of the inclined mesh was fixed in advance of the SONIC tests.

\subsection{Gravity-Dependent Choices}
\label{sec:lunar-supp}
% Method: human walk-run transition and continuous-contact walking limit
Humans change from walking to running near $\mathrm{Fr}=0.5$ at 0.4\,g and above, rising toward 0.9 to 1.1 at lunar-like gravity~\cite{kram1997effect,carr2009apollo}, and the 0.36 of suited Apollo crews has a \SI{95}{\percent} interval of $\pm0.11$~\cite{carr2009apollo}. Walking with continuous contact is possible up to $\mathrm{Fr}=1$, or \SI{1.12}{\metre\per\second} on the Moon, so every commanded speed can be walked at both gravities.
% Source: sqrt(1.62*0.78)=1.12.

% Method: air-time target relative to pendulum time, reward scaling variant
At each touchdown the air-time reward pays the swing time beyond \SI{0.5}{\second} and charges swings shorter than that, and the flight penalty applies at every command in the training range. The \SI{0.5}{\second} swing target is 0.72 of the pendulum time $\sqrt{L/g}=\SI{0.69}{\second}$ on the Moon and 1.77 of \SI{0.28}{\second} on Earth. Arm et al. multiply the joint-power weight by $g/g_E$ and the squared-torque weight by $(g/g_E)^2$~\cite{arm_efficient_2025}. The contact-detection thresholds scale as fractions of body weight. Local-weight and Earth-weight cost of transport differ by $g_E/g_L=6.06$, with Earth and lunar gravity $g_E$ and $g_L$, and local weight is primary.
% Source: sqrt(0.78/1.62)=0.694, sqrt(0.78/9.81)=0.282, 0.5/0.694=0.72, 0.5/0.282=1.77.

% Comparison: gravity scaling of energy and torque weights against Arm et al.
Unlike Arm et al., who scale them with gravity and, for tests on their offload rig, feed the gravity-compensation torques of the legs forward to the actuators~\cite{arm_efficient_2025}, we keep the energy and torque weights equal at both gravities and add no feedforward torque, so at the Moon the same weights act on smaller torques and powers.

\subsection{Terrain}
\label{sec:terrain-supp}
% Method: crater and rock statistics
Hartmann's distribution, extrapolated well below the crater sizes it was derived from, predicts \num{11.7} craters of 0.4 to \SI{2.4}{\metre} per \SI{100}{\metre\squared}~\cite{hartmann_saturation_1984}. Forty percent of the craters are fresh, with depth-to-diameter ratios in the \num{0.08} to \num{0.215} range measured on young lunar craters~\cite{hoover_depth_2024}. The rest are degraded, with ratios from \num{0.04} to \num{0.10} that we assume, for a mean of \num{0.10}. Rim height is \num{0.184} times depth, the ratio of Pike's relations $0.036D$ and $0.196D$ for fresh craters~\cite{pike_sizedependence_1977}. Rock height-to-diameter ratios follow those measured at the Chang'E-3 site~\cite{di_rock_2016}.

% Method: rock size-frequency model and abundance
Rocks follow the model of Golombek and Rapp for Mars and Earth analog sites~\cite{golombek_sizefrequency_1997}, applied to the Moon as by Di et al.~\cite{di_rock_2016}, with abundance $k=0.02$, above the \num{0.0125} they measured at the Chang'E-3 site.

% Method: undulation surface details
Undulations use a Hurst exponent of 0.9. Their amplitude is set on a large field so that the median absolute slope at a \SI{4}{\metre} baseline equals \SI{4.66}{\degree}, the median measured for lunar highlands from Lunar Reconnaissance Orbiter Camera Narrow Angle Camera (LROC NAC) elevation models~\cite{cai2020meter}. The surface is high-passed at the tile scale, because longer wavelengths are covered by the slope test. Rosenburg et al. report a Hurst exponent of 0.95 for highlands from the Lunar Orbiter Laser Altimeter (LOLA) over baselines of about \SI{17}{\metre} to \SI{2.7}{\kilo\metre}~\cite{rosenburg2011slopes}.

% Result: measured slope and rock statistics of the held-out tiles
The held-out tiles are therefore smoother than the cited highlands. Because each tile has its best-fit plane removed, the median absolute slope at a \SI{4}{\metre} baseline on the 32 held-out undulation tiles is \SI{1.6}{\degree}, or \SI{1.9}{\degree} when both ends of a baseline keep \SI{1.5}{\metre} from the tile edge, with \SI{0.9}{\degree} to \SI{5.0}{\degree} over tiles, against the \SI{4.66}{\degree} of the large field. The combined tiles give the same values within \SI{0.1}{\degree}. The 32 held-out rock tiles carry 4.0 rocks per tile, against 3.4 expected, and rocks cover 0.0032 of the tile area, against 0.0030 that the rock model with $k=0.02$ predicts for diameters of 0.2 to \SI{0.6}{\metre}.
% Source: review_2026-10-24/summary.md C19 (review_2026_10_24.py, generator metric median |dz|/L over x and y pairs): undulations MAS 4 m 1.59 deg (margin 0.4 m), 1.92 deg (margin 1.5 m, tiles 0.94-5.02); combined 1.61 / 1.93 (0.97-4.94). Rocks regenerated from seeds 2000-2031 with sample_rocks: 4.0 per 8 m tile (0-14, expected 3.4), area fraction 0.0032; Golombek-Rapp F_k(0.2)-F_k(0.6) = 0.0030 at k 0.02, 0.0008 at k 0.0125.

% Limitation: small rocks missing, same families in training and tests
Rocks under \SI{0.2}{\metre} are \SI{86}{\percent} of the rocks of at least \SI{0.1}{\metre} by count. Training and tests use the same four tile families, with 10 training tiles per family.

% Method: inclined mesh geometry
Falls on the inclined mesh are judged by height above the local surface. The mesh is a closed \SI{120}{\metre} slab \SI{1}{\metre} thick whose top face is meshed in \SI{0.5}{\metre} triangles, with the robot spawned heading along the grade.

\subsection{Soil Models}
\label{sec:rft-supp}
% Limitation: effects that 3D RFT omits and the extensions that address them
3D RFT, as used here, has no cohesion and no footprint history. Later work extends RFT to dynamic intrusion~\cite{agarwal_surprising_2021} and to cohesive powders~\cite{kerimoglu_extending_2025}, and neither extension is used in this study.

% Method: lunar regolith reference values and MPM settings
Lunar regolith at 0 to \SI{15}{\centi\metre} depth has a friction angle of \SI{42}{\degree} and a cohesion of \SI{0.52}{\kilo\pascal}, and its bulk density rises from 1.30 at the surface to \SI{1.92}{\gram\per\centi\metre\cubed} at depth, with 1.50 at 0 to \SI{15}{\centi\metre}~\cite{carrier_physical_1991}. MPM and CRM use \SI{2}{\centi\metre} voxels and particles. MPM uses a uniform bed with a Drucker--Prager friction coefficient of 0.99 and a yield stress of \SI{570}{\pascal}, matched to this friction angle and cohesion, and a foot--grain friction of 0.84. The grains are incompressible with no elastic modulus fitted, and no plate test relates the bed to the bootprint modulus. MPM runs at about 0.004 simulated seconds per GPU second for one robot on 1.4 to 2.6 million particles in a \SI{0.12}{\metre} bed.
% Source: LuPNT-isaac src/lupnt_lab_tasks/soil/lunar_mpm_soil.py:15-27 (phi 42, c 520 Pa, rho 1500 uniform, FRICTION 0.99, YIELD_STRESS 570); scripts/gm_loco/mpm_walk.py:45,139-143 (voxel 0.02 m default, young_modulus 1e15). The Newton form sqrt(J2) <= s + mu p was verified in source (vault RAL-paper-design, lunar soil parameters 2026-10-09), so 0.99 is the Drucker-Prager coefficient and not tan phi. Cost: NAVLab results/mpm_g1best/runs_v2 wall_per_sim_second_excl_first_step_s, median 274 s over the 48 Flat episodes (111-801, L40S / RTX 4090), mpm.particles 1.43-2.57 M; bed depth mpm.depth 0.12. Foot-grain friction 0.84 = regolith_action.py interface_friction default (lunar).

% Method: SCM Bekker set and CRM solver
The SCM Bekker set of Table 9.14 of the Lunar Sourcebook~\cite{carrier_physical_1991}, whose footnotes flag errors in the conversion of some values from English to SI units, is $k_c=\SI{1.4}{\kilo\newton\per\metre\squared}$, $k_\phi=\SI{820}{\kilo\newton\per\metre\cubed}$, $n=1$, $c=\SI{0.17}{\kilo\pascal}$, $\phi=\SI{35}{\degree}$, and $K=\SI{1.78}{\centi\metre}$ (listed as \SI{1.8}{\centi\metre}), with elastic stiffness \SI{2e8}{\pascal\per\metre} and damping \SI{3e4}{\pascal\second\per\metre}. This full set of eight values is the lunar soil of the Curiosity rover SCM demonstration distributed with Chrono. This Bekker fit gives SCM a lower friction angle and cohesion than the \SI{42}{\degree} and \SI{0.52}{\kilo\pascal} of MPM and CRM, so the engines model lunar regolith from one source but not as one identical material. Its sinkage modulus on the \SI{0.07}{\metre} sole is $k_c/b+k_\phi=\SI{0.84}{\mega\pascal\per\metre}$, close to the Apollo bootprint mean of \SI{0.80}{\mega\pascal\per\metre} (\SI{8}{\kilo\pascal\per\centi\metre}), though both come from the same source and the closeness is not an independent check. Chrono loads the robot from a URDF of the same asset, which is checked against the simulator model for link set, link offsets, masses, and inertias before use. On rigid flat ground the exported model reproduces the training engine. For 63 of the 72 combinations of policy, seed, and gravity, the survival over 256 episodes in the training engine lies inside the Wilson interval of the 32 Chrono episodes. Six of the nine exceptions are Earth-trained policies at the Moon, which survive 0.81 to 0.92 in Chrono against 0.64 to 0.75 in the training engine, two are seeds of \JointFlatRigid at Earth, and one is a seed of \MoonRoughRFT at the Moon that survives 0.86 in the training engine and every Chrono episode. CRM uses friction $\tan\SI{42}{\degree}$ with stiffness set for numerical stability. Its multibody system uses a different iterative solver from the Chrono default, because the default does not keep the articulated robot assembled even on rigid ground. CRM runs at 21 to 27 minutes per \SI{10}{\second} episode on one RTX 2080 Ti. The foot friction is 0.5 on rigid ground and in SCM and CRM.
% Source: CRM solver and cost: results/chrono_g1best/crm/README.md; rigid foot friction 0.5 (foot_friction).
% Source: Chrono vs Newton rigid, review_2026-10-24/summary.md C19 and review_numbers.json C19.chrono_vs_newton_per_seed (gate_pass from collab_results/chrono_g1best/summary.json): 72 rows, 9 fail: Earth-Flat-Rigid s42-44 Moon, Earth-Flat-RFT s43 Moon, Earth-Rough-Rigid s44 Moon, Earth-Rough-RFT s42 Moon, EarthMoon-Flat-Rigid s42/s43 Earth (Newton 0.965/0.879, Chrono 0.875/0.781), Moon-Rough-RFT s44 Moon (Newton 0.863, Chrono 1.00 [0.893, 1.0]). Seed means at the Moon for the Earth-trained: Chrono 0.81-0.92, Newton 0.64-0.75. Chrono reads joint effort limits from the USD (spec.effort_source).
% Source: SCM_SOIL in chrono_eval.py:101 equals all 8 values of chrono/src/demos/robot/curiosity/demo_ROBOT_Curiosity_SCM.cpp:171; URDF round-trip check newton_to_urdf.py:183 (links, offsets 1e-4 m, mass, inertia 1e-6).

% Method: target depths of the calibrated 3D RFT soils
The three calibrated lunar 3D RFT densities are fitted so that the walking foot sinkage of \MoonFlatRFT, averaged over its three seeds, at 0.5 and \SI{1.0}{\metre\per\second} matches each target bootprint depth scaled to the G1 foot~\cite{carrier_physical_1991}. \MoonRoughRFT is not used for the fit, because it loads a pitched foot whose ankle point stays above the surface. The test soils are thus tied to the gait of one evaluated policy, and lunar soil results are conditional on this stiffness. The fitted densities are effective parameters of 3D RFT and not physical bulk densities. The nominal value of about \SI{5200}{\kilo\gram\per\metre\cubed} is about 3.5 times the \SI{1500}{\kilo\gram\per\metre\cubed} bulk density of regolith at 0 to \SI{15}{\centi\metre}~\cite{carrier_physical_1991}, so at Earth gravity the lunar soil cells test a 3D RFT material with no physical counterpart. The compacted soil of \SI{16000}{\kilo\gram\per\metre\cubed} matches no bootprint and serves as a stress test beyond the training range rather than as a calibrated soil. The loose soil of \SI{1346}{\kilo\gram\per\metre\cubed} matches the Lunar Sourcebook modulus of \SI{2}{\kilo\pascal\per\centi\metre}, the value recommended for \SI{95}{\percent} confidence in acceptable settlement, with a G1 target depth of \SI{17.8}{\milli\metre}. The crater-rim soil of about \SI{2900}{\kilo\gram\per\metre\cubed} matches the \SI{1.6}{\centi\metre} rim bootprints, with a target depth of \SI{8.1}{\milli\metre}. The nominal lunar soil of about \SI{5200}{\kilo\gram\per\metre\cubed} matches the Lunar Sourcebook mean of about \SI{8}{\kilo\pascal\per\centi\metre}, with a target depth of \SI{4.5}{\milli\metre}. No measurement of foot-to-regolith friction is used for the surface friction of 0.6. In 3D RFT, MPM, and SCM the soft bed follows the tile surface on the combined held-out tile, with rocks rigid, and in MPM the combined-tile cells are tile episodes 0 and 1.
% Source: Lunar Sourcebook Ch. 9 p. 519 (Fig. 9.37, 776 bootprints): k = 8 kPa/cm mean, 2 kPa/cm for 95 % confidence in settlement. Verified 2026-10-20.
% Source: docs/regolith_calibration/insim_calibration.md table and Aditya-progress-2026-10-16.txt sec. 3 (targets 17.8 / 8.1 / 4.5 mm).

\label{sec:validation}
% Method: gravity scaling of the soil models brackets the measurements, bootprints as reference
3D RFT, linear in $g$, and SCM, independent of $g$, bracket the measured gravity scaling of regolith strength (Section~\ref{sec:rft}). The Apollo bootprints, with a mean subgrade modulus of \SI{8}{\kilo\pascal\per\centi\metre} and mean depths of \SI{0.7}{\centi\metre} between craters and \SI{1.6}{\centi\metre} on crater rims~\cite{carrier_physical_1991}, serve as a reference rather than a validation target. 3D RFT and MPM both gain strength with gravity, so their agreement alone does not show robustness to it.

\subsection{Evaluation Details}
\label{sec:protocol-supp}
% Setup: spawn, test conditions, fall bookkeeping
Every robot starts from the default joint pose with zero joint velocity. Tile cells spawn each robot at its standing height above the highest ground under its feet. In training, rough cells spawn robots at the tile-center height instead, so with the spawn offset a robot can start up to about 7 cm inside the relief. Foot friction in tests is fixed at 0.5, the effective value against the 0.3 ground, so a foot can hold static on a grade only up to the friction-cone limit $\tan^{-1}(0.5)=\SI{26.6}{\degree}$ at any gravity. The height and tilt fall rule of the Unitree recipe~\cite{unitree_rl_lab_2025} is applied here relative to the local ground. Each fall is recorded as a height or a tilt termination, and MPM episodes that leave the bed are excluded and counted.
% Source: up to 7 cm is the collaborator's estimate, not re-measured.
% Source: joint reset scale (1.0, 1.0) from gm_loco/flat_env_cfg.py:61, confirmed in the v2 run dump (events.reset_robot_joints.position_range [1.0, 1.0]), so there is no joint-reset randomization in training either.

% Setup: command profiles
Each episode commands one of eight profiles: forward at (1) 0.25, (2) 0.5, and (3) \SI{1.0}{\metre\per\second}, (4) backward at \SI{0.5}{\metre\per\second}, (5) lateral at \SI{0.3}{\metre\per\second}, (6) \SI{0.5}{\metre\per\second} while turning at \SI{0.5}{\radian\per\second}, and steps \SI{4}{\second} into the motion (7) from 0.25 to \SI{1.0}{\metre\per\second} and (8) from \SI{1.0}{\metre\per\second} to a hold.

% Definition: tracking error and gait statistics
Tracking error $e_v$, in m/s, is the mean absolute error of the forward and lateral base-frame velocities, averaged over the two components, over the second half of the last constant-command segment of each episode that covers it to its end (episode counts in Table~\ref{tab:ev_survivors}). For profile 8 this window is the zero-command hold, and yaw-rate error is not included. We also report step frequency, duty factor, and flight fraction over survivors.

% Definition: cost of transport as a mechanical proxy, with its results in the supplement
Cost of transport is $\mathrm{CoT}=E_{\rm abs}/(mgd_\parallel)$ with joint work $E_{\rm abs}=\int_0^{T_{\mathrm{ep}}}\sum_j|\tau_j\dot q_j|\,dt$ and local gravity $g$, a mechanical proxy that excludes electrical losses, and episodes with less than \SI{0.2}{\metre} of progress are excluded from it and counted.

% Result: MPM combined-bed episodes excluded for leaving the bed, with per-policy counts
On the MPM combined beds an episode with more than \SI{5}{\percent} of its frames off the bed is excluded together with its rigid pair. This removes 1 of the 36 lunar-gravity episodes of the six lunar-capable policies run in MPM, 2 of the 6 episodes of \EarthRoughRFT at the Moon, and 1 of the 6 combined-bed episodes of \EarthFlatRFT at Earth. Because leaving the bed depends on how the policy walks, this exclusion depends on the outcome. For the eight grid policies at their training gravity it removes 2 episodes, and keeping them changes no seed-mean MPM ratio by more than 0.01.

% Setup: episodes per cell and the restriction of SCM to the forward profiles
Each rigid and 3D RFT cell has 256 episodes per seed over the 8 profiles, 32 each, and in 3D RFT cells every environment is on a \SI{0.12}{\metre} soil layer, the deepest seen in training. Each SCM cell has 32 episodes per seed and ground, 8 profiles with 4 starts each. MPM and CRM run one episode per seed and forward profile (0.5 and \SI{1.0}{\metre\per\second}), 6 per policy and cell. Restricting SCM to these two profiles changes no sign of Table~\ref{tab:soil_results}.

% Definition: progress on turning and stop profiles, tracking windows and the e_v coverage rule (LuPNT-private projects/guillemc/RAL_Humanoid/gait_metrics.py _covered, COVER_SLACK_STEPS = 1.5, step_dt 0.02 s = sim_dt 0.005 x decimation 4)
The lateral profile runs half of its episodes to the left and half to the right, and the turning profile half each way. For the turning profile signed progress is the distance traveled along the turning path, and for the stop profile it includes the hold. The tracking window is the second half of the last constant-command segment, which is the last \SI{4}{\second} of the single-command profiles and the last \SI{2}{\second} of the second segment of the step profiles, and an episode enters $e_v$ only if its record reaches the end of the window to within 1.5 control steps (\SI{0.03}{\second} at \SI{50}{\hertz}), so an episode that falls earlier is left out.

% Method: approximate per-seed intervals of the sign rule and their half-widths
In every cell episodes are the sampling unit, and the interval on a survival difference is the per-seed Newcombe interval, which combines the two Wilson intervals. Both intervals treat the 256 episodes of a cell as independent trials, although the episodes of a cell share 32 tiles and 8 starts, and the difference interval also treats the two runs as independent although they share these tiles and starts. The intervals are therefore approximate. The original evaluation records store episodes without tile identifiers. A separate evaluation of the four combined-tile cells logged them, and a tile-cluster bootstrap on it changes no verdict (Section~\ref{sec:signrule-supp}). Few runs give unreliable interval estimates, for which Agarwal et al. recommend stratified bootstrap intervals~\cite{agarwal2021deep}, so we report every seed. With 256 episodes per cell the per-seed Newcombe interval of a survival difference has a half-width of about 0.05 at a baseline of 0.9 and 0.09 at 0.5, and with 32 episodes per profile about 0.22 at a baseline of 0.7 and 0.23 at 0.5, so smaller differences cannot meet the sign rule.
% Source: Newcombe half-width at equal rates, sqrt((p-l)^2+(u-p)^2) from the Wilson bounds (recomputed 2026-10-27): 0.053 at p=0.9 and 0.086 at 0.5 for n=256, 0.216 at 0.7 and 0.231 at 0.5 for n=32.

% Method: run-to-run spread of the GPU simulation, treated as no difference
The GPU simulation is not bitwise deterministic, so a cell re-run with identical seeds can give a different survival. Re-running all 18 3D RFT soil cells for each of the 12 policies and seeds, 648 evaluations at both gravities, with identical seeds changed survival in 313 of them, by at most 0.063. Every pass or fail of the seed sign rule is the same on the re-run. The binomial standard error of a survival of 0.5 over 256 episodes is 0.031, so the largest change is about two standard errors. The mean change is 0.007 in magnitude and \SI{95}{\percent} of the changes are at most 0.031. A seed mean over the three seeds of a cell changes by at most 0.035, and the per-seed soil progress ratio $\rho_v$ of the 3D RFT soil cells by at most 0.047. The tables report the original run, and we treat differences within this band as no difference.

\section{Result Details}
\label{sec:results-supp}
% Scope: what this supplement adds to the results of the main text
This section gives result details that the main text summarizes, in the order of its result subsections.

\subsection{Gravity Transfer}
\label{sec:res_gravity-supp}
\begin{figure*}[!tb]
  \centering
  \includegraphics[width=\textwidth]{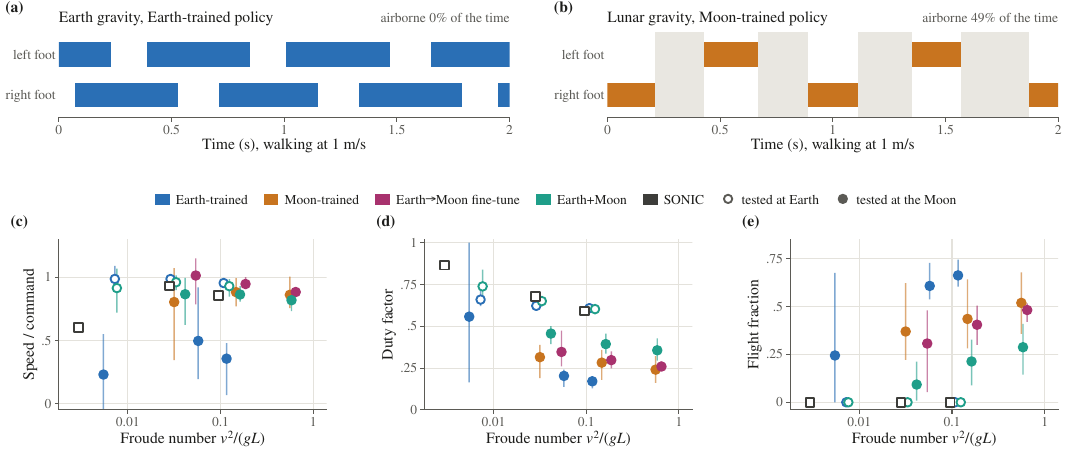}
  \caption{\textbf{Gait timing against Froude number.} (a, b) Foot contact timelines walking at \SI{1}{\metre\per\second} for an Earth-trained policy at Earth, airborne 0\% of the time, and a Moon-trained policy at the Moon, airborne 49\% of the time. (c--e) Speed over command, duty factor, and flight fraction against the Froude number $v^2/(gL)$ for each policy group of the legend and SONIC. Hollow markers are tested at Earth and filled markers at the Moon. Each marker is the mean over the policies and seeds of one group at one command speed of 0.25, 0.5, or \SI{1.0}{\metre\per\second} on flat rigid ground, 12 policy seeds for the Earth- and Moon-trained groups, 6 for the fine-tuned and gravity-randomized groups, and the single SONIC checkpoint. Gait statistics are computed over surviving episodes only. Each policy seed contributes the mean over its surviving episodes, and the bars span the lowest to the highest policy seed. Policy seeds with fewer than 25\% surviving episodes are left out, which removes SONIC at the Moon and the Moon-trained and fine-tuned policies at Earth.}
  \label{fig:froude}
\end{figure*}
% Source: fig:froude generator from the collaborator (collab_results/froude_figure/drafts/figures_v1/scripts/fig_froude.py: MIN_SURV 0.25 line 32, per-seed survivor mean lines 56-61, bar = seed min-max line 149, marker = seed mean line 151; README): 12/12/6/6/1 policy-seeds, Earth-trained at the Moon at 1 m/s has 9 after the cutoff. Panel (a) axis label reads Time (s) in the figure, needs Inkscape to re-export.
% Source: fig:froude generator from the collaborator (collab_results/froude_figure/drafts/figures_v1/scripts/fig_froude.py: MIN_SURV 0.25 line 32, per-seed survivor mean lines 56-61, bar = seed min-max line 149, marker = seed mean line 151; README): 12/12/6/6/1 policy-seeds, Earth-trained at the Moon at 1 m/s has 9 after the cutoff. Panel (a) axis label reads Time (s) in the figure, needs Inkscape to re-export.

% Result: gait timing against Froude number and the toe-first landing
Fig.~\ref{fig:froude} shows the gait timing behind the flight fractions of the main text. At the Moon the policies land on their toes with a short stance. For one seed of \MoonFlatRigid at the Moon on flat rigid ground the duty factor is 0.25, the flight fraction 0.49, and the step frequency \SI{3.0}{\hertz}, against 0.62, 0, and \SI{2.8}{\hertz} for \EarthFlatRigid at Earth. The GM-Loco reward terms~\cite{kamohara_granular_2026}, which penalize foot orientation but include no touchdown foot-pitch term, may allow this landing, and its low cost at lunar gravity may favor it.
% Source: toe-landing gait, Isaac eval FlatRigid, Moon-Flat-Rigid_s42 at the Moon vs Earth-Flat-Rigid_s42 at Earth (duty factor 0.25 vs 0.62, flight fraction 0.49 vs 0, step frequency 3.0 vs 2.8 Hz); GM-Loco reward table has a foot-orientation penalty and no touchdown foot-pitch term. No foot-pitch angle was measured, the toe landing is observed in the rollouts.

% Result: Moon-trained policies fall at Earth during the standing phase
The Moon-trained policies and the Earth-to-Moon fine-tunes fall at Earth on flat rigid ground in every episode, on average about \SI{1.3}{\second} after the start, and all but one of their 4,608 episodes fall before the \SI{2}{\second} standing phase ends.
% Source: eval_best FlatRigid_Earth summary.duration_s_falls, mean over the 18 Moon/fine-tune runs 1.29 s (per run 0.94-1.69 s). episodes_data of the same 18 files (2026-10-27): 4608 of 4608 episodes fall, 1 with duration_s >= 2.0.

% Result: the collapse at Earth is consistent with ankle overload in the crouched posture, knee and hip torque stay below their limits
The joint torques of the first \SI{1.3}{\second} at Earth show how these policies fall. For five of the six Moon-trained and fine-tuned policies the ankle pitch reaches its hard joint limit in 67 to \SI{100}{\percent} of the episodes, at a median of 0.52 to \SI{0.94}{\second}, and the ankle torque sits at its \SI{25}{\newton\metre} limit in 3.6 to \SI{10.6}{\percent} of the steps. Knee torque never reaches its limit on average, with a 99th percentile near 0.40 of the \SI{139}{\newton\metre} limit, and hip torque reaches its limit in at most \SI{0.5}{\percent} of the steps. The knees fold past \SI{90}{\degree} only in the last tenth of a second before the fall, after the ankle limit. The knee torque is close to the static torque the crouched posture needs, so the failure sequence is consistent with ankle overload in the learned crouched posture at Earth weight rather than with a lack of knee strength. We did not test this cause directly, for example by raising the ankle limits. The sixth policy, \MoonFlatRFT, slumps forward without reaching the joint limits and falls at \SI{1.43}{\second}. The Earth-trained policies at Earth never saturate a joint and stand with 12 to \SI{18}{\degree} of knee flexion.
% Source: output/RAL_Humanoid/review_2026-10-26/c9_torque/c9_table.md (analysis/run_c9_traces.py, c9_torque_analysis.py; eval_locomotion.py --joint_trace records robot.data.applied_torque), FlatRigid_Earth, 3 seeds x 64 episodes, window t <= 1.3 s on alive steps; saturation |tau| >= 0.99 effort limit from env.yaml (hip 88/139, knee 139, ankle 25 N m); limit hit within 0.02 rad of joint_pos_limits. Ankle pitch at limit [% ep]: Moon-Flat-Rigid 100, Moon-Rough-Rigid 67, Moon-Rough-RFT 100, Earth2Moon-Flat-Rigid 100, Earth2Moon-Rough-RFT 100, Moon-Flat-RFT 6. Sat. ankle 8.8/8.4/10.0/3.6/10.6 (Moon-Flat-RFT 0.5). Sat. hip <= 0.5, sat. knee 0.0. First ankle-limit contact median 0.52 s (Moon-Rough-RFT), 0.94 s (Moon-Flat-Rigid); knee passes 90 deg at median 0.92 s vs fall 0.99 s (Moon-Rough-RFT), 1.16 vs 1.29 s (Moon-Flat-Rigid). Knee |tau|/139 99th percentile about 0.40; static knee torque estimate about 20 N m at 54 deg and 35 N m at 100 deg flexion vs measured 17 N m at 0.5 s and 24-38 N m while flexed 90-110 deg. Independent recomputation from the npz traces 2026-10-26 (/tmp/claude-1001/c9check/).

\begin{figure}[!tb]
  \centering
  \includegraphics[width=\linewidth]{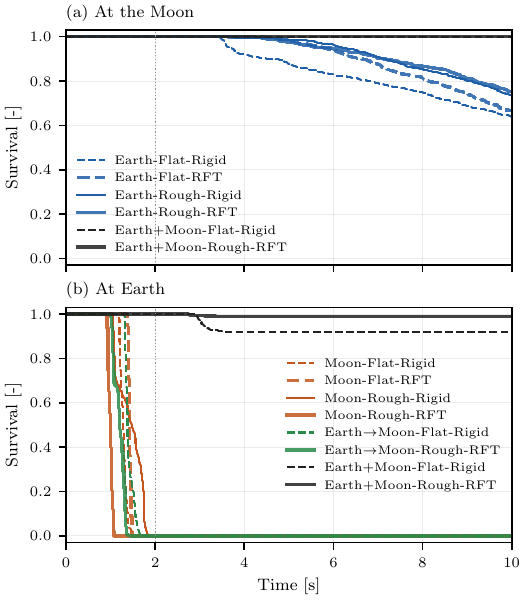}
  \caption{Survival against time on flat rigid ground at the gravity a policy was not trained for. Curves are Kaplan-Meier style estimates from the logged fall times, pooled over three seeds (768 episodes per policy), with survivors censored at the \SI{10}{\second} end of the episode, so each curve is the fraction of episodes not yet fallen. (a) Earth-trained and gravity-randomized policies at the Moon. (b) Moon-trained, fine-tuned, and gravity-randomized policies at Earth. The dotted line marks the end of the \SI{2}{\second} standing phase. The Earth-trained policies at the Moon still fall at \SI{10}{\second}, while the Moon-trained and fine-tuned policies at Earth fall during the standing phase, all but one of 4608 episodes before \SI{2}{\second}.}
  \label{fig:survival_time}
\end{figure}

% Result: Earth-trained policies at the Moon keep falling until the episode ends
Fig.~\ref{fig:survival_time} shows when the falls happen. On flat rigid ground at the Moon, \EarthFlatRigid falls 0, 61, 70, 62, and 84 times in the five \SI{2}{\second} intervals of the episode and \EarthRoughRigid 0, 3, 25, 84, and 89 times, out of 768 episodes each, so the \SI{10}{\second} survival of 0.64 to 0.75 of the Earth-trained policies at the Moon is an upper bound that a longer walk can only lower.
% Source: review_2026-10-24/summary.md C7 (duration_s of eval_best FlatRigid_{Moon,Earth}, pooled over seeds 42-44): falls per 2 s bin Earth-Flat-Rigid 0/61/70/62/84, Earth-Rough-Rigid 0/3/25/84/89; S(10 s) Earth-Flat-Rigid 0.64, Earth-Flat-RFT 0.67, Earth-Rough-Rigid 0.74, Earth-Rough-RFT 0.75. Figure copied from review_2026-10-24/survival_vs_time.pdf (review_2026_10_24.py).

% Result: cost of transport per gravity
The cost of transport of Section~\ref{sec:protocol-supp} is reported at each gravity separately, since it divides by the weight. On flat rigid ground at Earth the policies that walk there spend 0.42 to 0.81, and at the Moon the Moon-trained, fine-tuned, and gravity-randomized policies spend 1.4 to 2.8. The Earth-trained policies at the Moon spend 13 to 26, about 5 to 19 times more per meter than the policies trained for lunar gravity.
% Source: eval_best FlatRigid_{Earth,Moon} summary.cot.mean, mean over seeds 42-44 (checked 2026-10-19): Earth 0.42 (Earth-Rough-RFT) to 0.81 (Earth+Moon-Flat-Rigid); Moon 1.37 (Earth2Moon-Rough-RFT) to 2.83 (Earth+Moon-Flat-Rigid); Earth policies at the Moon 13.3 (Earth-Rough-Rigid) to 25.5 (Earth-Flat-RFT). TODO: CoT column per policy.

% Source: tables/tab_perprofile.tex, written by make_tab_keycells.py from eval_best TilesRigidCombined_{Earth,Moon} profile_summary.by_profile.survival.rate. Wilson intervals per profile are in the same JSONs (32 episodes per seed) and are not tabulated. SONIC is not in eval_best.

% Result: survival falls with commanded speed for the flat-trained policies, and the rough-trained ones hold across profiles at their training gravity
The pooled survival of Table~\ref{tab:gravity} hides a dependence on the command, which the file \texttt{perprofile.csv} gives for each of the eight command profiles on the rigid combined tiles, with 32 episodes per profile and seed. At Earth \EarthFlatRigid survives 0.90 of the forward \SI{0.25}{\metre\per\second} episodes on the combined tiles but 0.17 of the forward \SI{1.0}{\metre\per\second} ones, while \EarthRoughRigid and \EarthRoughRFT have a mean survival of 0.94 or above on every profile.

% Role: per-profile summary at the Moon and the profile reversal of the Earth rough-trained policies
At the Moon the fast forward profile is also the hardest one for the flat-trained policies of lunar gravity, with 0.62 for \MoonFlatRigid and 0.44 for \MoonFlatRFT, while the rough-trained policies that saw lunar gravity keep at least 0.91 on every profile. The Earth rough-trained policies at the Moon reverse this order, because they survive more of the forward \SI{1.0}{\metre\per\second} episodes than of the \SI{0.25}{\metre\per\second} ones, 0.55 against 0.43 for \EarthRoughRigid and 0.67 against 0.57 for \EarthRoughRFT, and their seeds differ most, by up to 0.57 on a single profile.

\subsection{Inclined Mesh and Tilted Gravity}
\label{sec:res_slope-supp}
% Comparison: the 10 degree inclined mesh against the tilted-gravity slope of the same grade
For the four rigid-trained policies of Table~\ref{tab:terrain}, mean survival on the \SI{10}{\degree} inclined mesh lies within 0.05 of that on the tilted-gravity slope of the same grade in every cell but three. \EarthFlatRigid at Earth survives 0.74 of the uphill mesh episodes against 0.81 with tilted gravity, \EarthRoughRigid at the Moon survives 0.62 of the downhill mesh episodes against 0.57, and \MoonRoughRigid at the Moon survives 0.86 of the downhill mesh episodes against 1.00, because one of its seeds survives only 0.58 of them.
% Source: eval_best SlopeRigid10{Up,Down} vs MeshSlope10{Up,Down}, mean over seeds 42-44 (count of cells beyond 0.05 from sign_rule_table.py: 3); Earth-Rough-Rigid MeshSlope10Down_Moon 0.62 vs SlopeRigid10Down_Moon 0.57; Moon-Rough-Rigid MeshSlope10Down_Moon seeds 1.00/0.58/1.00.

% Comparison: one Earth rough-trained seed on the 20 degree downhill tilted-gravity slope
On the tilted-gravity slopes at Earth, one seed of \EarthRoughRigid falls in most \SI{20}{\degree} downhill episodes, with survival 0.98, 0.99, and 0.05 over the three seeds (Table~\ref{tab:gravity}).

% Result: per-seed survival over the full slope sweep behind the slope figure
Tables~\ref{tab:slope_perseed} and~\ref{tab:slope_perseed_moon} give the survival of every seed on each grade of the slope sweep of Fig.~\ref{fig:slope}, at Earth and at the Moon.
% Source: tables/tab_slope_perseed.tex, written by make_tab_keycells.py from eval_best SlopeRigid{5,10,15,20,26,30}{Up,Down}_{Earth,Moon} (the rigid control JointRoughRigid from eval_c3_joint_rough_rigid); survival = 1 - fall_rate per seed.
\begin{table*}[!tb]
  \centering\scriptsize
  \caption{\textbf{Slope results per seed at Earth gravity.} Survival $S$ on rigid slopes of \SI{5}{\degree} to \SI{30}{\degree}, climbed (Up) or descended (Down), for each of the three seeds of every policy. RFT in policy names is 3D resistive force theory (3D RFT).}
  \label{tab:slope_perseed}
  \setlength{\tabcolsep}{3pt}
  \begin{tabular*}{\linewidth}{@{\extracolsep{\fill}}|l|c|cccccc|cccccc|}
  \hline
  & & \multicolumn{6}{c|}{\textbf{Up}} & \multicolumn{6}{c|}{\textbf{Down}} \\
  \hdrstrut \textbf{Policy} & \textbf{Seed} & \SI{5}{\degree} & \SI{10}{\degree} & \SI{15}{\degree} & \SI{20}{\degree} & \SI{26}{\degree} & \SI{30}{\degree} & \SI{5}{\degree} & \SI{10}{\degree} & \SI{15}{\degree} & \SI{20}{\degree} & \SI{26}{\degree} & \SI{30}{\degree} \\
  \hline\hline
  \EarthFlatRigid & 42 & 1.00 & 0.62 & 0.61 & 0.44 & 0.00 & 0.00 & 1.00 & 1.00 & 0.98 & 0.37 & 0.00 & 0.00 \\
  \EarthFlatRigid & 43 & 1.00 & 0.82 & 0.72 & 0.27 & 0.00 & 0.00 & 1.00 & 1.00 & 0.84 & 0.08 & 0.00 & 0.00 \\
  \EarthFlatRigid & 44 & 1.00 & 1.00 & 0.59 & 0.36 & 0.00 & 0.00 & 1.00 & 1.00 & 0.96 & 0.12 & 0.00 & 0.00 \\
  \hline
  \EarthFlatRFT & 42 & 1.00 & 1.00 & 0.70 & 0.02 & 0.00 & 0.00 & 1.00 & 0.95 & 0.89 & 0.33 & 0.00 & 0.00 \\
  \EarthFlatRFT & 43 & 1.00 & 1.00 & 0.50 & 0.00 & 0.00 & 0.00 & 1.00 & 0.87 & 0.83 & 0.21 & 0.00 & 0.00 \\
  \EarthFlatRFT & 44 & 1.00 & 1.00 & 0.99 & 0.21 & 0.00 & 0.00 & 1.00 & 1.00 & 0.96 & 0.23 & 0.00 & 0.00 \\
  \hline
  \EarthRoughRigid & 42 & 1.00 & 1.00 & 1.00 & 1.00 & 0.00 & 0.00 & 1.00 & 1.00 & 0.99 & 0.98 & 0.00 & 0.00 \\
  \EarthRoughRigid & 43 & 1.00 & 1.00 & 1.00 & 1.00 & 0.00 & 0.00 & 1.00 & 1.00 & 1.00 & 0.99 & 0.00 & 0.00 \\
  \EarthRoughRigid & 44 & 1.00 & 1.00 & 1.00 & 1.00 & 0.01 & 0.00 & 1.00 & 1.00 & 1.00 & 0.05 & 0.00 & 0.00 \\
  \hline
  \EarthRoughRFT & 42 & 1.00 & 1.00 & 1.00 & 1.00 & 0.00 & 0.00 & 1.00 & 1.00 & 1.00 & 1.00 & 0.02 & 0.00 \\
  \EarthRoughRFT & 43 & 1.00 & 1.00 & 1.00 & 1.00 & 0.01 & 0.00 & 1.00 & 1.00 & 1.00 & 0.93 & 0.00 & 0.00 \\
  \EarthRoughRFT & 44 & 1.00 & 1.00 & 1.00 & 0.97 & 0.00 & 0.00 & 1.00 & 1.00 & 1.00 & 1.00 & 0.00 & 0.00 \\
  \hline
  \MoonFlatRigid & 42 & 0.00 & 0.00 & 0.00 & 0.00 & 0.00 & 0.00 & 0.00 & 0.00 & 0.00 & 0.00 & 0.00 & 0.00 \\
  \MoonFlatRigid & 43 & 0.00 & 0.00 & 0.00 & 0.00 & 0.00 & 0.00 & 0.00 & 0.00 & 0.00 & 0.00 & 0.00 & 0.00 \\
  \MoonFlatRigid & 44 & 0.00 & 0.00 & 0.00 & 0.00 & 0.00 & 0.00 & 0.00 & 0.00 & 0.00 & 0.00 & 0.00 & 0.00 \\
  \hline
  \MoonFlatRFT & 42 & 0.00 & 0.00 & 0.00 & 0.00 & 0.00 & 0.00 & 0.00 & 0.00 & 0.00 & 0.00 & 0.00 & 0.00 \\
  \MoonFlatRFT & 43 & 0.00 & 0.00 & 0.00 & 0.00 & 0.00 & 0.00 & 0.00 & 0.00 & 0.00 & 0.00 & 0.00 & 0.00 \\
  \MoonFlatRFT & 44 & 0.00 & 0.00 & 0.00 & 0.00 & 0.00 & 0.00 & 0.00 & 0.00 & 0.00 & 0.00 & 0.00 & 0.00 \\
  \hline
  \MoonRoughRigid & 42 & 0.00 & 0.00 & 0.00 & 0.00 & 0.00 & 0.00 & 0.00 & 0.00 & 0.00 & 0.00 & 0.00 & 0.00 \\
  \MoonRoughRigid & 43 & 0.00 & 0.00 & 0.00 & 0.00 & 0.00 & 0.00 & 0.00 & 0.00 & 0.00 & 0.00 & 0.00 & 0.00 \\
  \MoonRoughRigid & 44 & 0.00 & 0.00 & 0.00 & 0.00 & 0.00 & 0.00 & 0.00 & 0.00 & 0.00 & 0.00 & 0.00 & 0.00 \\
  \hline
  \MoonRoughRFT & 42 & 0.00 & 0.00 & 0.00 & 0.00 & 0.00 & 0.00 & 0.00 & 0.00 & 0.00 & 0.00 & 0.00 & 0.00 \\
  \MoonRoughRFT & 43 & 0.00 & 0.00 & 0.00 & 0.00 & 0.00 & 0.00 & 0.00 & 0.00 & 0.00 & 0.00 & 0.00 & 0.00 \\
  \MoonRoughRFT & 44 & 0.00 & 0.00 & 0.00 & 0.00 & 0.00 & 0.00 & 0.00 & 0.00 & 0.00 & 0.00 & 0.00 & 0.00 \\
  \hline
  \FtFlatRigid & 42 & 0.00 & 0.00 & 0.00 & 0.00 & 0.00 & 0.00 & 0.00 & 0.00 & 0.00 & 0.00 & 0.00 & 0.00 \\
  \FtFlatRigid & 43 & 0.00 & 0.00 & 0.00 & 0.00 & 0.00 & 0.00 & 0.00 & 0.00 & 0.00 & 0.00 & 0.00 & 0.00 \\
  \FtFlatRigid & 44 & 0.00 & 0.00 & 0.00 & 0.00 & 0.00 & 0.00 & 0.00 & 0.00 & 0.00 & 0.00 & 0.00 & 0.00 \\
  \hline
  \FtRoughRFT & 42 & 0.00 & 0.00 & 0.00 & 0.00 & 0.00 & 0.00 & 0.00 & 0.00 & 0.00 & 0.00 & 0.00 & 0.00 \\
  \FtRoughRFT & 43 & 0.00 & 0.00 & 0.00 & 0.00 & 0.00 & 0.00 & 0.00 & 0.00 & 0.00 & 0.00 & 0.00 & 0.00 \\
  \FtRoughRFT & 44 & 0.00 & 0.00 & 0.00 & 0.00 & 0.00 & 0.00 & 0.00 & 0.00 & 0.00 & 0.00 & 0.00 & 0.00 \\
  \hline
  \JointFlatRigid & 42 & 1.00 & 0.70 & 0.39 & 0.01 & 0.00 & 0.00 & 1.00 & 1.00 & 0.66 & 0.32 & 0.00 & 0.00 \\
  \JointFlatRigid & 43 & 1.00 & 0.92 & 0.45 & 0.50 & 0.00 & 0.00 & 1.00 & 1.00 & 0.71 & 0.06 & 0.00 & 0.00 \\
  \JointFlatRigid & 44 & 1.00 & 0.66 & 0.60 & 0.00 & 0.00 & 0.00 & 1.00 & 0.58 & 0.10 & 0.00 & 0.00 & 0.00 \\
  \hline
  \JointRoughRFT & 42 & 1.00 & 1.00 & 1.00 & 1.00 & 0.00 & 0.00 & 1.00 & 1.00 & 1.00 & 0.74 & 0.00 & 0.00 \\
  \JointRoughRFT & 43 & 1.00 & 1.00 & 1.00 & 1.00 & 0.00 & 0.00 & 0.99 & 1.00 & 1.00 & 0.94 & 0.00 & 0.00 \\
  \JointRoughRFT & 44 & 1.00 & 1.00 & 1.00 & 1.00 & 0.01 & 0.00 & 1.00 & 1.00 & 1.00 & 0.91 & 0.00 & 0.00 \\
  \hline
  \JointRoughRigid & 42 & 1.00 & 1.00 & 1.00 & 1.00 & 0.00 & 0.00 & 1.00 & 1.00 & 1.00 & 0.76 & 0.00 & 0.00 \\
  \JointRoughRigid & 43 & 1.00 & 1.00 & 1.00 & 1.00 & 0.00 & 0.00 & 1.00 & 1.00 & 1.00 & 0.96 & 0.00 & 0.00 \\
  \JointRoughRigid & 44 & 1.00 & 1.00 & 1.00 & 0.96 & 0.00 & 0.00 & 1.00 & 1.00 & 0.97 & 0.75 & 0.00 & 0.00 \\
  \hline
  \end{tabular*}
\end{table*}
\begin{table*}[!tb]
  \centering\scriptsize
  \caption{\textbf{Slope results per seed at Moon gravity.} Survival $S$ on rigid slopes of \SI{5}{\degree} to \SI{30}{\degree}, climbed (Up) or descended (Down), for each of the three seeds of every policy. RFT in policy names is 3D resistive force theory (3D RFT).}
  \label{tab:slope_perseed_moon}
  \setlength{\tabcolsep}{3pt}
  \begin{tabular*}{\linewidth}{@{\extracolsep{\fill}}|l|c|cccccc|cccccc|}
  \hline
  & & \multicolumn{6}{c|}{\textbf{Up}} & \multicolumn{6}{c|}{\textbf{Down}} \\
  \hdrstrut \textbf{Policy} & \textbf{Seed} & \SI{5}{\degree} & \SI{10}{\degree} & \SI{15}{\degree} & \SI{20}{\degree} & \SI{26}{\degree} & \SI{30}{\degree} & \SI{5}{\degree} & \SI{10}{\degree} & \SI{15}{\degree} & \SI{20}{\degree} & \SI{26}{\degree} & \SI{30}{\degree} \\
  \hline\hline
  \EarthFlatRigid & 42 & 0.83 & 0.82 & 0.58 & 0.13 & 0.00 & 0.00 & 0.68 & 0.34 & 0.10 & 0.01 & 0.00 & 0.00 \\
  \EarthFlatRigid & 43 & 0.55 & 0.50 & 0.24 & 0.02 & 0.00 & 0.00 & 0.46 & 0.24 & 0.01 & 0.00 & 0.00 & 0.00 \\
  \EarthFlatRigid & 44 & 0.68 & 0.50 & 0.31 & 0.09 & 0.02 & 0.00 & 0.64 & 0.66 & 0.12 & 0.02 & 0.00 & 0.00 \\
  \hline
  \EarthFlatRFT & 42 & 0.56 & 0.41 & 0.18 & 0.06 & 0.00 & 0.00 & 0.65 & 0.46 & 0.16 & 0.19 & 0.01 & 0.00 \\
  \EarthFlatRFT & 43 & 0.36 & 0.17 & 0.09 & 0.01 & 0.00 & 0.00 & 0.55 & 0.28 & 0.20 & 0.03 & 0.01 & 0.00 \\
  \EarthFlatRFT & 44 & 0.80 & 0.61 & 0.42 & 0.07 & 0.02 & 0.00 & 0.72 & 0.57 & 0.21 & 0.10 & 0.02 & 0.00 \\
  \hline
  \EarthRoughRigid & 42 & 0.80 & 0.67 & 0.29 & 0.30 & 0.00 & 0.39 & 0.61 & 0.75 & 0.41 & 0.16 & 0.00 & 0.00 \\
  \EarthRoughRigid & 43 & 0.80 & 0.83 & 0.73 & 0.70 & 0.35 & 0.08 & 0.71 & 0.59 & 0.50 & 0.27 & 0.04 & 0.00 \\
  \EarthRoughRigid & 44 & 0.67 & 0.66 & 0.55 & 0.12 & 0.00 & 0.04 & 0.68 & 0.36 & 0.40 & 0.21 & 0.00 & 0.00 \\
  \hline
  \EarthRoughRFT & 42 & 0.79 & 0.82 & 0.77 & 0.57 & 0.27 & 0.00 & 0.66 & 0.36 & 0.20 & 0.04 & 0.01 & 0.00 \\
  \EarthRoughRFT & 43 & 0.92 & 0.94 & 0.80 & 0.71 & 0.50 & 0.15 & 0.88 & 0.82 & 0.54 & 0.36 & 0.01 & 0.00 \\
  \EarthRoughRFT & 44 & 0.62 & 0.56 & 0.48 & 0.40 & 0.38 & 0.00 & 0.52 & 0.45 & 0.06 & 0.09 & 0.00 & 0.00 \\
  \hline
  \MoonFlatRigid & 42 & 1.00 & 0.91 & 0.62 & 0.23 & 0.00 & 0.00 & 1.00 & 0.94 & 0.76 & 0.50 & 0.18 & 0.09 \\
  \MoonFlatRigid & 43 & 1.00 & 0.94 & 0.61 & 0.11 & 0.00 & 0.00 & 1.00 & 0.99 & 0.80 & 0.50 & 0.12 & 0.02 \\
  \MoonFlatRigid & 44 & 1.00 & 1.00 & 0.82 & 0.44 & 0.00 & 0.00 & 1.00 & 0.96 & 0.81 & 0.42 & 0.07 & 0.00 \\
  \hline
  \MoonFlatRFT & 42 & 1.00 & 1.00 & 0.58 & 0.39 & 0.00 & 0.00 & 1.00 & 1.00 & 0.85 & 0.42 & 0.07 & 0.00 \\
  \MoonFlatRFT & 43 & 0.98 & 0.77 & 0.52 & 0.21 & 0.01 & 0.00 & 0.95 & 1.00 & 0.82 & 0.55 & 0.25 & 0.00 \\
  \MoonFlatRFT & 44 & 0.99 & 0.89 & 0.57 & 0.20 & 0.00 & 0.00 & 1.00 & 0.99 & 0.86 & 0.35 & 0.06 & 0.00 \\
  \hline
  \MoonRoughRigid & 42 & 1.00 & 1.00 & 1.00 & 1.00 & 0.98 & 0.66 & 1.00 & 1.00 & 1.00 & 1.00 & 0.01 & 0.00 \\
  \MoonRoughRigid & 43 & 1.00 & 1.00 & 1.00 & 1.00 & 0.92 & 0.04 & 1.00 & 1.00 & 0.96 & 0.00 & 0.00 & 0.00 \\
  \MoonRoughRigid & 44 & 1.00 & 1.00 & 1.00 & 1.00 & 0.94 & 0.62 & 1.00 & 1.00 & 1.00 & 1.00 & 0.79 & 0.05 \\
  \hline
  \MoonRoughRFT & 42 & 1.00 & 1.00 & 1.00 & 0.91 & 0.89 & 0.55 & 1.00 & 1.00 & 1.00 & 0.99 & 0.88 & 0.40 \\
  \MoonRoughRFT & 43 & 1.00 & 1.00 & 1.00 & 0.99 & 0.62 & 0.00 & 1.00 & 1.00 & 1.00 & 0.99 & 0.66 & 0.00 \\
  \MoonRoughRFT & 44 & 0.91 & 0.71 & 0.73 & 0.53 & 0.41 & 0.00 & 1.00 & 1.00 & 0.99 & 0.98 & 0.65 & 0.00 \\
  \hline
  \FtFlatRigid & 42 & 1.00 & 1.00 & 0.90 & 0.39 & 0.05 & 0.00 & 1.00 & 0.98 & 0.85 & 0.67 & 0.02 & 0.00 \\
  \FtFlatRigid & 43 & 1.00 & 1.00 & 0.51 & 0.17 & 0.02 & 0.00 & 1.00 & 1.00 & 0.89 & 0.64 & 0.00 & 0.00 \\
  \FtFlatRigid & 44 & 1.00 & 1.00 & 0.70 & 0.38 & 0.08 & 0.02 & 1.00 & 0.96 & 0.82 & 0.70 & 0.31 & 0.00 \\
  \hline
  \FtRoughRFT & 42 & 1.00 & 1.00 & 0.95 & 0.84 & 0.75 & 0.00 & 1.00 & 1.00 & 1.00 & 0.99 & 0.92 & 0.68 \\
  \FtRoughRFT & 43 & 1.00 & 1.00 & 1.00 & 1.00 & 0.81 & 0.35 & 1.00 & 1.00 & 1.00 & 1.00 & 0.89 & 0.73 \\
  \FtRoughRFT & 44 & 1.00 & 1.00 & 1.00 & 1.00 & 0.99 & 0.77 & 1.00 & 1.00 & 1.00 & 1.00 & 0.93 & 0.52 \\
  \hline
  \JointFlatRigid & 42 & 1.00 & 0.76 & 0.49 & 0.02 & 0.00 & 0.00 & 1.00 & 0.96 & 0.71 & 0.22 & 0.04 & 0.00 \\
  \JointFlatRigid & 43 & 1.00 & 0.98 & 0.47 & 0.10 & 0.00 & 0.00 & 1.00 & 0.97 & 0.72 & 0.27 & 0.03 & 0.00 \\
  \JointFlatRigid & 44 & 1.00 & 0.80 & 0.40 & 0.05 & 0.00 & 0.00 & 0.75 & 0.82 & 0.25 & 0.03 & 0.00 & 0.00 \\
  \hline
  \JointRoughRFT & 42 & 1.00 & 1.00 & 1.00 & 1.00 & 0.85 & 0.18 & 1.00 & 1.00 & 1.00 & 0.97 & 0.70 & 0.00 \\
  \JointRoughRFT & 43 & 1.00 & 1.00 & 1.00 & 1.00 & 0.84 & 0.54 & 1.00 & 1.00 & 1.00 & 0.99 & 0.66 & 0.00 \\
  \JointRoughRFT & 44 & 1.00 & 1.00 & 0.98 & 0.97 & 0.92 & 0.24 & 1.00 & 1.00 & 1.00 & 0.91 & 0.48 & 0.00 \\
  \hline
  \JointRoughRigid & 42 & 1.00 & 1.00 & 1.00 & 1.00 & 0.97 & 0.73 & 1.00 & 1.00 & 1.00 & 0.84 & 0.64 & 0.04 \\
  \JointRoughRigid & 43 & 1.00 & 1.00 & 1.00 & 1.00 & 0.96 & 0.77 & 1.00 & 1.00 & 1.00 & 1.00 & 0.88 & 0.03 \\
  \JointRoughRigid & 44 & 1.00 & 1.00 & 1.00 & 1.00 & 0.96 & 0.79 & 1.00 & 1.00 & 0.98 & 0.86 & 0.89 & 0.00 \\
  \hline
  \end{tabular*}
\end{table*}

\subsection{Slope Traversal}
\label{sec:res_traversal-supp}
% Definition: friction-cone limit and the traversal outcome
Survival counts a robot as surviving while its pelvis stays within the height and tilt limits, also when it slides. With the test foot friction of 0.5 a foot can hold static only up to the friction-cone limit $\tan^{-1}(0.5)=\SI{26.6}{\degree}$ at any gravity, so the \SI{26}{\degree} grade lies \SI{0.6}{\degree} below the limit and the \SI{30}{\degree} grade beyond it. We therefore also report traversal $T$, the fraction of episodes that survive with positive signed progress along the command and with progress below 1.5 times the median progress of the same policy and seed over its survivors on flat rigid ground under the same command profile and gravity. Signed progress integrates the body-frame velocity along the commanded direction, so a robot that faces uphill and slides down reads negative, and the upper bound flags a robot that slides faster than it walks. $T$ is stricter than survival even on flat rigid ground, where it keeps at least 0.64 of the survivors of every policy at either gravity. On the sideways profile progress runs across the grade, so a robot that steps sideways while sliding downhill can still count as traversing.
% Source: review_2026-10-24/summary.md C1 (friction 0.5 static = dynamic in all 2014 eval logs that print it, ground 0.3; T definition; flat T/S minimum 0.644; lateral-profile limitation).

% Result: at 26 and 30 degrees the surviving robots slide, at 20 degrees the lunar-capable rough policies walk
Table~\ref{tab:slope_traversal} gives $S$ and $T$ on the 20, 26, and \SI{30}{\degree} grades. At the Moon on the \SI{26}{\degree} uphill grade, the rough-trained policies that saw lunar gravity keep a seed-mean survival of 0.64 to 0.95 but traverse only 0.14 to 0.29 of the episodes, and 59 to \SI{64}{\percent} of their survivors move backward. On the \SI{30}{\degree} downhill grade the forward-walking survivors of \FtRoughRFT cover 2.8 to 3.5 times the commanded distance per seed, so they slide down the grade, and its traversal is 0.07. At \SI{20}{\degree} at the Moon, \FtRoughRFT and \JointRoughRFT traverse 0.93 and 0.99 of the uphill episodes, and none of their uphill survivors moves backward.
% Source: review_2026-10-24/summary.md C1 key-cell table: 26 Up Moon S MRR 0.95 / MRT 0.64 / FRT 0.85 / JRT 0.87, T 0.19 / 0.14 / 0.21 / 0.29, survivors with progress < 0 0.64 / 0.64 / 0.59 / 0.61; 30 Down Moon FRT forward survivors progress / commanded 3.07 [2.84, 3.46], T 0.07; 20 Up Moon T FRT 0.93, JRT 0.99, backward survivors 0.00.
% Source: tables/tab_slope_traversal.tex, written by make_tab_keycells.py from eval_best SlopeRigid{20,26,30}{Up,Down}_{Moon,Earth} (the rigid control JointRoughRigid from eval_c3_joint_rough_rigid); S = 1 - fall_rate, T = slope_traversal.traversal (per-episode progress_m against the FlatRigid survivors of the same policy, seed, gravity and profile), mean over seeds.
\begin{table*}[!tb]
  \centering\scriptsize
  \caption{\textbf{Slope survival and traversal.} Seed-mean survival $S$ and traversal $T$ on rigid slopes of \SI{20}{\degree}, \SI{26}{\degree}, and \SI{30}{\degree}, climbed (Up) or descended (Down), at lunar and Earth gravity. An episode traverses when it survives and its signed progress along the command is positive and below 1.5 times the median progress of the same policy's and seed's survivors on flat rigid ground under the same command profile. With a foot friction coefficient of 0.5 the friction-cone limit is $\tan^{-1}(0.5) = \SI{26.6}{\degree}$, so on the \SI{26}{\degree} and \SI{30}{\degree} grades a robot can stay upright while it slides, and survival alone does not show that it walks the slope. RFT in policy names is 3D resistive force theory (3D RFT).}
  \label{tab:slope_traversal}
  \setlength{\tabcolsep}{3pt}
  \begin{tabular*}{\linewidth}{@{\extracolsep{\fill}}|l|cc|cc|cc|cc|cc|cc|}
  \hline
  & \multicolumn{6}{c|}{\textbf{Up}} & \multicolumn{6}{c|}{\textbf{Down}} \\
  & \multicolumn{2}{c|}{\SI{20}{\degree}} & \multicolumn{2}{c|}{\SI{26}{\degree}} & \multicolumn{2}{c|}{\SI{30}{\degree}} & \multicolumn{2}{c|}{\SI{20}{\degree}} & \multicolumn{2}{c|}{\SI{26}{\degree}} & \multicolumn{2}{c|}{\SI{30}{\degree}} \\
  \hdrstrut \textbf{Policy} & $S$ & $T$ & $S$ & $T$ & $S$ & $T$ & $S$ & $T$ & $S$ & $T$ & $S$ & $T$ \\
  \hline\hline
  \multicolumn{13}{|l|}{\textbf{At Moon}} \\ \hline
  \EarthRoughRigid & 0.37 & 0.03 & 0.12 & 0.00 & 0.17 & 0.00 & 0.21 & 0.04 & 0.01 & 0.00 & 0.00 & 0.00 \\
  \EarthRoughRFT & 0.56 & 0.12 & 0.38 & 0.05 & 0.05 & 0.00 & 0.16 & 0.02 & 0.01 & 0.00 & 0.00 & 0.00 \\
  \hline
  \MoonRoughRigid & 1.00 & 0.83 & 0.95 & 0.19 & 0.44 & 0.04 & 0.67 & 0.54 & 0.26 & 0.07 & 0.02 & 0.00 \\
  \MoonRoughRFT & 0.81 & 0.72 & 0.64 & 0.14 & 0.18 & 0.04 & 0.99 & 0.74 & 0.73 & 0.16 & 0.13 & 0.02 \\
  \FtRoughRFT & 0.95 & 0.93 & 0.85 & 0.21 & 0.37 & 0.05 & 0.99 & 0.77 & 0.91 & 0.33 & 0.64 & 0.07 \\
  \hline
  \JointFlatRigid & 0.06 & 0.02 & 0.00 & 0.00 & 0.00 & 0.00 & 0.17 & 0.01 & 0.02 & 0.00 & 0.00 & 0.00 \\
  \JointRoughRFT & 0.99 & 0.99 & 0.87 & 0.29 & 0.32 & 0.05 & 0.96 & 0.91 & 0.61 & 0.20 & 0.00 & 0.00 \\
  \JointRoughRigid & 1.00 & 0.90 & 0.96 & 0.25 & 0.77 & 0.12 & 0.90 & 0.83 & 0.80 & 0.51 & 0.02 & 0.00 \\
  \hline
  \multicolumn{13}{|l|}{\textbf{At Earth}} \\ \hline
  \EarthRoughRigid & 1.00 & 0.99 & 0.00 & 0.00 & 0.00 & 0.00 & 0.68 & 0.64 & 0.00 & 0.00 & 0.00 & 0.00 \\
  \EarthRoughRFT & 0.99 & 0.96 & 0.00 & 0.00 & 0.00 & 0.00 & 0.98 & 0.91 & 0.01 & 0.00 & 0.00 & 0.00 \\
  \hline
  \MoonRoughRigid & 0.00 & 0.00 & 0.00 & 0.00 & 0.00 & 0.00 & 0.00 & 0.00 & 0.00 & 0.00 & 0.00 & 0.00 \\
  \MoonRoughRFT & 0.00 & 0.00 & 0.00 & 0.00 & 0.00 & 0.00 & 0.00 & 0.00 & 0.00 & 0.00 & 0.00 & 0.00 \\
  \FtRoughRFT & 0.00 & 0.00 & 0.00 & 0.00 & 0.00 & 0.00 & 0.00 & 0.00 & 0.00 & 0.00 & 0.00 & 0.00 \\
  \hline
  \JointFlatRigid & 0.17 & 0.03 & 0.00 & 0.00 & 0.00 & 0.00 & 0.13 & 0.00 & 0.00 & 0.00 & 0.00 & 0.00 \\
  \JointRoughRFT & 1.00 & 0.97 & 0.00 & 0.00 & 0.00 & 0.00 & 0.86 & 0.77 & 0.00 & 0.00 & 0.00 & 0.00 \\
  \JointRoughRigid & 0.98 & 0.80 & 0.00 & 0.00 & 0.00 & 0.00 & 0.82 & 0.67 & 0.00 & 0.00 & 0.00 & 0.00 \\
  \hline
  \end{tabular*}
\end{table*}

\subsection{Terrain per Seed}
\label{sec:res_terrain_seed-supp}
\begin{table*}[!tb]
  \centering\footnotesize
  \caption{\textbf{Survival and tracking on held-out terrain at both gravities. RFT in policy names is 3D resistive force theory (3D RFT).}}
  \label{tab:terrain}
  \setlength{\tabcolsep}{3pt}
  \begin{tabular*}{\linewidth}{@{\extracolsep{\fill}}|l|l|ccccc|cc|c|}
\hline
& & \multicolumn{5}{c|}{\textbf{Tiles, $S\uparrow$}} & \multicolumn{2}{c|}{\textbf{Mesh, $S\uparrow$}} & \textbf{Comb.} \\
\hdrstrut \textbf{Policy} & \textbf{Test} & Flat & Crat. & Rock & Und. & Comb. & Up & Down & $e_v\downarrow$ \\
\hline\hline
\EarthFlatRigid & Earth & 1.00 & 0.98\,{\tiny[0.98,0.99]} & 0.98 & 0.57\,{\tiny[0.52,0.60]} & 0.53\,{\tiny[0.50,0.56]} & 0.74\,{\tiny[0.62,0.92]} & 1.00 & 0.10\,{\tiny[0.09,0.11]} \\
\EarthFlatRigid & Moon & 0.64\,{\tiny[0.56,0.70]} & 0.66\,{\tiny[0.57,0.71]} & 0.68\,{\tiny[0.57,0.73]} & 0.45\,{\tiny[0.34,0.50]} & 0.45\,{\tiny[0.34,0.50]} & 0.64\,{\tiny[0.51,0.80]} & 0.39\,{\tiny[0.19,0.57]} & 0.25\,{\tiny[0.23,0.27]} \\
\EarthRoughRigid & Earth & 1.00 & 1.00 & 1.00 & 0.99\,{\tiny[0.99,1.00]} & 0.98\,{\tiny[0.98,0.99]} & 1.00 & 1.00 & 0.04 \\
\EarthRoughRigid & Moon & 0.74\,{\tiny[0.62,0.80]} & 0.74\,{\tiny[0.66,0.79]} & 0.76\,{\tiny[0.71,0.82]} & 0.47\,{\tiny[0.37,0.55]} & 0.43\,{\tiny[0.31,0.55]} & 0.75\,{\tiny[0.65,0.85]} & 0.62\,{\tiny[0.55,0.73]} & 0.26\,{\tiny[0.25,0.27]} \\
\hline
\MoonFlatRigid & Earth & 0.00 & 0.00 & 0.00 & 0.00 & 0.00 & 0.00 & 0.00 & -- \\
\MoonFlatRigid & Moon & 1.00 & 0.99\,{\tiny[0.98,1.00]} & 1.00 & 0.85\,{\tiny[0.80,0.91]} & 0.80\,{\tiny[0.77,0.86]} & 0.94\,{\tiny[0.84,1.00]} & 0.98\,{\tiny[0.96,1.00]} & 0.08\,{\tiny[0.07,0.09]} \\
\MoonRoughRigid & Earth & 0.00 & 0.00 & 0.00 & 0.00 & 0.00 & 0.00 & 0.00 & -- \\
\MoonRoughRigid & Moon & 1.00 & 1.00 & 1.00 & 0.99\,{\tiny[0.99,1.00]} & 0.99\,{\tiny[0.98,0.99]} & 1.00 & 0.86\,{\tiny[0.58,1.00]} & 0.10\,{\tiny[0.09,0.11]} \\
\hline
SONIC & Earth & 1.00 & 1.00 & 1.00 & 0.91 & 0.90 & 1.00 & 1.00 & 0.20 \\
SONIC & Moon & 0.10 & 0.10 & 0.09 & 0.05 & 0.08 & 0.06 & 0.00 & 0.32 \\
\hline
\end{tabular*}

  \tabnote{Survival $S$ on rigid held-out tiles of each family and on a physically inclined \SI{10}{\degree} mesh with vertical gravity, and velocity tracking error $e_v$ in m/s on the combined tiles over the episodes that reach the last steady window, whose counts are in Table~\ref{tab:ev_survivors}. Crat., Und., and Comb. are the crater, undulation, and combined tiles. Rigid results of the RFT-trained policies are in Tables~\ref{tab:gravity} and~\ref{tab:route}. Values are means over seeds, with the seed range in brackets when seeds differ.}
    % Source SONIC: NAVLab results/sonic_paper/tab_terrain_sonic.json; 256 episodes per cell, Wilson 95 % intervals (flat, mesh) or a bootstrap over the 32 tiles (tiles) in that file and tab_terrain_sonic.tex; mesh with 0.5 m cells (incline_terrain.py cell 0.5); e_v over episodes reaching the steady windows (Earth 230, Moon 20).
  % Source tab_terrain.tex (other columns, written by projects/guillemc/RAL_Humanoid/figures/make_tab_keycells.py): output/RAL_Humanoid/eval_best/<policy>_s<seed>__{FlatRigid,TilesRigid{Craters,Rocks,Undulations,Combined},MeshSlope10{Up,Down}}_{Earth,Moon}.json, 256 episodes per cell, mean over seeds 42-44; mesh cells on the fixed closed mesh (LuPNT-private fd3ceadab); e_v = pooled mean of the vx and vy steady-window errors (profile_summary.pooled.tracking), over the episodes that reach the last steady window (the SONIC row's definition), and -- where no episode reaches it.
\end{table*}

% Result: undulations are the costly tile feature for flat-trained policies, and the cross-gravity losses per family
At Earth, \EarthFlatRigid survives within 0.02 of \EarthRoughRigid on flat, crater, and rock tiles but falls to about half survival on undulations and combined tiles, and \MoonFlatRigid loses less on the same tiles at the Moon (Table~\ref{tab:terrain}). Undulations are therefore the costly feature for the flat-trained policies at both gravities, while craters and rocks cost little. \EarthFlatRigid, which already falls often on undulations and combined tiles at Earth, loses survival at the Moon on flat, crater, and rock tiles. Tracking error on the combined tiles is also much larger for the Earth-trained policies at the Moon than at Earth.

% Source: tables/tab_perseed_terrain.tex, written by make_tab_keycells.py from eval_best {FlatRigid,TilesRigid*,MeshSlope10*}_{Earth,Moon}; v_par/v_c = progress_m / move_s / commanded speed, mean over the three forward profiles (progress includes a fall's partial progress); slip = profile_summary.pooled.gait.foot_slip_ms.mean. Episodes carry no tile index and the JSONs hold no foot clearance, so tile-bootstrap intervals and clearance cannot be generated from eval_best.

% Result: seed spread of the terrain results, progress speed of the Moon-trained policies at the Moon
The per-seed values behind Table~\ref{tab:terrain} are in \texttt{perseed\_terrain.csv}, which also gives the progress speed over the commanded speed and the foot slip speed on the combined tiles. The Moon-trained policies reach 0.73 to 0.86 of the commanded forward speed on the combined tiles at the Moon, and \EarthRoughRigid reaches 0.93 to 0.96 at Earth. At the Moon, however, \EarthRoughRigid makes almost no forward progress on these tiles, with 0.18, 0.15, and $-0.03$ of the commanded speed over the three seeds, and the two Earth rigid-trained policies slip at 0.21 to \SI{0.31}{\metre\per\second} there, against 0.08 to 0.10 at Earth. A single seed drives the largest spread, as one seed of \MoonRoughRigid survives 0.58 of the \SI{10}{\degree} downhill mesh episodes at the Moon while the other two survive all of them (Section~\ref{sec:res_slope-supp}), and on the same mesh the seeds of \EarthFlatRigid at the Moon span 0.19 to 0.57.

\subsection{Soil Training on Lunar Soil at Earth Gravity}
\label{sec:res_soil_earth-supp}
% Source: tables/tab_earthregolith.tex, written by projects/guillemc/RAL_Humanoid/figures/make_tab_keycells.py from eval_best SoilFlat{,Regolith}_Earth and SoilTilesCombined{,Regolith}_Earth.

% Result: survival of the Earth policies on the nominal lunar soil at Earth gravity in 3D RFT
On the nominal lunar soil in 3D RFT at Earth gravity (\texttt{earth\_regolith.csv}), which lies outside the Earth training range and is stiffer than the Earth sand, with higher density and internal friction, all four Earth-trained policies survive 0.99 to 1.00 of the flat-soil episodes. On the combined tiles under this soil, \EarthRoughRigid and \EarthRoughRFT both survive 0.98 of episodes, and \EarthFlatRFT survives 0.68 against 0.38 for \EarthFlatRigid. On the softer Earth sand, the same tiles give 0.87 against 0.30 for the rough pair and 0.69 against 0.27 for the flat pair. In 3D RFT, soil training therefore helps the rough pair only on Earth sand, while the flat pair gains on the tiles under both soils. The seed ranges of the two policies of a pair do not overlap in any of these three tile comparisons, with 0.62 to 0.73 against 0.34 to 0.41 for the flat pair under lunar soil, 0.61 to 0.76 against 0.22 to 0.34 under Earth sand, and 0.72 to 0.95 against 0.20 to 0.43 for the rough pair under Earth sand.
% Source: eval_best SoilFlatRegolith_Earth and SoilTilesCombinedRegolith_Earth success_rate, 256 episodes per seed, mean over seeds 42-44 (seed ranges: Earth-Flat-Rigid tiles 0.34-0.41, Earth-Flat-RFT 0.62-0.73, Earth-Rough-Rigid 0.97-0.99, Earth-Rough-RFT 0.97-0.98). Earth sand: tab_keycells Soil columns at Earth.

% Result: survival of the Earth policies on the nominal lunar soil at Earth gravity in MPM
In MPM on the same soil at Earth gravity, the rough pair survives every flat-bed and combined-bed episode. \EarthFlatRigid and \EarthFlatRFT each fall in one of 6 flat-bed episodes, which they survive on the rigid floor of that engine, and survive 3 of 6 and 4 of 5 combined-bed episodes, the same counts as on the rigid tiles. One combined-bed episode of \EarthFlatRFT leaves the bed and is excluded. With 6 episodes per cell MPM can show only large differences, and it shows none between the two flat-pair policies.
% Source: NAVLab results/mpm_g1best_comb/reg981_summary.{md,json} (summarize_reg981.py), 48 episodes, Earth-Flat-RFT_s42 tile0 p2 excluded (39 % of frames off the bed). Rigid pairs: mpm_g1best/runs_v2 hard200 (flat, 6/6 for all four) and mpm_g1best_comb/hard hard200_tile (comb, 3/6, 4/5, 6/6, 6/6).

% Result: held-out engines at Earth gravity on lunar regolith detect no soil-training effect in progress
At Earth gravity on lunar regolith in the held-out engines, MPM and SCM agree on the sign of the soil-training difference in progress for the flat pair and disagree for the rough pair (Table~\ref{tab:soil_results}). With soil training the flat pair keeps more of its rigid-ground progress on the flat bed in both engines, as in 3D RFT and CRM. On the combined bed it keeps less in both engines, in MPM by only 0.02, and there both MPM ratios of the flat pair are above 1 mainly because rigid references fall early. The rough pair is ahead on the flat MPM bed and behind on the combined one, the reverse of SCM. No difference meets the agreement rule of Section~\ref{sec:res_soilmodel}, a descriptive rule without intervals that asks for the same sign in all three seeds in both MPM and SCM, and single falls in 6 MPM episodes set the MPM means, so we find no held-out soil-training effect in progress at Earth on lunar soil.
% Source: seed means RFT minus rigid. MPM flat bed: flat pair .86-.76 = +.10, rough pair 1.06-1.03 = +.03. MPM comb: flat pair 1.24-1.26 = -.02, rough pair 1.02-1.04 = -.02. SCM flat bed: flat pair 1.05-.96 = +.09, rough pair 1.03-1.06 = -.03. SCM comb: flat pair 1.18-1.37 = -.19, rough pair .95-.91 = +.04. MPM comb per seed (reg981_summary.json): Earth-Flat-Rigid 2.14/1.07/.55, s42 rigid tile1 p3 fell at 0.91 m while MPM reached 6.24 m. Earth-Flat-RFT 1.07/1.05/1.59, s44 rigid tile0 p2 fell at 0.00 m. MPM flat-bed falls: Earth-Flat-Rigid s42 p3 at 3.2 s, Earth-Flat-RFT s43 p3 at 3.2 s.

% Caveat: two 3D RFT failures on flat Earth sand, their causes and times
Two policies fail on flat Earth sand in 3D RFT although they walk on flat rigid ground at Earth. \JointFlatRigid falls in all 768 flat-sand episodes, all but 9 of them by tilt, at a median time of 1.9 to \SI{2.2}{\second} per seed, so 343 of the falls come within the \SI{2}{\second} standing phase. It survives 0.88 to 0.96 per seed on rigid ground and 0.88 to 1.00 on the stiffer lunar soil at Earth gravity. \EarthRoughRigid falls on flat sand by tilt at median times of 3.1 to \SI{5.7}{\second}, with seed survival 0.54, 0.11, and 0.34, while it survives every rigid episode and \EarthFlatRigid survives 0.92 to 0.99 of the sand episodes. Almost all of these failures trigger the tilt criterion rather than the height criterion. The logged data hold no force or sinkage traces, so they cannot show whether these falls are a real loss of balance on soft soil or an artifact of the per-foot 3D RFT wrench.
% Source: review_2026-10-24/summary.md C15 (eval_best, run "best"): EarthMoon-Flat-Rigid SoilFlat_Earth s42/s43/s44 S 0.00, 256 falls each, causes s44 247 bad_orientation + 3 base_height + 6 both (s42, s43 all bad_orientation), medians 2.18/1.88/2.04 s, falls during stand 61/157/125 = 343; FlatRigid_Earth 0.96/0.88/0.91; SoilFlatRegolith_Earth 0.88/0.99/1.00. Earth-Rough-Rigid SoilFlat_Earth S 0.54/0.11/0.34, all bad_orientation, medians 5.66/3.86/3.14 s; FlatRigid_Earth 1.00 x3. Earth-Flat-Rigid SoilFlat_Earth 0.92/0.97/0.99. Re-run gives the same picture.

\subsection{Soil Progress Ratio over Survivors}
\label{sec:res_rhov-supp}
% Source: 3D RFT at the Moon, nominal lunar soil, forward profiles 2 and 3, mean over seeds 42-44, rho_v over all episodes (as in the main tables) and over surviving episodes only, values supplied and checked by the coordinator on 2026-10-19 from output/RAL_Humanoid/eval_best/<policy>_s<seed>__{SoilTilesCombined,TilesRigidCombined}_Moon.json.
\begin{table}[!tb]
  \centering\footnotesize
  \caption{\textbf{Soil progress ratio over all episodes and over survivors.} 3D resistive force theory (3D RFT, written RFT in policy names) combined tiles under the nominal lunar soil at the Moon, forward profiles, mean over seeds.}
  \label{tab:rhov_survivors}
  \setlength{\tabcolsep}{4pt}
  \begin{tabular*}{\linewidth}{@{\extracolsep{\fill}}|l|cc|}
    \hline
    \hdrstrut \textbf{Policy} & All & Survivors \\
    \hline\hline
    \MoonFlatRigid & 1.35 & 1.13 \\
    \MoonFlatRFT & 1.46 & 1.16 \\
    \FtFlatRigid & 1.27 & 1.14 \\
    \JointFlatRigid & 1.69 & 1.19 \\
    \hline
    \MoonRoughRigid & 0.87 & 0.89 \\
    \MoonRoughRFT & 1.08 & 1.05 \\
    \FtRoughRFT & 1.04 & 1.02 \\
    \JointRoughRFT & 1.04 & 1.04 \\
    \hline
  \end{tabular*}
\end{table}

% Result: survivors-only rho_v compares condition-specific survivor populations, falls on the rigid reference inflate the combined-tile values
The soil progress ratio $\rho_v$ of the main text includes progress before a fall, so a policy that falls more often on the rigid reference than on soil reads above 1. On flat soil at the Moon this hardly matters. Over all episodes $\rho_v$ lies between 0.89 and 1.14 for every policy, and over survivors between 0.89 and 1.07, with only \MoonRoughRFT changing, from 1.14 to 1.04. On the combined tiles the values well above 1 come mostly from falls on the rigid tiles (Table~\ref{tab:rhov_survivors}). The survivors of the soil run and of the rigid run are condition-specific populations, and the evaluation files do not pair episodes, so they can be different tiles and starts. The survivor values therefore compare these two populations and do not isolate a speed benefit of soil. Between them the four flat-trained policies trained at lunar gravity still progress 13 to \SI{19}{\percent} faster on soil-covered tiles than on rigid tiles, and the rough-trained ones stay within \SI{11}{\percent} of their rigid progress.

\subsection{Soil Training per Seed}
\label{sec:res_soil_seed-supp}
% Source: tables/tab_perseed_soil.tex, written by make_tab_keycells.py: 3D RFT from eval_best, MPM from collab_results/mpm_g1best{,_comb}, SCM and CRM from the per-seed entries of scm_table.json and crm_table.json. Per-seed excluded MPM episodes are in comb_summary.

% Source: tables/tab_soil_vs_rigid.tex, written by make_tab_keycells.py: survival and profile_summary.pooled.cot.mean of SoilFlat{Regolith_Earth,_Moon} and SoilTilesCombined{Regolith_Earth,_Moon} over FlatRigid and TilesRigidCombined at the policy's gravity. Sinkage is not in the JSONs and stays an open TODO.

% Source: tables/tab_soil_falls.tex, written by make_tab_keycells.py from profile_summary.pooled.fall_causes of the soil cells of tab_soil_vs_rigid (bad_orientation, base_height, bad_orientation+base_height).

% Source: tables/tab_soil_geometries.tex, written by make_tab_keycells.py from eval_best SoilFlat_{Earth,Moon} and SoilTiles{Craters,Rocks,Undulations,Combined}_{Earth,Moon}. The Earth cells are Earth sand because only the flat and combined cells exist on lunar regolith at Earth. eval_best holds five soil geometries, not six.

% Definition: geometries and episodes behind the soil progress ratio
In Table~\ref{tab:soil_results}, Flat is a flat bed and Comb. a bed that follows the held-out combined tiles, all 32 in 3D RFT and SCM and two in MPM. $\rho_v$ uses profiles 2 and 3, 64 of the 256 episodes per seed in 3D RFT and 2 per seed in MPM and CRM, and all 32 episodes per seed and ground in SCM. MPM combined-bed episodes that leave the bed are excluded (Section~\ref{sec:protocol}). The per-seed values are in \texttt{perseed\_soil.csv} and the seed ranges in \texttt{soil\_rhov\_seed\_ranges.csv}.

% Result: soil training at Earth on rigid ground, the 20 degree uphill grade, and flat sand
At Earth, we detect no consistent survival loss from soil training on rigid flat ground or rigid tiles, but it lowers survival on the \SI{20}{\degree} uphill grade for the flat pair (Tables~\ref{tab:gravity} and~\ref{tab:keycells-full}). On flat sand the difference between \EarthFlatRFT and \EarthFlatRigid is not consistent across seeds.

% Result: per-seed spread of the soil progress ratio and the single-seed effects behind its means
The seeds behind the means of Table~\ref{tab:soil_results} are in \texttt{perseed\_soil.csv}, and three seed effects stand out there. One seed of \MoonFlatRFT reads 2.51 on the MPM combined beds against 1.03 and 0.93 for the other two, one seed of \MoonRoughRFT reads 1.45 on flat 3D RFT soil against 1.02 and 0.94, while the three seeds of \EarthFlatRigid on the MPM combined beds read 2.14, 1.07, and 0.55, so the first two means each rest on one seed and the third changes sign relative to 1 between seeds. The low flat-bed MPM values of 0.44 for one seed of \EarthFlatRigid and 0.49 for one seed of \EarthFlatRFT each come from a single fall among the two episodes of that seed.

% Role: fall causes on soil, compact summary of the moved fall-cause table
In 3D RFT every fall on soil is a tilt termination, and no fall meets the height criterion, with 768 episodes per cell, each policy at its training gravity on the nominal lunar soil (\texttt{soil\_falls.csv}). On the combined tiles under soil the Earth-trained policies fall 480, 245, 13, and 16 times, in the order \EarthFlatRigid, \EarthFlatRFT, \EarthRoughRigid, and \EarthRoughRFT, and the Moon-trained policies 33, 29, 15, and 3 times in the same order. On flat soil only \EarthFlatRigid, with 6 falls, and \MoonFlatRFT, with one, fall at all.

% Role: soil against the rigid run of the same geometry, survival change and cost of transport
Relative to the rigid run of the same geometry (\texttt{soil\_vs\_rigid.csv}), soil changes survival by at most 0.05 on flat ground for every policy. On the combined tiles the flat-trained Moon policies gain 0.15 and 0.23 in survival on soil and spend 0.52 and 0.49 of the rigid cost of transport, whereas \EarthFlatRigid at Earth gravity on lunar soil loses 0.15 and the other policies change by at most 0.03. On soil the Earth-trained policies spend 1.49 to 2.33 times their rigid cost of transport, while the Moon-trained policies spend 0.88 to 1.21 times it on flat soil.

% Role: survival over the five soil geometries, compact summary of the moved geometry table
Over the five soil geometries in 3D RFT (\texttt{soil\_geometries.csv}), the Moon-trained policies at the Moon survive at least 0.96 of the episodes on every geometry. The Earth-trained policies on Earth sand survive fewer undulation and combined-tile episodes than flat, crater, or rock ones, with 0.32 and 0.27 against 0.87 to 0.96 for \EarthFlatRigid, 0.78 and 0.69 against 0.97 to 1.00 for \EarthFlatRFT, and 0.92 and 0.87 against 0.98 to 0.99 for \EarthRoughRFT. \EarthRoughRigid is the exception, since it survives 0.28 to 0.33 on every geometry, with seeds from 0.10 to 0.55.

% Method: pooled 95% half-width of a three-seed mean of the soil progress ratio per engine, not a power-based detectable effect
The seed spread of $\rho_v$ in \texttt{perseed\_soil.csv} bounds what $\rho_v$ can show. We summarize it by the approximate \SI{95}{\percent} half-width of a $t$ interval on a three-seed mean, 2.48 times the seed standard deviation pooled over the eight policies, which summarizes the between-seed uncertainty of $\rho_v$. The half-width is not a power-based minimum detectable effect. It is 0.28 on flat beds and 0.24 on the combined tiles in 3D RFT, 0.40 and 1.10 in MPM, 0.18 and 0.19 in SCM, and 0.09 on flat beds in CRM. Pooled over the four Moon-trained policies only, as in the main text, it is 0.37 and 0.33 in 3D RFT, 0.15 and 1.12 in MPM, 0.24 and 0.13 in SCM, and 0.13 in CRM. Because the pooled value averages the seed variances of different policies, we base any statement about a single policy on its own interval in Table~\ref{tab:rhov_intervals} below.

% Result: per-policy t intervals of the soil progress ratio, the Moon-trained policies in the held-out engines
Because the pooled half-width averages the seed variances, it understates the uncertainty of the policies with the largest seed spread. Table~\ref{tab:rhov_intervals} therefore gives each policy's seed mean of $\rho_v$ with its own \SI{95}{\percent} $t$ interval over the three seeds. For the Moon-trained policies in MPM, SCM, and CRM, no interval lies entirely below 1. Apart from \MoonFlatRFT on the MPM combined beds, whose interval spans $-0.71$ to 3.69, the lower ends of these intervals reach 0.69, for \MoonRoughRigid on the flat MPM bed and the combined SCM bed. Losses of up to about 0.3 in $\rho_v$ therefore remain possible for single policies, and with two degrees of freedom the intervals are themselves imprecise. On the combined beds at Earth gravity, however, four Earth-trained policies show losses whose intervals exclude 1, namely \EarthFlatRigid and \EarthFlatRFT in 3D RFT and \EarthRoughRigid and \EarthRoughRFT in SCM, so the statements above about losses not excluded refer to the Moon-trained rows.
% Source: tables/tab_rhov_intervals.tex, written by make_tab_keycells.py write_rhov_intervals (LuPNT 8a666bcfa) from the per-seed values of tab_perseed_soil (t(0.975,2) = 4.303, mean +- t s/sqrt(3)); the same function prints the pooled half-widths (all 8: 0.275 0.236 0.395 1.098 0.179 0.190 0.091; Moon-trained 4: 0.370 0.326 0.146 1.118 0.241 0.128 0.126), equal to review_2026-10-24 C16. Moon-trained held-out lower ends: Moon-Flat-Rigid 0.97/0.91/1.02/0.93/1.06, Moon-Flat-RFT 0.90/-0.71/0.76/0.99/0.89, Moon-Rough-Rigid 0.69/0.72/0.75/0.69/0.95, Moon-Rough-RFT 0.87/0.70/1.00/0.96/0.85 (MPM flat, MPM comb, SCM flat, SCM comb, CRM).

\begin{table*}[!tb]
  \centering\scriptsize
  \caption{\textbf{Soil progress ratio with per-policy intervals.} Seed mean of $\rho_v$ of Table~\ref{tab:soil_results} with its \SI{95}{\percent} Student $t$ interval over the three seeds of \texttt{perseed\_soil.csv} in brackets. With two degrees of freedom the interval spans $\pm 2.48$ seed standard deviations. Every engine uses the nominal lunar soil, so the Earth rows are at Earth gravity on lunar soil. Columns are 3D resistive force theory (3D RFT, written RFT in policy names), material point method (MPM), soil contact model (SCM), and continuous representation model (CRM) beds, flat or combined (Comb.).}
  \label{tab:rhov_intervals}
  \setlength{\tabcolsep}{3pt}
  \begin{tabular*}{\linewidth}{@{\extracolsep{\fill}}|l|cc|cc|cc|c|}
\hline
& \multicolumn{2}{c|}{\textbf{3D RFT}} & \multicolumn{2}{c|}{\textbf{MPM}} & \multicolumn{2}{c|}{\textbf{SCM}} & \textbf{CRM} \\
\hdrstrut \textbf{Policy} & Flat & Comb. & Flat & Comb. & Flat & Comb. & Flat \\
\hline\hline
\EarthFlatRigid & 1.04\,{\tiny[0.91,1.17]} & 0.80\,{\tiny[0.75,0.86]} & 0.76\,{\tiny[0.06,1.46]} & 1.26\,{\tiny[$-$0.76,3.27]} & 0.96\,{\tiny[0.90,1.01]} & 1.37\,{\tiny[1.05,1.70]} & 0.98\,{\tiny[0.96,1.00]} \\
\EarthFlatRFT & 1.07\,{\tiny[1.03,1.10]} & 0.91\,{\tiny[0.83,0.98]} & 0.86\,{\tiny[0.05,1.68]} & 1.24\,{\tiny[0.49,1.99]} & 1.05\,{\tiny[1.04,1.06]} & 1.18\,{\tiny[0.84,1.53]} & 1.07\,{\tiny[1.04,1.09]} \\
\EarthRoughRigid & 1.05\,{\tiny[0.90,1.20]} & 0.93\,{\tiny[0.86,1.01]} & 1.03\,{\tiny[1.01,1.06]} & 1.04\,{\tiny[0.93,1.15]} & 1.06\,{\tiny[1.01,1.11]} & 0.91\,{\tiny[0.88,0.94]} & 1.03\,{\tiny[1.01,1.05]} \\
\EarthRoughRFT & 0.94\,{\tiny[0.80,1.07]} & 0.93\,{\tiny[0.84,1.01]} & 1.06\,{\tiny[0.96,1.16]} & 1.02\,{\tiny[0.92,1.12]} & 1.03\,{\tiny[0.89,1.16]} & 0.95\,{\tiny[0.91,0.99]} & 1.07\,{\tiny[1.04,1.11]} \\
\hline
\MoonFlatRigid & 1.04\,{\tiny[1.04,1.04]} & 1.35\,{\tiny[1.25,1.44]} & 1.00\,{\tiny[0.97,1.04]} & 1.05\,{\tiny[0.91,1.18]} & 1.04\,{\tiny[1.02,1.06]} & 1.01\,{\tiny[0.93,1.09]} & 1.09\,{\tiny[1.06,1.11]} \\
\MoonFlatRFT & 1.01\,{\tiny[0.85,1.16]} & 1.46\,{\tiny[1.00,1.93]} & 1.00\,{\tiny[0.90,1.11]} & 1.49\,{\tiny[$-$0.71,3.69]} & 0.96\,{\tiny[0.76,1.16]} & 1.05\,{\tiny[0.99,1.11]} & 0.97\,{\tiny[0.89,1.05]} \\
\MoonRoughRigid & 0.89\,{\tiny[0.63,1.14]} & 0.87\,{\tiny[0.47,1.27]} & 0.92\,{\tiny[0.69,1.15]} & 0.92\,{\tiny[0.72,1.11]} & 1.19\,{\tiny[0.75,1.62]} & 0.92\,{\tiny[0.69,1.15]} & 1.13\,{\tiny[0.95,1.31]} \\
\MoonRoughRFT & 1.14\,{\tiny[0.46,1.82]} & 1.08\,{\tiny[0.87,1.29]} & 1.01\,{\tiny[0.87,1.16]} & 1.03\,{\tiny[0.70,1.35]} & 1.05\,{\tiny[1.00,1.09]} & 0.99\,{\tiny[0.96,1.03]} & 1.01\,{\tiny[0.85,1.16]} \\
\hline
\end{tabular*}

\end{table*}

% Method: MPM off-bed exclusions
Two MPM combined-bed episodes leave the bed and are excluded, one of \EarthFlatRFT and one of \MoonFlatRigid. Including them changes a seed-mean $\rho_v$ by at most 0.01 and a seed-mean progress difference between soil and rigid ground by at most \SI{0.09}{\metre}.
% Source: review_2026-10-24/summary.md C16: pooled between-seed SD (root mean of per-policy variances, all 8 policies) x t(0.975,2)/sqrt(3) = 2.48: 3D RFT Flat 0.275, Comb. 0.236; MPM Flat 0.395, Comb. 1.098; SCM Flat 0.179, Comb. 0.190; CRM Flat 0.091. Off-bed (> 5 % of frames): Earth-Flat-RFT s42, Moon-Flat-Rigid s43; with vs without exclusion, seed-mean rho_v 1.24 vs 1.23 and 1.05 vs 1.06, seed-mean delta progress 1.03 vs 0.94 m and 0.25 vs 0.29 m.

\subsection{Calibrated Lunar Soils}
\label{sec:res_lunarsoils-supp}
% Source: output/RAL_Humanoid/eval_best/<policy>_s<seed>__SoilFlat*_Moon.json and __SoilTilesCombined*_Moon.json, tabular generated by projects/guillemc/RAL_Humanoid/figures/make_tab_keycells.py
\begin{table*}[!tb]
  \centering\footnotesize
  \caption{\textbf{Survival on the three calibrated lunar soils and one compacted stress test at the Moon.} 3D resistive force theory (3D RFT, written RFT in policy names) soil over a \SI{0.12}{\metre} layer, flat or following the 32 held-out combined tiles, 256 episodes per cell, mean over seeds. Loose, rim, and nominal match bootprint stiffness for the G1 foot. Compacted is twice as stiff as the stiffest lunar training soil.}
  \label{tab:lunar_soils}
  \setlength{\tabcolsep}{3pt}
  \begin{tabular*}{\linewidth}{@{\extracolsep{\fill}}|l|cccc|cccc|}
\hline
& \multicolumn{4}{c|}{\textbf{Flat soil}} & \multicolumn{4}{c|}{\textbf{Combined tiles under soil}} \\
\hdrstrut \textbf{Policy} & Loose & Rim & Nominal & Compacted & Loose & Rim & Nominal & Compacted \\
\hline\hline
\EarthFlatRigid & 0.99\,{\tiny[0.98,1.00]} & 0.95\,{\tiny[0.88,1.00]} & 0.92\,{\tiny[0.84,0.98]} & 0.87\,{\tiny[0.79,0.93]} & 0.89\,{\tiny[0.83,0.92]} & 0.86\,{\tiny[0.79,0.92]} & 0.80\,{\tiny[0.70,0.88]} & 0.69\,{\tiny[0.58,0.75]} \\
\EarthFlatRFT & 0.99\,{\tiny[0.98,1.00]} & 0.98\,{\tiny[0.93,1.00]} & 0.96\,{\tiny[0.93,0.98]} & 0.88\,{\tiny[0.83,0.94]} & 0.86\,{\tiny[0.77,0.93]} & 0.83\,{\tiny[0.73,0.91]} & 0.76\,{\tiny[0.63,0.89]} & 0.67\,{\tiny[0.55,0.80]} \\
\EarthRoughRigid & 0.97\,{\tiny[0.91,1.00]} & 0.96\,{\tiny[0.92,0.99]} & 0.96\,{\tiny[0.93,0.99]} & 0.86\,{\tiny[0.72,0.94]} & 0.85\,{\tiny[0.82,0.87]} & 0.85\,{\tiny[0.80,0.87]} & 0.77\,{\tiny[0.76,0.78]} & 0.72\,{\tiny[0.64,0.78]} \\
\EarthRoughRFT & 0.90\,{\tiny[0.77,1.00]} & 0.85\,{\tiny[0.67,1.00]} & 0.84\,{\tiny[0.68,1.00]} & 0.80\,{\tiny[0.62,0.98]} & 0.83\,{\tiny[0.63,0.96]} & 0.81\,{\tiny[0.59,0.96]} & 0.78\,{\tiny[0.52,0.96]} & 0.71\,{\tiny[0.50,0.92]} \\
\hline
\MoonFlatRigid & 1.00 & 1.00 & 1.00 & 1.00 & 0.93\,{\tiny[0.91,0.96]} & 0.95\,{\tiny[0.93,0.98]} & 0.96\,{\tiny[0.94,0.98]} & 0.94\,{\tiny[0.92,0.95]} \\
\MoonFlatRFT & 1.00 & 1.00\,{\tiny[0.99,1.00]} & 1.00 & 1.00 & 0.95\,{\tiny[0.93,0.95]} & 0.95\,{\tiny[0.94,0.96]} & 0.96\,{\tiny[0.94,0.98]} & 0.95\,{\tiny[0.94,0.96]} \\
\MoonRoughRigid & 1.00 & 1.00 & 1.00 & 1.00 & 0.94\,{\tiny[0.86,0.99]} & 0.97\,{\tiny[0.95,0.99]} & 0.98\,{\tiny[0.96,0.99]} & 0.98\,{\tiny[0.95,1.00]} \\
\MoonRoughRFT & 1.00 & 1.00 & 1.00 & 1.00 & 0.99\,{\tiny[0.98,0.99]} & 0.99\,{\tiny[0.98,1.00]} & 1.00\,{\tiny[0.99,1.00]} & 0.99\,{\tiny[0.99,1.00]} \\
\hline
\FtFlatRigid & 1.00 & 1.00 & 1.00 & 1.00 & 0.95\,{\tiny[0.93,0.98]} & 0.96\,{\tiny[0.94,0.98]} & 0.96\,{\tiny[0.94,0.99]} & 0.95\,{\tiny[0.91,0.96]} \\
\FtRoughRFT & 1.00 & 1.00 & 1.00 & 1.00 & 0.99\,{\tiny[0.99,1.00]} & 0.99\,{\tiny[0.99,1.00]} & 1.00 & 0.99\,{\tiny[0.98,1.00]} \\
\JointFlatRigid & 1.00 & 1.00 & 1.00 & 1.00 & 0.82\,{\tiny[0.77,0.90]} & 0.86\,{\tiny[0.83,0.91]} & 0.86\,{\tiny[0.82,0.89]} & 0.86\,{\tiny[0.83,0.88]} \\
\JointRoughRFT & 1.00 & 1.00 & 1.00 & 1.00 & 0.99\,{\tiny[0.98,1.00]} & 0.99\,{\tiny[0.99,1.00]} & 1.00 & 1.00 \\
\hline
\end{tabular*}

\end{table*}

% Result: lunar soil stiffness hardly matters for the lunar-trained policies, Earth policies at the Moon survive more on softer soil
Soil stiffness hardly matters to the Moon-trained and fine-tuned policies (Table~\ref{tab:lunar_soils}). Across the three calibrated lunar soils and the compacted stress-test soil, which lies outside the training range, they keep at least 0.93 of the tile episodes and nearly all flat ones. \JointFlatRigid keeps all flat episodes but only 0.82 to 0.86 of the tile episodes. The Earth-trained policies at the Moon are the exception. They survive more often on softer soil, with 0.99 on loose flat soil against 0.87 on compacted soil for \EarthFlatRigid.

% Result: soil progress ratio on the four test soils rises with stiffness
For the Moon-trained, fine-tuned, and gravity-randomized policies on flat soil, $\rho_v$ rises monotonically from loose to compacted soil, with seed means of 0.77 to 1.07 on loose soil and 1.01 to 1.22 on compacted soil, so they progress faster on stiffer soil relative to rigid ground. On the combined tiles the order is the same for every one of them but \JointFlatRigid, whose $\rho_v$ of 1.53 to 1.69 comes from frequent falls on the rigid tiles. The Earth-trained policies at the Moon read 1.34 to 2.32 on every soil and geometry for the same reason.

% Caveat: 3D RFT soil is easier than rigid ground for policies tested out of distribution
3D RFT soil is easier than rigid ground at the Moon for the Earth-trained policies. Across the three calibrated lunar soils and the compacted stress test, their seed-mean survival on the combined tiles exceeds that on the rigid tiles by 0.17 to 0.45, and on flat soil \EarthFlatRigid, \EarthFlatRFT, and \EarthRoughRigid exceed their flat rigid survival by 0.12 to 0.35. The policies trained for lunar gravity survive about as often on flat rigid ground as on flat soil at the Moon, so on flat ground the effect is confined to policies tested outside their training gravity. On the combined tiles the flat-trained Moon policies also gain on soil (Section~\ref{sec:res_soil_seed-supp}).
% Source: review_2026-10-24/summary.md C16 (three calibrated soils and the compacted stress test, rho_v seed means, forward profiles) and C15 (soil cells more than 0.10 above rigid). Flat loose rho_v lunar-capable 0.77 (MRR) to 1.07 (MRT), compacted 1.01 (MRR) to 1.22 (MRT); comb. EarthMoon-Flat-Rigid 1.53/1.64/1.69/1.61; Earth-trained min 1.34 (ERR flat loose), max 2.32 (EFT flat nominal). Easier-than-rigid deltas: comb. EFR +0.25..+0.45, EFT +0.24..+0.42, ERR +0.29..+0.42, ERT +0.17..+0.29; flat EFR +0.23..+0.35, EFT +0.21..+0.33, ERR +0.12..+0.23 (ERT flat exceeds 0.10 only on loose, +0.14).

\subsection{Seed Sign Rule}
\label{sec:signrule-supp}
% Method: the sign rule applied to every stated survival comparison, a seed-replication heuristic
Tables~\ref{tab:signrule-supp}, \ref{tab:signrule-c3}, and~\ref{tab:signrule-withdrawn} apply the sign rule of Section~\ref{sec:protocol} to 54 survival comparisons, with the per-seed Newcombe interval on each difference over 256 episodes per seed. \Cref{tab:signrule-supp} and \Cref{tab:signrule-c3} list the 42 comparisons that the main text and this supplement state, with the 13 comparisons of the recommended policy against its rigid control in the second table, and all of them meet the rule. \Cref{tab:signrule-withdrawn} lists the other 12, comparisons that the text states only in part or not at all. Ten of them do not meet the rule, and two on the \SI{26}{\degree} grade meet it but are no longer stated. We keep all 54 in the test so that the count of comparisons, and with it the bound on chance passes, stays complete. The rule is a seed-replication heuristic and not a test with a controlled error rate. If seed effects are symmetric about zero, all three seeds share one sign with probability 1/4, and the interval condition can only lower this, so a comparison without a true difference passes at a rate of at most 1/4. Without a correction for multiple comparisons, this bound allows up to about 14 chance passes in expectation over the 54 comparisons. A comparison that spans several cells is shown by its weakest cell. Where the rule is not met, the text states only the part that passes, for example the cells or seeds in which it holds, and \Cref{tab:signrule-withdrawn} marks the row as narrowed, or as withdrawn where the text states none of it. In all three tables $>$ and $<$ mark a stated difference, $\approx$ no detected difference in either direction, and $\not>$ and $\not<$ no detected gain or loss. These rows mean only that the rule does not license the excluded difference, not that the policies are equivalent. Rows on the 26 and \SI{30}{\degree} grades are scored on traversal $T$ (Section~\ref{sec:res_traversal-supp}).

% Method: tile-cluster check of the sign rule on the combined-tile cells
Because the episodes of a cell share tiles, we re-ran the four combined-tile cells (rigid and soil, at both gravities) with the tile of each episode logged. All four cells use the same 32 tiles with one episode per command profile on each tile, and every policy sees the same tile and profile assignment. A per-seed bootstrap over the 32 tiles, resampling the same tiles for both policies of a comparison, gives a survival variance a median 1.12 times the binomial variance (0.47 to 2.32 over policies and seeds). The paired intervals of a difference are usually narrower than the Newcombe intervals, because a hard tile tends to be hard for both policies. With these intervals in place of the Newcombe intervals, none of the 27 comparisons of Tables~\ref{tab:signrule-supp} and~\ref{tab:signrule-withdrawn} that involve these cells changes its verdict. The re-run itself agrees with the original evaluations within binomial noise, with a mean absolute change in survival of 0.013 and at most 0.035 in seed means.
% Source: output/RAL_Humanoid/review_2026-10-26/c11_tiles/c11_summary.md, c11_signrule_cluster.csv, c11_cells.csv (144 evals, 12 policies x 3 seeds x TilesRigidCombined/SoilTilesCombined x Earth/Moon, tile ids per episode); script output/.../c11_tiles/script_snapshot/c11_cluster.py with sign_rule_table.py at LuPNT 2ccc2e119 (reproduces the 41 table verdicts, 31 met). Design effect per cell median: rigid Earth 1.32, rigid Moon 1.09, soil Earth 1.24, soil Moon 1.00. Largest rerun gap Earth-Rough-RFT s44 rigid Moon 65 vs 96 survivors (z 2.95), 2 of 144 runs |z|>1.96 vs 7.2 expected.

\begin{table*}[!tb]
  \centering\scriptsize
  \caption{\textbf{Seed sign rule for each stated comparison.} The rule is a seed-replication heuristic with approximate per-seed intervals. Survival difference per seed for the weakest cell of each comparison, and whether the rule holds, that is, all three seeds share the sign and each per-seed Newcombe \SI{95}{\percent} interval excludes zero, or, for a stated non-detection, that the rule does not license the excluded difference. Candidate claims that the text withdrew or narrowed are in \Cref{tab:signrule-withdrawn}. RFT in the comparisons is 3D resistive force theory (3D RFT).}
  \label{tab:signrule-supp}
  \begin{tabular*}{\linewidth}{@{\extracolsep{\fill}}p{0.70\textwidth}cc}
\hline
\textbf{Comparison (claim, weakest sub-comparison)} & \textbf{Difference per seed} & \textbf{Rule met} \\
\hline\hline
$\approx$~no detected difference between \EarthFlatRigid and \EarthRoughRigid on flat, crater, and rock tiles (Earth) (3 cells), weakest: \EarthRoughRigid vs \EarthFlatRigid, rocks @Earth & $+0.02$, $+0.02$, $+0.02$ & yes \\
$>$~\EarthFlatRigid falls to about half survival on undulations and combined tiles (Earth), vs \EarthRoughRigid (2 cells), weakest: \EarthRoughRigid vs \EarthFlatRigid, undulations @Earth & $+0.47$, $+0.40$, $+0.39$ & yes \\
$<$~\MoonFlatRigid loses less on the same tiles at the Moon than \EarthFlatRigid at Earth (rough-minus-flat gap of the rigid pair smaller at the Moon) (2 cells), weakest: undulations: (MoonRough-MoonFlat) @Moon minus (EarthRough-EarthFlat) @Earth & $-0.30$, $-0.21$, $-0.31$ & yes \\
$\approx$~no detected difference between \MoonFlatRigid and \MoonRoughRigid on craters and rocks at the Moon (2 cells), weakest: \MoonRoughRigid vs \MoonFlatRigid, craters @Moon & $+0.00$, $+0.02$, $+0.00$ & yes \\
$<$~\EarthRoughRigid loses survival at the Moon on every tile family (Moon vs Earth cell) (5 cells), weakest: \EarthRoughRigid, rocks: Moon minus Earth & $-0.26$, $-0.18$, $-0.29$ & yes \\
$<$~\MoonFlatRigid falls at Earth on every tile family (Earth vs Moon cell) (5 cells), weakest: \MoonFlatRigid, combined tiles: Earth minus Moon & $-0.77$, $-0.79$, $-0.86$ & yes \\
$<$~\MoonRoughRigid falls at Earth on every tile family (Earth vs Moon cell) (5 cells), weakest: \MoonRoughRigid, combined tiles: Earth minus Moon & $-0.99$, $-0.98$, $-0.99$ & yes \\
$\not>$~no detected gain of \EarthRoughRigid over \EarthFlatRigid on the combined tiles at the Moon, weakest: \EarthRoughRigid vs \EarthFlatRigid, combined tiles @Moon & $-0.18$, $+0.21$, $-0.07$ & yes \\
$>$~RFT-trained \EarthRoughRFT survives flat sand and combined tiles under sand more often than \EarthRoughRigid (2 cells), weakest: \EarthRoughRFT vs \EarthRoughRigid, combined tiles soil @Earth & $+0.66$, $+0.75$, $+0.30$ & yes \\
$\not<$~no detected survival loss from RFT training on rigid flat ground or rigid tiles (\EarthFlatRFT vs \EarthFlatRigid at Earth) (5 cells), weakest: \EarthFlatRFT vs \EarthFlatRigid, flat @Earth & $-0.01$, $+0.00$, $+0.00$ & yes \\
$\not<$~no detected survival loss from RFT training on rigid flat ground or rigid tiles (\EarthRoughRFT vs \EarthRoughRigid at Earth) (5 cells), weakest: \EarthRoughRFT vs \EarthRoughRigid, flat @Earth & $+0.00$, $+0.00$, $+0.00$ & yes \\
$\not<$~no detected survival loss from RFT training on rigid flat ground or rigid tiles (\MoonFlatRFT vs \MoonFlatRigid at Moon) (5 cells), weakest: \MoonFlatRFT vs \MoonFlatRigid, flat @Moon & $+0.00$, $-0.01$, $+0.00$ & yes \\
$\not<$~no detected survival loss from RFT training on rigid flat ground or rigid tiles (\MoonRoughRFT vs \MoonRoughRigid at Moon) (5 cells), weakest: \MoonRoughRFT vs \MoonRoughRigid, flat @Moon & $+0.00$, $+0.00$, $-0.14$ & yes \\
$<$~RFT training lowers survival on the 20$^\circ$ uphill grade for the flat pair (Earth), weakest: \EarthFlatRFT vs \EarthFlatRigid, 20$^\circ$ uphill @Earth & $-0.42$, $-0.27$, $-0.14$ & yes \\
$>$~on lunar soil at Earth gravity the flat pair keeps a survival gain on tiles (\EarthFlatRFT above \EarthFlatRigid), weakest: \EarthFlatRFT vs \EarthFlatRigid, combined tiles lunar soil @Earth & $+0.28$, $+0.32$, $+0.32$ & yes \\
$\approx$~on lunar soil at Earth gravity no detected survival difference on tiles for the rough pair (\EarthRoughRFT vs \EarthRoughRigid), weakest: \EarthRoughRFT vs \EarthRoughRigid, combined tiles lunar soil @Earth & $+0.01$, $-0.02$, $+0.00$ & yes \\
$\approx$~at the Moon no detected survival difference between \MoonFlatRFT and \MoonFlatRigid on any soil cell (5 cells), weakest: \MoonFlatRFT vs \MoonFlatRigid, flat soil @Moon & $+0.00$, $-0.00$, $+0.00$ & yes \\
$\approx$~at the Moon no detected survival difference between \MoonRoughRFT and \MoonRoughRigid on any soil cell (5 cells), weakest: \MoonRoughRFT vs \MoonRoughRigid, flat soil @Moon & $+0.00$, $+0.00$, $+0.00$ & yes \\
$>$~\JointFlatRigid keeps most of its Earth survival while \MoonFlatRigid and \FtFlatRigid walk in no Earth episode (2 cells), weakest: \JointFlatRigid vs \MoonFlatRigid, flat @Earth & $+0.96$, $+0.88$, $+0.91$ & yes \\
$\approx$~on rigid combined tiles at the Moon no detected survival difference among the three rough RFT strategies (3 cells), weakest: \MoonRoughRFT vs \JointRoughRFT, combined tiles @Moon & $-0.00$, $-0.02$, $-0.04$ & yes \\
$>$~only \JointRoughRFT keeps walking at Earth (rigid combined tiles) (2 cells), weakest: \JointRoughRFT vs \MoonRoughRFT, combined tiles @Earth & $+0.96$, $+0.94$, $+0.96$ & yes \\
$>$~\EarthFlatRigid survives its 3D RFT soil cells more often than the rigid ones at the Moon (2 cells), weakest: \EarthFlatRigid, flat soil minus flat rigid @Moon & $+0.29$, $+0.28$, $+0.29$ & yes \\
$>$~\EarthFlatRFT survives its 3D RFT soil cells more often than the rigid ones at the Moon (2 cells), weakest: \EarthFlatRFT, flat soil minus flat rigid @Moon & $+0.30$, $+0.35$, $+0.23$ & yes \\
$>$~\EarthRoughRigid survives its 3D RFT soil cells more often than the rigid ones at the Moon (2 cells), weakest: \EarthRoughRigid, flat soil minus flat rigid @Moon & $+0.15$, $+0.19$, $+0.31$ & yes \\
$>$~\EarthFlatRigid at the Moon survives 0.99 on loose flat soil against 0.87 on compacted soil, weakest: \EarthFlatRigid, loose minus compacted flat soil @Moon & $+0.07$, $+0.19$, $+0.10$ & yes \\
$>$~rough training lifts survival on combined tiles from 0.53 to 0.98 at Earth (\EarthRoughRigid vs \EarthFlatRigid), weakest: \EarthRoughRigid vs \EarthFlatRigid, combined tiles @Earth & $+0.48$, $+0.48$, $+0.42$ & yes \\
$>$~rough training lifts survival on combined tiles from 0.80 to 0.99 at the Moon (\MoonRoughRigid vs \MoonFlatRigid), weakest: \MoonRoughRigid vs \MoonFlatRigid, combined tiles @Moon & $+0.22$, $+0.20$, $+0.14$ & yes \\
$\approx$~the combined-tile rough gain is detected only at the training gravity (Moon-trained pair at Earth: no detected gain), weakest: \MoonRoughRigid vs \MoonFlatRigid, combined tiles @Earth & $+0.00$, $+0.00$, $+0.00$ & yes \\
$\not<$~at the Moon on every cell up to 20$^\circ$ no detected loss of gravity-randomized \JointRoughRFT against any lunar-trained policy (36 cells), weakest: \JointRoughRFT vs \MoonFlatRigid, flat @Moon & $+0.00$, $+0.00$, $+0.00$ & yes \\
\hline
\end{tabular*}

\end{table*}

\begin{table*}[!tb]
  \centering\scriptsize
  \caption{\textbf{Sign-rule comparisons of the recommended policy against its rigid control.} The 13 comparisons of \JointRoughRFT against \JointRoughRigid, with the same columns and scoring as Table~\ref{tab:signrule-supp}. RFT is 3D resistive force theory (3D RFT).}
  \label{tab:signrule-c3}
  \begin{tabular*}{\linewidth}{@{\extracolsep{\fill}}p{0.70\textwidth}cc}
\hline
\textbf{Comparison (claim, weakest sub-comparison)} & \textbf{Difference per seed} & \textbf{Rule met} \\
\hline\hline
$>$~rigid control: \JointRoughRFT survives Earth flat sand more often than \JointRoughRigid, weakest: \JointRoughRFT vs \JointRoughRigid, flat soil @Earth & $+0.91$, $+0.96$, $+0.70$ & yes \\
$>$~rigid control: \JointRoughRFT survives Earth combined tiles under sand more often than \JointRoughRigid, weakest: \JointRoughRFT vs \JointRoughRigid, combined tiles soil @Earth & $+0.72$, $+0.86$, $+0.64$ & yes \\
$>$~rigid control: \JointRoughRFT survives combined tiles on lunar soil at Earth gravity more often than \JointRoughRigid, weakest: \JointRoughRFT vs \JointRoughRigid, combined tiles lunar soil @Earth & $+0.06$, $+0.07$, $+0.05$ & yes \\
$\approx$~rigid control: no detected difference between \JointRoughRFT and \JointRoughRigid on flat at Earth, weakest: \JointRoughRFT vs \JointRoughRigid, flat @Earth & $+0.00$, $-0.02$, $+0.00$ & yes \\
$\approx$~rigid control: no detected difference between \JointRoughRFT and \JointRoughRigid on combined tiles at Earth, weakest: \JointRoughRFT vs \JointRoughRigid, combined tiles @Earth & $+0.02$, $-0.04$, $-0.00$ & yes \\
$\approx$~rigid control: no detected difference between \JointRoughRFT and \JointRoughRigid on 20$^\circ$ uphill at Earth, weakest: \JointRoughRFT vs \JointRoughRigid, 20$^\circ$ uphill @Earth & $+0.00$, $+0.00$, $+0.04$ & yes \\
$\approx$~rigid control: no detected difference between \JointRoughRFT and \JointRoughRigid on 20$^\circ$ downhill at Earth, weakest: \JointRoughRFT vs \JointRoughRigid, 20$^\circ$ downhill @Earth & $-0.02$, $-0.02$, $+0.16$ & yes \\
$\approx$~rigid control: no detected difference between \JointRoughRFT and \JointRoughRigid on flat at Moon, weakest: \JointRoughRFT vs \JointRoughRigid, flat @Moon & $+0.00$, $+0.00$, $+0.00$ & yes \\
$\approx$~rigid control: no detected difference between \JointRoughRFT and \JointRoughRigid on combined tiles at Moon, weakest: \JointRoughRFT vs \JointRoughRigid, combined tiles @Moon & $+0.01$, $+0.00$, $+0.00$ & yes \\
$\approx$~rigid control: no detected difference between \JointRoughRFT and \JointRoughRigid on 20$^\circ$ uphill at Moon, weakest: \JointRoughRFT vs \JointRoughRigid, 20$^\circ$ uphill @Moon & $-0.00$, $+0.00$, $-0.03$ & yes \\
$\approx$~rigid control: no detected difference between \JointRoughRFT and \JointRoughRigid on 20$^\circ$ downhill at Moon, weakest: \JointRoughRFT vs \JointRoughRigid, 20$^\circ$ downhill @Moon & $+0.13$, $-0.01$, $+0.05$ & yes \\
$\approx$~rigid control: no detected difference between \JointRoughRFT and \JointRoughRigid on flat soil at Moon, weakest: \JointRoughRFT vs \JointRoughRigid, flat soil @Moon & $+0.00$, $+0.00$, $+0.00$ & yes \\
$\approx$~rigid control: no detected difference between \JointRoughRFT and \JointRoughRigid on combined tiles soil at Moon, weakest: \JointRoughRFT vs \JointRoughRigid, combined tiles soil @Moon & $+0.01$, $+0.01$, $-0.00$ & yes \\
\hline
\end{tabular*}

\end{table*}

\begin{table*}[!tb]
  \centering\scriptsize
  \caption{\textbf{Withdrawn and narrowed candidate claims.} Comparisons we tested that the text states only in part (narrowed) or not at all (withdrawn), scored as in Table~\ref{tab:signrule-supp}, with traversal in place of survival on the 26 and \SI{30}{\degree} grades. RFT is 3D resistive force theory (3D RFT).}
  \label{tab:signrule-withdrawn}
  \begin{tabular*}{\linewidth}{@{\extracolsep{\fill}}p{0.62\textwidth}ccc}
\hline
\textbf{Comparison (claim, weakest sub-comparison)} & \textbf{Difference per seed} & \textbf{Rule met} & \textbf{In text} \\
\hline\hline
$<$~\EarthFlatRigid loses survival at the Moon on every tile family (Moon vs Earth cell) (5 cells), weakest: \EarthFlatRigid, combined tiles: Moon minus Earth & $-0.02$, $-0.17$, $-0.06$ & no & narrowed \\
$>$~at the Moon the Earth rough-trained policies survive the 20$^\circ$ uphill grade more often than the flat-trained ones (2 cells), weakest: \EarthRoughRigid vs \EarthFlatRigid, 20$^\circ$ uphill @Moon & $+0.17$, $+0.68$, $+0.03$ & no & withdrawn \\
$>$~at Earth, rough-trained policies hold survival to 20$^\circ$ while flat-trained decline (rough above flat at 20$^\circ$, both directions, both pairs) (4 cells), weakest: \EarthRoughRigid vs \EarthFlatRigid, 20$^\circ$ downhill @Earth & $+0.62$, $+0.91$, $-0.06$ & no & narrowed \\
$<$~at Earth, flat-trained policies already decline from 10$^\circ$ (10$^\circ$ survival below flat ground) (4 cells), weakest: \EarthFlatRFT, 10$^\circ$ uphill @Earth minus flat & $+0.01$, $+0.00$, $+0.00$ & no & withdrawn \\
$<$~at the Moon, flat-trained policies fall to near zero by 26$^\circ$ uphill (26$^\circ$ minus flat ground) (2 cells)$^\dagger$, weakest: \MoonFlatRFT, 26$^\circ$ up minus flat @Moon & $-1.00$, $-0.99$, $-1.00$ & yes & withdrawn \\
$>$~at the Moon, rough-trained lunar policies decline more gradually: above flat-trained at 26$^\circ$ uphill (2 cells)$^\dagger$, weakest: \MoonRoughRFT vs \MoonFlatRFT, 26$^\circ$ uphill @Moon & $+0.21$, $+0.14$, $+0.07$ & yes & withdrawn \\
$>$~at 30$^\circ$ the best result is \FtRoughRFT downhill at the Moon (above every other policy) (11 cells)$^\dagger$, weakest: \FtRoughRFT vs \MoonRoughRigid, 30$^\circ$ downhill @Moon & $+0.10$, $+0.09$, $+0.00$ & no & withdrawn \\
$>$~RFT-trained \EarthFlatRFT survives flat sand and combined tiles under sand more often than \EarthFlatRigid (2 cells), weakest: \EarthFlatRFT vs \EarthFlatRigid, flat soil @Earth & $+0.08$, $+0.03$, $+0.01$ & no & narrowed \\
$<$~RFT training lowers survival on the 20$^\circ$ uphill grade for the flat pair (Moon), weakest: \MoonFlatRFT vs \MoonFlatRigid, 20$^\circ$ uphill @Moon & $+0.16$, $+0.10$, $-0.24$ & no & withdrawn \\
$>$~\EarthRoughRFT survives its 3D RFT soil cells more often than the rigid ones at the Moon (2 cells), weakest: \EarthRoughRFT, flat soil minus flat rigid @Moon & $+0.05$, $+0.06$, $+0.14$ & no & narrowed \\
$>$~\MoonFlatRigid survives its 3D RFT soil cells more often than the rigid ones at the Moon (2 cells), weakest: \MoonFlatRigid, flat soil minus flat rigid @Moon & $+0.00$, $+0.00$, $+0.00$ & no & narrowed \\
$>$~\MoonFlatRFT survives its 3D RFT soil cells more often than the rigid ones at the Moon (2 cells), weakest: \MoonFlatRFT, flat soil minus flat rigid @Moon & $+0.00$, $+0.01$, $+0.00$ & no & narrowed \\
\hline
\multicolumn{4}{@{}p{\linewidth}@{}}{$^\dagger$Traversal $T$ on the \SI{26}{\degree} and \SI{30}{\degree} grades, near or above the \SI{26.6}{\degree} friction-cone limit, where a robot can slide while upright. $T$ counts an episode that survives with positive progress along the command below 1.5 times its flat-ground progress. Other rows use survival $S$.} \\
\end{tabular*}

\end{table*}
% Source: tables/tab_signrule.tex, tab_signrule_c3.tex (the 13 C3 rows, split off 2026-10-27 because the first table overflowed the page) and tab_signrule_withdrawn.tex written by LuPNT-private projects/guillemc/RAL_Humanoid/figures/sign_rule_table.py (LuPNT 65f7fc65c, 2026-10-27: claim ids in NARROWED go to the second table; the script asserts that every claim outside it passes; C3 rigid-control rows C3-a..c and C3-<cell> x10 added, per-seed numbers reproduce eval_c3_joint_rough_rigid/c3_analysis_stdout.txt). 54 rows, 44 pass and 10 fail; stated 42 (all pass, 29 + 13 rigid control), narrowed 6 (T5-EFR, T9, G1-flat, G6-ERT, G6-MFR, G6-MFT, all fail), withdrawn 6 (T8, T10, T13, G3-Moon fail; T11, T12 pass, on traversal T at 26 deg). Chance-pass bound 54/4 = 13.5. TODO: recheck the counts if the tables are regenerated again.

\subsection{Rough versus Soil Training per Seed}
\label{sec:roughvssoil-supp}
% Comparison: per-seed test that rough training matters more than soil training
Table~\ref{tab:roughvssoil} tests the main-text statement that rough training matters more than soil training on tiles and slopes. The rough gain is the survival of the rough-trained rigid policy minus that of the flat-trained rigid policy, the soil gain is the survival of the flat-trained 3D RFT policy minus that of the flat-trained rigid policy, and both are measured on the same cell at the training gravity. Their difference equals the survival of the rough-trained rigid policy minus that of the flat-trained 3D RFT policy, and the sign rule of Section~\ref{sec:protocol} applies to it with the per-seed Newcombe interval. The statement holds in all three seeds on the combined rigid tiles and on the \SI{20}{\degree} uphill grade at both gravities, 4 of 12 cells. On the \SI{30}{\degree} grades the difference is scored on traversal $T$, because a robot can slide there while upright (Section~\ref{sec:res_traversal-supp}). It fails on the combined tiles under Earth sand, where the soil gain exceeds the rough gain in all three seeds, and under the lunar soil at the Moon, where the sign changes between seeds. On the \SI{20}{\degree} downhill grade it fails in one seed at each gravity, in which the rough-trained rigid policy survives less than the flat-trained one. At \SI{30}{\degree} at Earth every policy falls. At the Moon the difference in traversal is zero in one seed on the \SI{30}{\degree} uphill grade and in two seeds on the downhill grade, because almost no policy traverses either.

\begin{table*}[!tb]
  \centering\scriptsize
  \caption{\textbf{Rough-training gain against soil-training gain per seed.} Survival gain of rough over flat training and of 3D resistive force theory (3D RFT) over rigid training on the same cell of combined (Comb.) tiles at the training gravity, scored on traversal on the \SI{30}{\degree} rows marked $^\dagger$, for seeds 42, 43, and 44, and their difference. Earth rows use the Earth-trained policies and Earth sand for the soil cell, Moon rows the Moon-trained policies and the nominal lunar soil. The sign rule is met when the difference is positive with a per-seed Newcombe \SI{95}{\percent} interval that excludes zero in all three seeds.}
  \label{tab:roughvssoil}
  \begin{tabular*}{\linewidth}{@{\extracolsep{\fill}}|l|l|ccc|ccc|ccc|c|}
\hline
& & \multicolumn{3}{c|}{\textbf{Rough gain}} & \multicolumn{3}{c|}{\textbf{Soil gain}} & \multicolumn{3}{c|}{\textbf{Rough minus soil}} & \\
\hdrstrut \textbf{Gravity} & \textbf{Cell} & 42 & 43 & 44 & 42 & 43 & 44 & 42 & 43 & 44 & \textbf{Rule met} \\
\hline\hline
Earth & Comb. tiles, rigid & $+0.48$ & $+0.48$ & $+0.42$ & $+0.12$ & $+0.19$ & $+0.20$ & $+0.36$ & $+0.29$ & $+0.22$ & yes \\
Earth & Comb. tiles, soil & $+0.06$ & $-0.04$ & $+0.09$ & $+0.48$ & $+0.52$ & $+0.27$ & $-0.42$ & $-0.56$ & $-0.18$ & no \\
Earth & 20$^\circ$ up & $+0.56$ & $+0.73$ & $+0.64$ & $-0.42$ & $-0.27$ & $-0.14$ & $+0.98$ & $+1.00$ & $+0.79$ & yes \\
Earth & 20$^\circ$ down & $+0.62$ & $+0.91$ & $-0.06$ & $-0.04$ & $+0.12$ & $+0.11$ & $+0.66$ & $+0.78$ & $-0.18$ & no \\
Earth & 30$^\circ$ up$^\dagger$ & $+0.00$ & $+0.00$ & $+0.00$ & $+0.00$ & $+0.00$ & $+0.00$ & $+0.00$ & $+0.00$ & $+0.00$ & no \\
Earth & 30$^\circ$ down$^\dagger$ & $+0.00$ & $+0.00$ & $+0.00$ & $+0.00$ & $+0.00$ & $+0.00$ & $+0.00$ & $+0.00$ & $+0.00$ & no \\
\hline
Moon & Comb. tiles, rigid & $+0.22$ & $+0.20$ & $+0.14$ & $-0.04$ & $-0.08$ & $-0.09$ & $+0.25$ & $+0.27$ & $+0.23$ & yes \\
Moon & Comb. tiles, soil & $+0.04$ & $+0.02$ & $+0.01$ & $+0.01$ & $+0.04$ & $-0.04$ & $+0.03$ & $-0.03$ & $+0.05$ & no \\
Moon & 20$^\circ$ up & $+0.77$ & $+0.89$ & $+0.56$ & $+0.16$ & $+0.10$ & $-0.24$ & $+0.61$ & $+0.79$ & $+0.80$ & yes \\
Moon & 20$^\circ$ down & $+0.50$ & $-0.50$ & $+0.58$ & $-0.08$ & $+0.05$ & $-0.07$ & $+0.58$ & $-0.55$ & $+0.65$ & no \\
Moon & 30$^\circ$ up$^\dagger$ & $+0.02$ & $+0.00$ & $+0.09$ & $+0.00$ & $+0.00$ & $+0.00$ & $+0.02$ & $+0.00$ & $+0.09$ & no \\
Moon & 30$^\circ$ down$^\dagger$ & $+0.00$ & $+0.00$ & $+0.01$ & $+0.00$ & $+0.00$ & $+0.00$ & $+0.00$ & $+0.00$ & $+0.01$ & no \\
\hline
\multicolumn{12}{@{}p{\linewidth}@{}}{$^\dagger$Traversal $T$ on the \SI{26}{\degree} and \SI{30}{\degree} grades, near or above the \SI{26.6}{\degree} friction-cone limit, where a robot can slide while upright. $T$ counts an episode that survives with positive progress along the command below 1.5 times its flat-ground progress. Other rows use survival $S$.} \\
\end{tabular*}

\end{table*}
% Source: tables/tab_roughvssoil.tex written by LuPNT-private projects/guillemc/RAL_Humanoid/figures/sign_rule_table.py from eval_best, 256 episodes per cell and seed. Gains from \EarthRoughRigid, \EarthFlatRigid, \EarthFlatRFT at Earth and \MoonRoughRigid, \MoonFlatRigid, \MoonFlatRFT at the Moon.

\subsection{Rigid Control of the Gravity-Randomized Policy}
\label{sec:rigidcontrol-supp}
% Comparison: per-seed differences between Earth+Moon-Rough-RFT and its rigid control, held-out soils not run
The rigid control \JointRoughRigid shares the gravity randomization and rough training of \JointRoughRFT but trains on rigid ground only, and we test it on the same 56 cells with the same evaluator. Because the two policies trained on different GPU hosts, seed indices do not pair runs. We therefore compare seed means and spreads, and apply the sign rule of Section~\ref{sec:protocol} to the differences \JointRoughRFT minus \JointRoughRigid by seed index only to keep the test of the main text. On Earth sand the differences of 0.57 to 0.96 exceed every seed spread. The sign rule licenses a gain of \JointRoughRFT on Earth sand, with per-seed differences of $+0.91$, $+0.96$, and $+0.70$ on flat sand and $+0.72$, $+0.86$, and $+0.64$ on the combined sand tiles, and likewise of $+0.57$ to $+0.96$ on the crater, rock, and undulation sand tiles at Earth. It also licenses a small gain on the combined tiles under the nominal lunar soil at Earth gravity, of $+0.06$, $+0.07$, and $+0.05$. In contrast, the sign rule licenses no difference on any rigid cell at either gravity, including every tile family, the slopes up to \SI{20}{\degree}, and the traversal $T$ on the 26 and \SI{30}{\degree} grades, nor on any 3D RFT soil cell at lunar gravity. On the \SI{20}{\degree} downhill grade, for example, the differences are $-0.02$, $-0.02$, and $+0.16$ at Earth and $+0.13$, $-0.01$, and $+0.05$ at the Moon. We did not run the held-out MPM, SCM, and CRM soils for the rigid control, so this comparison rests on 3D RFT, the training contact model of \JointRoughRFT.
% Source: output/RAL_Humanoid/eval_c3_joint_rough_rigid/c3_summary.md, c3_results.json and c3_analysis_stdout.txt (analysis projects/guillemc/RAL_Humanoid/analysis/c3_joint_rough_rigid.py, LuPNT 0dd639798; sign rule as sign_rule_table.py at 2ccc2e119, reproduces the C8 numbers). EarthMoon-Rough-Rigid s42-44 model_10000, 56 cells x 256 episodes, ws6 RTX 5090; SoilTiles{Craters,Rocks,Undulations}_Earth +0.88/+0.95/+0.70, +0.89/+0.96/+0.69, +0.79/+0.91/+0.57; SoilFlatRegolith_Earth 0.00 in all seeds (none). Re-running EarthMoon-Rough-RFT with today's evaluator moves 8 of 15 cells by at most 0.023 and changes no verdict.

\subsection{Seed Ranges of the Main-Text Tables}
\label{sec:seedranges-supp}
% Scope: seed ranges of every cell of the main-text tables
Table~\ref{tab:keycells-full} repeats Table~\ref{tab:gravity} of the main text with the range over the three seeds in brackets wherever the seeds differ, and the seed ranges of Table~\ref{tab:soil_results} are in \texttt{soil\_rhov\_seed\_ranges.csv}, next to the per-policy intervals of Table~\ref{tab:rhov_intervals}. Table~\ref{tab:keycells-full} also adds the \SI{30}{\degree} grades and the six Moon-trained and fine-tuned policies at Earth. Like Table~\ref{tab:gravity}, it also lists the rigid control \JointRoughRigid (Section~\ref{sec:rigidcontrol-supp}). Tables~\ref{tab:terrain} and~\ref{tab:route} give their seed ranges directly.

\begin{table*}[!tb]
  \centering\footnotesize
  \caption{\textbf{Key test cells with seed ranges and the \SI{30}{\degree} grades.} Survival on the key test cells at both gravities, with Earth sand in the soil columns at Earth and lunar soil at the Moon. Above the \SI{26.6}{\degree} friction-cone limit the survivors mostly slide, so the \SI{30}{\degree} columns report survival and not traversal (Table~\ref{tab:slope_traversal}). RFT in policy names is 3D resistive force theory (3D RFT).}
  \label{tab:keycells-full}
  \setlength{\tabcolsep}{3pt}
  \begin{tabular*}{\linewidth}{@{\extracolsep{\fill}}|l|cc|cc|cc|cc|}
\hline
& \multicolumn{2}{c|}{\textbf{Flat}} & \multicolumn{2}{c|}{\textbf{Combined tiles}} & \multicolumn{2}{c|}{\textbf{Slope \SI{20}{\degree}}} & \multicolumn{2}{c|}{\textbf{Slope \SI{30}{\degree}}} \\
\hdrstrut \textbf{Policy} & Rigid & Soil & Rigid & Soil & Up & Down & Up & Down \\
\hline\hline
\multicolumn{9}{|l|}{\textbf{At Earth}} \\ \hline
\EarthFlatRigid & 1.00 & 0.96\,{\tiny[0.92,0.99]} & 0.53\,{\tiny[0.50,0.56]} & 0.27\,{\tiny[0.22,0.34]} & 0.36\,{\tiny[0.27,0.44]} & 0.19\,{\tiny[0.08,0.37]} & 0.00 & 0.00 \\
\EarthFlatRFT & 1.00\,{\tiny[0.99,1.00]} & 1.00 & 0.70\,{\tiny[0.62,0.76]} & 0.69\,{\tiny[0.61,0.76]} & 0.08\,{\tiny[0.00,0.21]} & 0.26\,{\tiny[0.21,0.33]} & 0.00 & 0.00 \\
\EarthRoughRigid & 1.00 & 0.33\,{\tiny[0.11,0.54]} & 0.98\,{\tiny[0.98,0.99]} & 0.30\,{\tiny[0.20,0.43]} & 1.00 & 0.68\,{\tiny[0.05,0.99]} & 0.00 & 0.00 \\
\EarthRoughRFT & 1.00 & 0.99\,{\tiny[0.98,1.00]} & 0.98\,{\tiny[0.97,0.98]} & 0.87\,{\tiny[0.72,0.95]} & 0.99\,{\tiny[0.97,1.00]} & 0.98\,{\tiny[0.93,1.00]} & 0.00 & 0.00 \\
\hline
\MoonFlatRigid & 0.00 & 0.00 & 0.00 & 0.00 & 0.00 & 0.00 & 0.00 & 0.00 \\
\MoonFlatRFT & 0.00 & 0.00 & 0.00 & 0.00 & 0.00 & 0.00 & 0.00 & 0.00 \\
\MoonRoughRigid & 0.00 & 0.00 & 0.00 & 0.00 & 0.00 & 0.00 & 0.00 & 0.00 \\
\MoonRoughRFT & 0.00 & 0.00 & 0.00 & 0.00 & 0.00 & 0.00 & 0.00 & 0.00 \\
\hline
\FtFlatRigid & 0.00 & 0.00 & 0.00 & 0.00 & 0.00 & 0.00 & 0.00 & 0.00 \\
\FtRoughRFT & 0.00 & 0.00 & 0.00 & 0.00 & 0.00 & 0.00 & 0.00 & 0.00 \\
\JointFlatRigid & 0.92\,{\tiny[0.88,0.96]} & 0.00 & 0.29\,{\tiny[0.24,0.34]} & 0.00 & 0.17\,{\tiny[0.00,0.50]} & 0.13\,{\tiny[0.00,0.32]} & 0.00 & 0.00 \\
\JointRoughRFT & 0.99\,{\tiny[0.96,1.00]} & 0.98\,{\tiny[0.95,1.00]} & 0.95\,{\tiny[0.94,0.96]} & 0.84\,{\tiny[0.75,0.89]} & 1.00 & 0.86\,{\tiny[0.74,0.94]} & 0.00 & 0.00 \\
\JointRoughRigid & 0.99\,{\tiny[0.98,1.00]} & 0.13\,{\tiny[0.04,0.26]} & 0.96\,{\tiny[0.95,0.97]} & 0.10\,{\tiny[0.01,0.18]} & 0.98\,{\tiny[0.96,1.00]} & 0.82\,{\tiny[0.75,0.96]} & 0.00 & 0.00 \\
\hline
\multicolumn{9}{|l|}{\textbf{At Moon}} \\ \hline
\EarthFlatRigid & 0.64\,{\tiny[0.56,0.70]} & 0.92\,{\tiny[0.84,0.98]} & 0.45\,{\tiny[0.34,0.50]} & 0.80\,{\tiny[0.70,0.88]} & 0.08\,{\tiny[0.02,0.13]} & 0.01\,{\tiny[0.00,0.02]} & 0.00 & 0.00 \\
\EarthFlatRFT & 0.67\,{\tiny[0.58,0.75]} & 0.96\,{\tiny[0.93,0.98]} & 0.43\,{\tiny[0.35,0.56]} & 0.76\,{\tiny[0.63,0.89]} & 0.05\,{\tiny[0.01,0.07]} & 0.11\,{\tiny[0.03,0.19]} & 0.00 & 0.00 \\
\EarthRoughRigid & 0.74\,{\tiny[0.62,0.80]} & 0.96\,{\tiny[0.93,0.99]} & 0.43\,{\tiny[0.31,0.55]} & 0.77\,{\tiny[0.76,0.78]} & 0.37\,{\tiny[0.12,0.70]} & 0.21\,{\tiny[0.16,0.27]} & 0.17\,{\tiny[0.04,0.39]} & 0.00 \\
\EarthRoughRFT & 0.75\,{\tiny[0.53,0.94]} & 0.84\,{\tiny[0.68,1.00]} & 0.54\,{\tiny[0.38,0.74]} & 0.78\,{\tiny[0.52,0.96]} & 0.56\,{\tiny[0.40,0.71]} & 0.16\,{\tiny[0.04,0.36]} & 0.05\,{\tiny[0.00,0.15]} & 0.00 \\
\hline
\MoonFlatRigid & 1.00 & 1.00 & 0.80\,{\tiny[0.77,0.86]} & 0.96\,{\tiny[0.94,0.98]} & 0.26\,{\tiny[0.11,0.44]} & 0.47\,{\tiny[0.42,0.50]} & 0.00 & 0.04\,{\tiny[0.00,0.09]} \\
\MoonFlatRFT & 1.00\,{\tiny[0.99,1.00]} & 1.00 & 0.74\,{\tiny[0.71,0.76]} & 0.96\,{\tiny[0.94,0.98]} & 0.26\,{\tiny[0.20,0.39]} & 0.44\,{\tiny[0.35,0.55]} & 0.00 & 0.00 \\
\MoonRoughRigid & 1.00 & 1.00 & 0.99\,{\tiny[0.98,0.99]} & 0.98\,{\tiny[0.96,0.99]} & 1.00 & 0.67\,{\tiny[0.00,1.00]} & 0.44\,{\tiny[0.04,0.66]} & 0.02\,{\tiny[0.00,0.05]} \\
\MoonRoughRFT & 0.95\,{\tiny[0.86,1.00]} & 1.00 & 0.97\,{\tiny[0.95,0.99]} & 1.00\,{\tiny[0.99,1.00]} & 0.81\,{\tiny[0.53,0.99]} & 0.99\,{\tiny[0.98,0.99]} & 0.18\,{\tiny[0.00,0.55]} & 0.13\,{\tiny[0.00,0.40]} \\
\hline
\FtFlatRigid & 1.00 & 1.00 & 0.84\,{\tiny[0.82,0.87]} & 0.96\,{\tiny[0.94,0.99]} & 0.31\,{\tiny[0.17,0.39]} & 0.67\,{\tiny[0.64,0.70]} & 0.01\,{\tiny[0.00,0.02]} & 0.00 \\
\FtRoughRFT & 1.00 & 1.00 & 0.98\,{\tiny[0.98,0.99]} & 1.00 & 0.95\,{\tiny[0.84,1.00]} & 0.99\,{\tiny[0.99,1.00]} & 0.37\,{\tiny[0.00,0.77]} & 0.64\,{\tiny[0.52,0.73]} \\
\JointFlatRigid & 1.00 & 1.00 & 0.53\,{\tiny[0.52,0.56]} & 0.86\,{\tiny[0.82,0.89]} & 0.06\,{\tiny[0.02,0.10]} & 0.17\,{\tiny[0.03,0.27]} & 0.00 & 0.00 \\
\JointRoughRFT & 1.00 & 1.00 & 0.99\,{\tiny[0.99,1.00]} & 1.00 & 0.99\,{\tiny[0.97,1.00]} & 0.96\,{\tiny[0.91,0.99]} & 0.32\,{\tiny[0.18,0.54]} & 0.00 \\
\JointRoughRigid & 1.00 & 1.00 & 0.99\,{\tiny[0.98,0.99]} & 0.99\,{\tiny[0.99,1.00]} & 1.00 & 0.90\,{\tiny[0.84,1.00]} & 0.77\,{\tiny[0.73,0.79]} & 0.02\,{\tiny[0.00,0.04]} \\
\hline
\end{tabular*}

  \tabnote{Each value is the fraction of 256 episodes without a fall, pooled over the eight command profiles, as the mean over seeds with the seed range in brackets when seeds differ. Soil cells are 3D RFT under a \SI{0.12}{\metre} layer, with an Earth medium sand at Earth and the nominal lunar soil at the Moon, and 3D RFT is the training contact model of the RFT-trained policies only. Tile cells use the 32 held-out combined tiles. Slopes are rigid, and the \SI{30}{\degree} grade lies beyond the \SI{26}{\degree} training range.}
\end{table*}
% Source: soil columns, SoilFlat_Earth / SoilTilesCombined_Earth use EARTH_TEST_SOIL (rho_c 1500, mu_int 0.55); SoilFlat_Moon / SoilTilesCombined_Moon use the nominal lunar soil (rho_c 5200, mu_int 0.9), equal to the Nominal columns of tab_lunarsoils.

\subsection{Training Strategies per Seed}
\label{sec:res_route_seed-supp}
\begin{table*}[!tb]
  \centering\footnotesize
  \caption{\textbf{Comparison of training strategies for the Moon.} RFT in policy names and in $\rho_v^{\rm RFT}$ is 3D resistive force theory (3D RFT), and $\rho_v^{\rm MPM}$ uses the material point method (MPM).}
  \label{tab:route}
  \setlength{\tabcolsep}{3pt}
  \begin{tabular*}{\linewidth}{@{\extracolsep{\fill}}|l|cccc|cc|cc|}
\hline
& \multicolumn{4}{c|}{\textbf{Moon}} & \multicolumn{2}{c|}{\textbf{Earth}} & \multicolumn{2}{c|}{\textbf{Total}} \\
\hdrstrut \textbf{Policy} & $S\uparrow$ & $e_v\downarrow$ & $\rho_v^{\rm RFT}$ & $\rho_v^{\rm MPM}$ & $S\uparrow$ & $e_v\downarrow$ & Iter. & Steps \\
\hline\hline
\MoonFlatRigid & 1.00 & 0.03\,{\tiny[0.03,0.04]} & -- & -- & 0.00 & -- & 10 & 0.98 \\
\FtFlatRigid & 1.00 & 0.03\,{\tiny[0.02,0.03]} & -- & -- & 0.00 & -- & 15 & 1.47 \\
\JointFlatRigid & 1.00 & 0.02\,{\tiny[0.02,0.03]} & -- & -- & 0.92\,{\tiny[0.88,0.96]} & 0.02\,{\tiny[0.01,0.02]} & 10 & 0.98 \\
\EarthFlatRigid & 0.64\,{\tiny[0.56,0.70]} & 0.16\,{\tiny[0.15,0.18]} & -- & -- & 1.00 & 0.01 & 10 & 0.98 \\
\hline
\MoonRoughRFT & 0.97\,{\tiny[0.95,0.99]} & 0.10\,{\tiny[0.08,0.12]} & 1.08\,{\tiny[1.03,1.18]} & 1.03\,{\tiny[0.89,1.15]} & 0.00 & -- & 10 & 0.98 \\
\FtRoughRFT & 0.98\,{\tiny[0.98,0.99]} & 0.11\,{\tiny[0.09,0.12]} & 1.04\,{\tiny[1.02,1.05]} & 0.97\,{\tiny[0.94,1.00]} & 0.00 & -- & 15 & 1.47 \\
\JointRoughRFT & 0.99\,{\tiny[0.99,1.00]} & 0.10\,{\tiny[0.10,0.11]} & 1.04\,{\tiny[1.02,1.07]} & 0.97\,{\tiny[0.88,1.09]} & 0.95\,{\tiny[0.94,0.96]} & 0.07\,{\tiny[0.05,0.08]} & 10 & 0.98 \\
\EarthRoughRFT & 0.54\,{\tiny[0.38,0.74]} & 0.28\,{\tiny[0.27,0.29]} & 1.99\,{\tiny[1.20,2.49]} & --$^\dagger$ & 0.98\,{\tiny[0.97,0.98]} & 0.05\,{\tiny[0.04,0.05]} & 10 & 0.98 \\
\hline
\end{tabular*}

  \tabnote{Upper block on rigid flat ground. Lower block on rigid combined tiles for survival $S$ and tracking error $e_v$ [m/s], and on the 3D RFT and MPM combined beds for the soil progress ratio $\rho_v$. Earth columns give retention. Iterations in thousands, steps in billions, both including the parent of a fine-tune. Means over three seeds with the seed range in brackets when seeds differ, and the episode counts behind $e_v$ in Table~\ref{tab:ev_survivors}. A dash marks a cell that does not apply. The 1.99 of \EarthRoughRFT comes from falls on the rigid reference. $^\dagger$Undefined, as the rigid reference makes short or negative progress.}
  % Source S, e_v, rho_v^RFT: output/RAL_Humanoid/eval_best/<policy>_s<seed>__{FlatRigid,TilesRigidCombined}_{Moon,Earth}.json (e_v as in tab:terrain, -- if no episode reaches the steady window) and rho_v^RFT = SoilTilesCombined_Moon over TilesRigidCombined_Moon forward progress (as in tab:soil_results) for all four lower rows, tabular by make_tab_keycells.py.
  % Source rho_v^MPM: NAVLab results/mpm_g1best_comb/comb_summary.{md,json} (summarize_comb.py), same protocol as the MPM Comb. column of tab:soil_results, Moon policies and fine-tune/joint at 1.62 on lunar soil, \EarthRoughRFT at 1.62 (rows \texttt{Earth-Rough-RFT\_atmoon}); per seed: \FtRoughRFT 0.94-1.00, \JointRoughRFT 0.88-1.09, all 6/6 survive; \EarthRoughRFT rigid progress -0.67 to 0.90 m over the 4 kept episodes, 2 of 6 excluded off the bed.
\end{table*}

% Source: tables/tab_perseed_route.tex, written by make_tab_keycells.py from eval_best {FlatRigid,TilesRigidCombined,SoilTilesCombined}_{Moon,Earth} and collab_results/mpm_g1best_comb. Iterations and steps are not per seed.

% Result: seed spread of the training strategies, the rho_v of the Earth rough-trained policy
The seeds of \EarthRoughRFT differ widely at the Moon, with survival 0.51, 0.74, and 0.38 and a 3D RFT soil progress ratio of 2.49, 1.20, and 2.27 over a rigid reference that falls often (\texttt{perseed\_route.csv}). The lunar-capable strategies vary by at most 0.05 in Moon survival across seeds, and the seeds of the three rough-trained ones among them span 0.88 to 1.15 in the MPM soil progress ratio.

\subsection{Episodes behind the Tracking Error}
\label{sec:res_ev-supp}
% Caveat: tracking error averages over the episodes that cover the last steady window to its end (gait_metrics.py COVER_SLACK_STEPS = 1.5)
The tracking error $e_v$ of Tables~\ref{tab:terrain} and~\ref{tab:route} averages over the episodes whose record reaches the end of the last steady window to within 1.5 control steps (\SI{0.03}{\second}), and Table~\ref{tab:ev_survivors} gives their number, which for \EarthRoughRigid on the combined tiles at the Moon is only 79 to 142 of 256 per seed, so a low $e_v$ there describes a selected population.
% Source: review_2026-10-24/summary.md C20, Earth-Rough-Rigid TilesRigidCombined_Moon n_ev 79 / 142 / 110.
% Source: tables/tab_ev_survivors.tex, written by make_tab_keycells.py from eval_best; survivors = episodes with survived true, e_v episodes = profile_summary.pooled.tracking.err_vx.n, the episodes steady_ev averages.
\begin{table}[!tb]
  \centering\scriptsize
  \caption{\textbf{Episodes behind each tracking error.} For every $e_v$ of Tables~\ref{tab:terrain} and~\ref{tab:route}, the number of surviving episodes (Surv.) and of episodes that enter $e_v$ ($n$), out of 256 per seed. An episode enters $e_v$ if its record reaches the end of the last steady window to within 1.5 control steps (\SI{0.03}{\second}), so only a fall in that last \SI{0.03}{\second} enters $e_v$ without surviving. A dash in the main tables marks a cell in which no episode of any seed enters $e_v$. RFT in policy names is 3D resistive force theory (3D RFT).}
  \label{tab:ev_survivors}
  \setlength{\tabcolsep}{1.8pt}
  \begin{tabular*}{\linewidth}{@{\extracolsep{\fill}}|l|l|c|cc|cc|cc|}
  \hline
  & & & \multicolumn{2}{c|}{\textbf{Seed 42}} & \multicolumn{2}{c|}{\textbf{Seed 43}} & \multicolumn{2}{c|}{\textbf{Seed 44}} \\
  \hdrstrut \textbf{Policy} & \textbf{Test} & $e_v\downarrow$ & Surv. & $n$ & Surv. & $n$ & Surv. & $n$ \\
  \hline\hline
  \multicolumn{9}{|l|}{\textbf{Table~\ref{tab:terrain}, rigid combined tiles}} \\ \hline
  \EarthFlatRigid & Earth & 0.10\,{\tiny[0.09,0.11]} & 129 & 130 & 132 & 132 & 144 & 144 \\
  \EarthFlatRigid & Moon & 0.25\,{\tiny[0.23,0.27]} & 125 & 127 & 88 & 88 & 129 & 129 \\
  \EarthRoughRigid & Earth & 0.04 & 251 & 251 & 254 & 254 & 251 & 251 \\
  \EarthRoughRigid & Moon & 0.26\,{\tiny[0.25,0.27]} & 79 & 79 & 142 & 142 & 110 & 110 \\
  \MoonFlatRigid & Earth & -- & 0 & 0 & 0 & 0 & 0 & 0 \\
  \MoonFlatRigid & Moon & 0.08\,{\tiny[0.07,0.09]} & 198 & 198 & 201 & 201 & 219 & 219 \\
  \MoonRoughRigid & Earth & -- & 0 & 0 & 0 & 0 & 0 & 0 \\
  \MoonRoughRigid & Moon & 0.10\,{\tiny[0.09,0.11]} & 254 & 254 & 251 & 251 & 254 & 254 \\
  \hline
  \multicolumn{9}{|l|}{\textbf{Table~\ref{tab:route}, flat rigid ground}} \\ \hline
  \MoonFlatRigid & Moon & 0.03\,{\tiny[0.03,0.04]} & 256 & 256 & 256 & 256 & 256 & 256 \\
  \MoonFlatRigid & Earth & -- & 0 & 0 & 0 & 0 & 0 & 0 \\
  \FtFlatRigid & Moon & 0.03\,{\tiny[0.02,0.03]} & 256 & 256 & 256 & 256 & 256 & 256 \\
  \FtFlatRigid & Earth & -- & 0 & 0 & 0 & 0 & 0 & 0 \\
  \JointFlatRigid & Moon & 0.02\,{\tiny[0.02,0.03]} & 256 & 256 & 256 & 256 & 256 & 256 \\
  \JointFlatRigid & Earth & 0.02\,{\tiny[0.01,0.02]} & 247 & 247 & 225 & 225 & 233 & 233 \\
  \EarthFlatRigid & Moon & 0.16\,{\tiny[0.15,0.18]} & 179 & 181 & 143 & 143 & 169 & 169 \\
  \EarthFlatRigid & Earth & 0.01 & 256 & 256 & 256 & 256 & 256 & 256 \\
  \hline
  \multicolumn{9}{|l|}{\textbf{Table~\ref{tab:route}, rigid combined tiles}} \\ \hline
  \MoonRoughRFT & Moon & 0.10\,{\tiny[0.08,0.12]} & 254 & 254 & 248 & 248 & 243 & 243 \\
  \MoonRoughRFT & Earth & -- & 0 & 0 & 0 & 0 & 0 & 0 \\
  \FtRoughRFT & Moon & 0.11\,{\tiny[0.09,0.12]} & 250 & 250 & 250 & 250 & 253 & 253 \\
  \FtRoughRFT & Earth & -- & 0 & 0 & 0 & 0 & 0 & 0 \\
  \JointRoughRFT & Moon & 0.10\,{\tiny[0.10,0.11]} & 255 & 255 & 254 & 254 & 253 & 253 \\
  \JointRoughRFT & Earth & 0.07\,{\tiny[0.05,0.08]} & 247 & 247 & 240 & 240 & 245 & 246 \\
  \EarthRoughRFT & Moon & 0.28\,{\tiny[0.27,0.29]} & 130 & 130 & 190 & 191 & 96 & 97 \\
  \EarthRoughRFT & Earth & 0.05\,{\tiny[0.04,0.05]} & 249 & 249 & 249 & 249 & 251 & 251 \\
  \hline
  \end{tabular*}
\end{table}

\bibliographystyle{IEEEtran}
\bibliography{refs}

% Generated by IEEEtran.bst, version: 1.14 (2015/08/26)
\begin{thebibliography}{10}
\providecommand{\url}[1]{#1}
\csname url@samestyle\endcsname
\providecommand{\newblock}{\relax}
\providecommand{\bibinfo}[2]{#2}
\providecommand{\BIBentrySTDinterwordspacing}{\spaceskip=0pt\relax}
\providecommand{\BIBentryALTinterwordstretchfactor}{4}
\providecommand{\BIBentryALTinterwordspacing}{\spaceskip=\fontdimen2\font plus
\BIBentryALTinterwordstretchfactor\fontdimen3\font minus
  \fontdimen4\font\relax}
\providecommand{\BIBforeignlanguage}[2]{{%
\expandafter\ifx\csname l@#1\endcsname\relax
\typeout{** WARNING: IEEEtran.bst: No hyphenation pattern has been}%
\typeout{** loaded for the language `#1'. Using the pattern for}%
\typeout{** the default language instead.}%
\else
\language=\csname l@#1\endcsname
\fi
#2}}
\providecommand{\BIBdecl}{\relax}
\BIBdecl

\bibitem{nasa_artemis_2020}
\BIBentryALTinterwordspacing
{NASA}, ``Artemis plan: {NASA's} lunar exploration program overview,'' National
  Aeronautics and Space Administration, Washington, DC, USA, Tech. Rep., Sep.
  2020, accessed Oct. 2026. [Online]. Available:
  \url{https://www.nasa.gov/wp-content/uploads/2020/12/artemis_plan-20200921.pdf}
\BIBentrySTDinterwordspacing

\bibitem{nasa_experience_2023}
{NASA Johnson Space Center}, ``{NASA} experience enables advance of small
  business's humanoid robot,''
  \url{https://www.nasa.gov/centers-and-facilities/johnson/nasa-experience-enables-advance-of-small-businesss-humanoid-robot/},
  2023, accessed Oct. 2026.

\bibitem{diftler_robonaut_2011}
M.~A. Diftler, F.~A. Permenter, B.~K. Hargrave, R.~Platt, R.~T. Savely, R.~O.
  Ambrose, J.~S. Mehling, M.~E. Abdallah, N.~A. Radford, L.~B. Bridgwater,
  A.~M. Sanders, R.~S. Askew, D.~M. Linn, and J.~D. Yamokoski, ``Robonaut 2 --
  the first humanoid robot in space,'' in \emph{Proc. IEEE Int. Conf. Robotics
  and Automation (ICRA)}, 2011, pp. 2178--2183.

\bibitem{carr2009apollo}
C.~E. Carr and J.~McGee, ``The {Apollo} number: Space suits, self-support, and
  the walk-run transition,'' \emph{PLoS ONE}, vol.~4, no.~8, p. e6614, 2009.

\bibitem{thuro_characterization_2021}
A.~Thuro and L.~A. Stirling, ``Characterization of the {Apollo} astronaut lunar
  extravehicular activity falls and near-falls,'' in \emph{Proc. IEEE Aerospace
  Conference}, 2021, pp. 1--6.

\bibitem{radosavovic_realworld_2024}
I.~Radosavovic, T.~Xiao, B.~Zhang, T.~Darrell, J.~Malik, and K.~Sreenath,
  ``Real-world humanoid locomotion with reinforcement learning,'' \emph{Science
  Robotics}, vol.~9, no.~89, p. eadi9579, 2024.

\bibitem{radosavovic_challenging_2024}
I.~Radosavovic, S.~Kamat, T.~Darrell, and J.~Malik, ``Learning humanoid
  locomotion over challenging terrain,'' \emph{arXiv preprint
  arXiv:2410.03654}, 2024.

\bibitem{hartmann_saturation_1984}
W.~K. Hartmann, ``Does crater ``saturation equilibrium'' occur in the solar
  system?'' \emph{Icarus}, vol.~60, no.~1, pp. 56--74, 1984.

\bibitem{di_rock_2016}
K.~Di, B.~Xu, M.~Peng, Z.~Yue, Z.~Liu, W.~Wan, L.~Li, and J.~Zhou, ``Rock
  size-frequency distribution analysis at the {Chang'E-3} landing site,''
  \emph{Planetary and Space Science}, vol. 120, pp. 103--112, 2016.

\bibitem{carrier_physical_1991}
W.~D. Carrier, III, G.~R. Olhoeft, and W.~Mendell, ``Physical properties of the
  lunar surface,'' in \emph{Lunar Sourcebook: A User's Guide to the Moon},
  G.~H. Heiken, D.~Vaniman, and B.~M. French, Eds.\hskip 1em plus 0.5em minus
  0.4em\relax Cambridge University Press, 1991, ch.~9, pp. 475--594.

\bibitem{klein_training_2025}
\BIBentryALTinterwordspacing
B.~Klein and J.~Roberts, ``Training humanoid robots to walk in lunar gravity
  using reinforcement learning,'' in \emph{Proc. Australasian Conference on
  Robotics and Automation (ACRA)}.\hskip 1em plus 0.5em minus 0.4em\relax ARAA,
  2025, queensland University of Technology. [Online]. Available:
  \url{https://eprints.qut.edu.au/262116/}
\BIBentrySTDinterwordspacing

\bibitem{arm_efficient_2025}
P.~Arm, O.~Fischer, J.~Church, A.~Fuhrer, H.~Kolvenbach, and M.~Hutter,
  ``Efficient learning-based control of a legged robot in lunar gravity,''
  arXiv preprint arXiv:2509.10128, 2025.

\bibitem{kamohara_granular_2026}
J.~Kamohara, F.~Wu, A.~N. Zong, D.~I. Goldman, Y.~Nakka, S.~Hutchinson, and
  Y.~Zhao, ``{GM-Loco}: Terrain-adaptive humanoid locomotion on granular
  media,'' \emph{arXiv preprint arXiv:2609.10286}, 2026.

\bibitem{belmontebaeza_learning_2026}
A.~Belmonte-Baeza, M.~Cazorla, G.~J. Garc{\'i}a, C.~J. P{\'e}rez-Del-Pulgar,
  and J.~Pomares, ``Learning safety-constrained whole-body loco-manipulation
  for lunar surface operations,'' arXiv preprint arXiv:2610.15486, 2026.

\bibitem{vyas_locomotion_2026}
Y.~J. Vyas, ``Locomotion analysis of a quadruped interacting with the lunar
  granular surface,'' arXiv preprint arXiv:2606.10273, 2026.

\bibitem{li_terradynamics_2013}
C.~Li, T.~Zhang, and D.~I. Goldman, ``A terradynamics of legged locomotion on
  granular media,'' \emph{Science}, vol. 339, no. 6126, pp. 1408--1412, 2013.

\bibitem{agarwal_mechanistic_2023}
S.~Agarwal, D.~I. Goldman, and K.~Kamrin, ``Mechanistic framework for
  reduced-order models in soft materials: Application to three-dimensional
  granular intrusion,'' \emph{Proceedings of the National Academy of Sciences},
  vol. 120, no.~4, p. e2214017120, 2023.

\bibitem{luo_sonic_2026}
\BIBentryALTinterwordspacing
Z.~Luo, Y.~Yuan, T.~Wang \emph{et~al.}, ``{SONIC}: Supersizing motion tracking
  for natural humanoid whole-body control,'' \emph{Science Robotics}, vol.~11,
  no. 117, p. eaed4592, 2026. [Online]. Available:
  \url{https://arxiv.org/abs/2511.07820}
\BIBentrySTDinterwordspacing

\bibitem{peng_simtoreal_2018}
X.~B. Peng, M.~Andrychowicz, W.~Zaremba, and P.~Abbeel, ``Sim-to-real transfer
  of robotic control with dynamics randomization,'' in \emph{Proc. IEEE Int.
  Conf. Robot. Autom. (ICRA)}, 2018, pp. 3803--3810.

\bibitem{he_ultra_2026}
X.~He, S.~Xu, X.~Li, R.~Dong, L.~Bian, Y.-X. Wang, and L.-Y. Gui, ``{ULTRA}:
  Unified multimodal control for autonomous humanoid whole-body
  loco-manipulation,'' arXiv preprint arXiv:2603.03279, 2026.

\bibitem{kolvenbach_spacebok_2019}
P.~Arm, R.~Zenkl, P.~Barton \emph{et~al.}, ``{SpaceBok}: A dynamic legged robot
  for space exploration,'' in \emph{Proc. IEEE Int. Conf. Robotics and
  Automation (ICRA)}, 2019, pp. 6288--6294.

\bibitem{olsen_olympus_2025}
J.~A. Olsen, G.~Malczyk, and K.~Alexis, ``Olympus: A jumping quadruped for
  planetary exploration utilizing reinforcement learning for in-flight attitude
  control,'' in \emph{Proc. IEEE International Conference on Robotics and
  Automation (ICRA)}, 2025.

\bibitem{unitree_rl_lab_2025}
{Unitree Robotics}, ``Unitree {RL} lab: Reinforcement learning environments for
  {Unitree} robots, built on {Isaac Lab},''
  \url{https://github.com/unitreerobotics/unitree_rl_lab}, 2025, open-source
  software, Apache-2.0 license. Accessed Oct. 2026.

\bibitem{griffin_walking_1999}
T.~M. Griffin, N.~A. Tolani, and R.~Kram, ``Walking in simulated reduced
  gravity: Mechanical energy fluctuations and exchange,'' \emph{Journal of
  Applied Physiology}, vol.~86, no.~1, pp. 383--390, 1999.

\bibitem{pavei_skipping_2015}
G.~Pavei, C.~M. Biancardi, and A.~E. Minetti, ``Skipping vs. running as the
  bipedal gait of choice in hypogravity,'' \emph{Journal of Applied
  Physiology}, vol. 119, no.~1, pp. 93--100, 2015.

\bibitem{kram1997effect}
R.~Kram, A.~Domingo, and D.~P. Ferris, ``Effect of reduced gravity on the
  preferred walk--run transition speed,'' \emph{Journal of Experimental
  Biology}, vol. 200, no.~4, pp. 821--826, 1997.

\bibitem{de_witt_lunar_2014}
J.~K. De~Witt, W.~B. Edwards, M.~M. Scott-Pandorf, J.~R. Norcross, and M.~L.
  Gernhardt, ``The preferred walk to run transition speed in actual lunar
  gravity,'' \emph{Journal of Experimental Biology}, vol. 217, no.~18, pp.
  3200--3203, 2014.

\bibitem{chen_bipedal_2026}
X.~Chen, X.~Huang, P.~Shan, J.~Yi, and T.~Liu, ``Bipedal-walking-dynamics model
  on granular terrains,'' in \emph{Proc. IEEE Int. Conf. Robotics and
  Automation (ICRA)}, 2026.

\bibitem{kerimoglu_flowable_2026}
\BIBentryALTinterwordspacing
D.~Kerimoglu, J.~Kamohara, J.~Maeng, Z.~Yoon, S.~Hutchinson, Y.~Zhao, and D.~I.
  Goldman, ``Robust bipedal locomotion on flowable slopes via foot-driven
  terrain manipulation,'' arXiv preprint arXiv:2607.11855, 2026. [Online].
  Available: \url{https://arxiv.org/abs/2607.11855}
\BIBentrySTDinterwordspacing

\bibitem{newton_contributors_2025}
{The Newton Contributors}, ``{Newton}: {GPU}-accelerated physics simulation for
  robotics and simulation research,''
  \url{https://github.com/newton-physics/newton}, 2025, open-source software.

\bibitem{costantino_lowvelocity_2011}
D.~J. Costantino, J.~Bartell, K.~Scheidler, and P.~Schiffer, ``Low-velocity
  granular drag in reduced gravity,'' \emph{Physical Review E}, vol.~83, no.~1,
  p. 011305, 2011.

\bibitem{murdoch_experimental_2017}
N.~Murdoch, I.~Avila~Martinez, C.~Sunday, E.~Zenou, O.~Cherrier, A.~Cadu, and
  Y.~Gourinat, ``An experimental study of low-velocity impacts into granular
  material in reduced gravity,'' \emph{Monthly Notices of the Royal
  Astronomical Society}, vol. 468, pp. 1259--1272, 2017.

\bibitem{chen_lowgravity_2026}
J.~Chen, R.~Li, and S.~Fu, ``Influence of low gravity on the penetration
  resistance of lunar regolith,'' \emph{npj Microgravity}, vol.~12, p.~18,
  2026.

\bibitem{niksirat_reducedgravity_2020}
\BIBentryALTinterwordspacing
P.~Niksirat, A.~Daca, and K.~Skonieczny, ``The effects of reduced-gravity on
  planetary rover mobility,'' \emph{The International Journal of Robotics
  Research}, vol.~39, no.~7, pp. 797--811, 2020. [Online]. Available:
  \url{https://journals.sagepub.com/doi/abs/10.1177/0278364920913945}
\BIBentrySTDinterwordspacing

\bibitem{rudin_learning_2022}
N.~Rudin, D.~Hoeller, P.~Reist, and M.~Hutter, ``Learning to walk in minutes
  using massively parallel deep reinforcement learning,'' in \emph{Proc.
  Conference on Robot Learning}, ser. Proceedings of Machine Learning Research,
  vol. 164, 2022, pp. 91--100.

\bibitem{casadesusvila_lupnt_2025}
G.~Casadesus~Vila, K.~Iiyama, and G.~Gao, ``{LuPNT}: An open-source simulator
  for lunar communications, positioning, navigation, and timing,'' in
  \emph{Proc. IEEE Aerospace Conf.}, 2025, pp. 1--18.

\bibitem{cai2020meter}
Y.~Cai and W.~Fa, ``Meter-scale topographic roughness of the {Moon}: The effect
  of small impact craters,'' \emph{Journal of Geophysical Research: Planets},
  vol. 125, no.~8, p. e2020JE006429, 2020.

\bibitem{sulsky_particle_1994}
D.~Sulsky, Z.~Chen, and H.~L. Schreyer, ``A particle method for
  history-dependent materials,'' \emph{Computer Methods in Applied Mechanics
  and Engineering}, vol. 118, no. 1--2, pp. 179--196, 1994.

\bibitem{tasora_overview_2017}
\BIBentryALTinterwordspacing
A.~Tasora, D.~Mangoni, and D.~Negrut, ``An overview of the chrono soil contact
  model (scm) implementation,'' Simulation-Based Engineering Laboratory,
  University of Wisconsin-Madison, Tech. Rep. TR-2017-01, 2017. [Online].
  Available:
  \url{https://sbel.wisc.edu/wp-content/uploads/sites/569/2018/05/TR-2017-01.pdf}
\BIBentrySTDinterwordspacing

\bibitem{unjhawala_physicsbased_2025}
H.~Unjhawala, L.~Bakke, H.~Zhang, M.~Taylor, G.~Arivoli, R.~Serban, and
  D.~Negrut, ``A physics-based continuum model for versatile, scalable, and
  fast terramechanics simulation,'' arXiv preprint arXiv:2507.05643, 2025.

\bibitem{brown_interval_2001}
L.~D. Brown, T.~T. Cai, and A.~DasGupta, ``Interval estimation for a binomial
  proportion,'' \emph{Statistical Science}, vol.~16, no.~2, pp. 101--133, 2001.

\bibitem{newcombe_interval_1998}
R.~G. Newcombe, ``Interval estimation for the difference between independent
  proportions: Comparison of eleven methods,'' \emph{Statistics in Medicine},
  vol.~17, no.~8, pp. 873--890, 1998.

\bibitem{wolczyk_finetuning_2024}
M.~Wolczyk, B.~Cupia{\l}, M.~Ostaszewski, M.~Bortkiewicz, M.~Zajac, R.~Pascanu,
  {\L}.~Kuci{\'n}ski, and P.~Mi{\l}o{\'s}, ``Fine-tuning reinforcement learning
  models is secretly a forgetting mitigation problem,'' in \emph{Proc. Int.
  Conf. Machine Learning (ICML)}, 2024, pp. 53\,039--53\,078.

\bibitem{ash_warmstarting_2020}
J.~T. Ash and R.~P. Adams, ``On warm-starting neural network training,'' in
  \emph{Advances in Neural Information Processing Systems}, H.~Larochelle,
  M.~Ranzato, R.~Hadsell, M.~F. Balcan, and H.~Lin, Eds., vol.~33.\hskip 1em
  plus 0.5em minus 0.4em\relax Curran Associates, Inc., 2020, pp. 3884--3894.

\bibitem{sonic_modelcard_2026}
\BIBentryALTinterwordspacing
{NVIDIA GEAR Team}, ``{SONIC} model card,'' 2026, accessed October 5, 2026.
  [Online]. Available:
  \url{https://nvlabs.github.io/GR00T-WholeBodyControl/model_card.html}
\BIBentrySTDinterwordspacing

\bibitem{sonic_planner_2026}
\BIBentryALTinterwordspacing
------, ``Kinematic planner {ONNX} model reference,'' 2026, accessed October 5,
  2026. [Online]. Available:
  \url{https://nvlabs.github.io/GR00T-WholeBodyControl/references/planner_onnx.html}
\BIBentrySTDinterwordspacing

\bibitem{hoover_depth_2024}
R.~H. Hoover, S.~J. Robbins, B.~M. Hynek, and P.~O. Hayne, ``Depth-to-diameter
  ratios of fresh craters on the {Moon} and implications for surface age
  estimates,'' \emph{The Planetary Science Journal}, vol.~5, no.~1, p.~26,
  2024.

\bibitem{pike_sizedependence_1977}
R.~J. Pike, ``Size-dependence in the shape of fresh impact craters on the
  {Moon},'' in \emph{Impact and Explosion Cratering}, D.~J. Roddy, R.~O. Pepin,
  and R.~B. Merrill, Eds.\hskip 1em plus 0.5em minus 0.4em\relax Pergamon
  Press, 1977, pp. 489--509.

\bibitem{golombek_sizefrequency_1997}
M.~Golombek and D.~Rapp, ``Size-frequency distributions of rocks on {Mars} and
  {Earth} analog sites: Implications for future landed missions,''
  \emph{Journal of Geophysical Research: Planets}, vol. 102, no.~E2, pp.
  4117--4129, 1997.

\bibitem{rosenburg2011slopes}
M.~A. Rosenburg, O.~Aharonson, J.~W. Head, M.~A. Kreslavsky, E.~Mazarico, G.~A.
  Neumann, D.~E. Smith, M.~H. Torrence, and M.~T. Zuber, ``Global surface
  slopes and roughness of the {Moon} from the {Lunar} {Orbiter} {Laser}
  {Altimeter},'' \emph{Journal of Geophysical Research}, vol. 116, no.~E2, p.
  E02001, 2011.

\bibitem{agarwal_surprising_2021}
S.~Agarwal, A.~Karsai, D.~I. Goldman, and K.~Kamrin, ``Surprising simplicity in
  the modeling of dynamic granular intrusion,'' \emph{Science Advances},
  vol.~7, no.~17, p. eabe0631, 2021.

\bibitem{kerimoglu_extending_2025}
D.~Kerimoglu, E.~Marteau, D.~Soto, and D.~I. Goldman, ``Extending granular
  resistive force theory to cohesive powder-scale media,'' \emph{Journal of
  Terramechanics}, vol. 120, p. 101058, 2025.

\bibitem{agarwal2021deep}
R.~Agarwal, M.~Schwarzer, P.~S. Castro, A.~Courville, and M.~G. Bellemare,
  ``Deep reinforcement learning at the edge of the statistical precipice,'' in
  \emph{Advances in Neural Information Processing Systems}, vol.~34, 2021, pp.
  29\,304--29\,320.

\end{thebibliography}
\end{document}